%% file: neurips_2025.tex
\documentclass{article}

\usepackage[preprint]{neurips_2025}

\usepackage[utf8]{inputenc} 
\usepackage[T1]{fontenc}    
\usepackage{hyperref}       
\usepackage{url}            
\usepackage{booktabs}       
\usepackage{amsfonts}       
\usepackage{nicefrac}       
\usepackage{microtype}      
\usepackage[frozencache,cachedir=.]{minted}
\usepackage[dvipsnames]{xcolor}
\usepackage{colortbl}
\usepackage{titletoc}   
\usepackage{listings}

\usepackage{pifont}
\usepackage{fontawesome}
\usepackage{enumitem}
\usepackage[most, breakable, skins]{tcolorbox}

\usepackage{makecell}
\usepackage{multirow}
\usepackage{caption}
\usepackage{wrapfig}
\usepackage{adjustbox} 
\usepackage{subfigure}
\usepackage{anyfontsize}

\definecolor{citecolor}{HTML}{1741DC}
\definecolor{pink}{HTML}{DB6199}
\hypersetup{
    colorlinks,
    linkcolor=citecolor,   
    urlcolor=pink,
    citecolor=citecolor
}

\newcommand{\benchmark}{\textsc{M$^2$Eval}}
\newcommand{\instruct}{\textsc{M$^2$c-Instruct}}
\newcommand{\coder}{\textsc{M$^2$-Coder}}
\newcommand{\codegraph}{Visual Workflow}

\newcommand{\rightsym}{\textcolor{green}{\ding{51}}}
\newcommand{\wrongsym}{\textcolor{red}{\ding{55}}}

\title{Multilingual Multimodal Software Developer 
for \\ Code Generation}
\author{
  {\bf Linzheng Chai}\textsuperscript{1},\quad
  {\bf Jian Yang}\textsuperscript{1}\thanks{Corresponding Author.},\quad 
  {\bf Shukai Liu}\textsuperscript{1},\quad
  {\bf Wei Zhang}\textsuperscript{1},\quad
  {\bf Liran Wang}\textsuperscript{1}, \\
  {\bf Ke Jin}\textsuperscript{1},\quad 
  {\bf Tao Sun}\textsuperscript{1},\quad
  {\bf Congnan Liu}\textsuperscript{2},\quad 
  {\bf Chenchen Zhang}\textsuperscript{3},\quad
  {\bf Hualei Zhu}\textsuperscript{1},\\
  {\bf Jiaheng Liu}\textsuperscript{4},\quad
  {\bf Xianjie Wu}\textsuperscript{1},\quad 
  {\bf Ge Zhang}\textsuperscript{3},\quad
  {\bf Tianyu Liu}\textsuperscript{3},\quad 
  {\bf Zhoujun Li}\textsuperscript{1} \\
 \textsuperscript{\rm 1}Beihang University,\quad
 \textsuperscript{\rm 2}Alibaba Group,\quad
\textsuperscript{\rm 3}M-A-P,\quad
 \textsuperscript{\rm 4}Nanjing University \\
  \texttt{chailinzheng@buaa.edu.cn,~~jiayang@buaa.edu.cn } \\
  \textcolor{blue}{\coder:} \url{https://github.com/MCEVAL/MMCoder}
}

\begin{document}

\maketitle

\begin{abstract}

The rapid advancement of Large Language Models (LLMs) has significantly improved code generation, yet most models remain text-only, neglecting crucial visual aids like diagrams and flowcharts used in real-world software development. To bridge this gap, we introduce \coder{}, a Multilingual Multimodal software developer. \coder{} integrates visual design inputs—Unified Modeling Language (UML) diagrams and flowcharts (termed \codegraph{})—with textual instructions to enhance code generation accuracy and architectural alignment. To enable this, we developed \instruct{}, a diverse multimodal instruction-tuning dataset including visual-workflow-based code generation, allowing \coder{} to synthesize textual and graphical information like human developers, distinct from prior work on narrow tasks. Furthermore, we introduce \benchmark{}, a new benchmark for evaluating multimodal code generation, addressing existing text-only limitations. Our evaluations using \benchmark{} highlight significant remaining challenges for models in precise visual information capture, instruction following, and advanced programming knowledge.
Our work aims to revolutionize industrial programming by enabling LLMs to interpret and implement complex specifications conveyed through both text and visual designs.

\end{abstract}

\section{Introduction}
The emergence of large language models (LLMs) and Large Multimodal Models(LMMs), such as Claude3~\citep{claude3} and GPT4o/GPT4.5~\citep{gpt4,gpt45}, has shown exceptional performance in various code-related tasks, particularly in code generation. Meanwhile, open-source code LLMs like StarCoder~\citep{starcoder}, DeepSeekCoder~\citep{deepseek_coder} also deliver competitive results on foundational text-only code generation benchmarks, such as LiveCodeBench~\citep{livecodebench}, Aider~\citep{aider}, SWE-Bench~\citep{swe_bench}, and BigCodeBench~\citep{bigcodebench}. 
However, these code LLMs involve processing exclusively textual inputs. This contrasts sharply with industrial programming, where translating complex requirements into functional code often relies heavily on visual aids like diagrams, UI mockups, and schematics to improve clarity and collaboration. 

\begin{figure*}[t]
\begin{center}
    \includegraphics[width=0.9\textwidth]{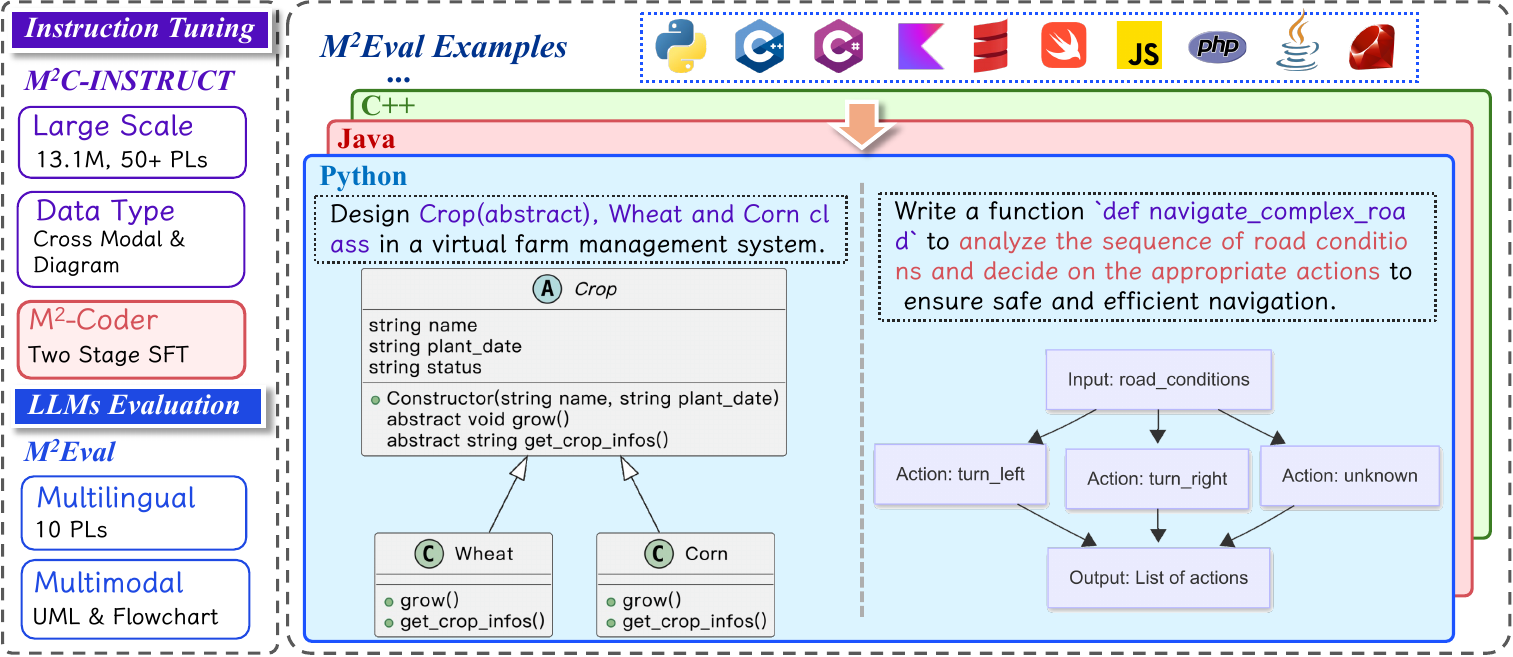}
    \caption{
    Overview of \coder{}, \instruct{} and \benchmark{}. We constructed \instruct{}, an instruction tuning dataset with over 13.1 million instances, to enhance the multilingual multimodal programming capabilities of \coder{}. For evaluating \coder{}, we curate \benchmark{}, a multimodal benchmark for code evaluation across 10 programming languages.
    }
    \label{fig:intro}
    \vspace{-15pt}
\end{center}
\end{figure*}

There has been some recent work on visual input for code LMMs, such as Design2Code~\citep{design2code}, Web2Code~\citep{web2code}, MatPlotBench~\citep{yang2024matplotagent}, and ChartCoder~\citep{zhao2025chartcoder}. These methods focus on restoring the source code from the image input, including graph-to-code generation. Most previous works restore the functional code that can render the corresponding image, so the code generation task is limited only to the narrow field of code-based drawing (creating visuals using programming languages). In the industrial scenario, the software architect often first provides a visual representation of system architecture/design patterns/process workflows in \autoref{fig:intro}, and then the software developer implements the actual code. Unlike human developers, who seamlessly integrate textual and graphical information to understand intricate specifications, current code LLMs operate solely on text-based inputs, limiting their effectiveness in scenarios where visual context is critical. 


To bridge the gap between current code generation models and the nuanced understanding of human software developers, we propose the \textbf{M}ultilingual \textbf{M}ultimodal software developer for \textbf{Code} generation (\coder{}). This model is designed to effectively incorporate visual software designing information directly into the code generation pipeline. We utilize UML (Unified Modeling Language) diagrams and flowcharts as ``\codegraph{}'', as these provide indispensable visual representations of system architecture, processes, and logic before coding commences. As depicted in \autoref{fig:intro}, UML diagrams facilitate complex system design and documentation, while flowcharts clarify algorithmic logic, thus enabling \coder{} to grasp higher-level design intent. Specifically, \coder{} undergoes a two-stage training: initial pre-training on diverse large-scale multimodal code-related instruction samples, followed by fine-tuning on a high-quality instruction corpus emphasizing visual-workflow-based code generation. Finally, to accurately evaluate the performance of code LMMs like \coder{}, we introduce a new benchmark, \benchmark{}.
The contributions and insights are summarized as:
\vspace{-0.2cm}
\begin{itemize}[leftmargin=*]
    \item We introduce \instruct{}, a large-scale multilingual multimodal code generation dataset comprising over 13.1M samples across 50 programming languages. Building upon this, we propose \coder{}, a novel multimodal code generation model trained on \instruct{} that leverages visual design inputs, such as UML diagrams and flowcharts, to improve code generation accuracy and alignment with architectural intent.
    \vspace{-0.10cm}
    \item We present \benchmark{}, a new benchmark for evaluating multilingual multimodal code generation. \benchmark{} significantly expands the scope of evaluation in this field by incorporating a broader range of programming languages and diverse task types.
    \vspace{-0.10cm}
    \item Our systematic evaluation of \coder{} and popular LMMs on \benchmark{} demonstrates our 7B \coder{}'s competitiveness with larger 70B+ LMMs. Crucially, our analysis uncovers persistent LMM limitations in precise visual understanding, instruction following, and applying higher-order programming concepts, offering valuable directions for the field.
    \vspace{-0.10cm}
\end{itemize}


%

\section{\benchmark{} Benchmark}

\subsection{Task Definition}


Given the $k$-th language $L_k$ from the set ${L_1, ..., L_K}$ of $K$ programming languages(PLs), we input the instruction $I^{L_k}$ and diagram $D$ into the LMM $\mathcal{M}$ to generate a code-related response $c^{L_k}$, sampled from the distribution $P(c^{L_k} \mid I^{L_k}, D; \mathcal{M})$. We then extract the code from $c^{L_k}$ and execute it with the corresponding test cases $u^{L_k}$ to obtain the final output. The output is expected to match the results specified by the test cases.

The process can be described as:
\begin{MiddleEquation}
\begin{align}
    r^{L_{k}} = \mathbb{I}(P(c^{L_{k}}|I^{L_{k}}, D;\mathcal{M}); u^{L_{k}})
    \label{eval_code_generation}
\end{align}
\end{MiddleEquation}where $\mathbb{I}(\cdot)$ is the indicator function by executing the generated code with the given test cases $u^{L_{k}}$. when the generated code $c^{L_{k}}$ passes all test cases, the evaluation result $r=1$, else $r=0$.

\subsection{Data Curation Process}

\begin{figure*}[ht]
\begin{center}
    \includegraphics[width=0.95\textwidth]{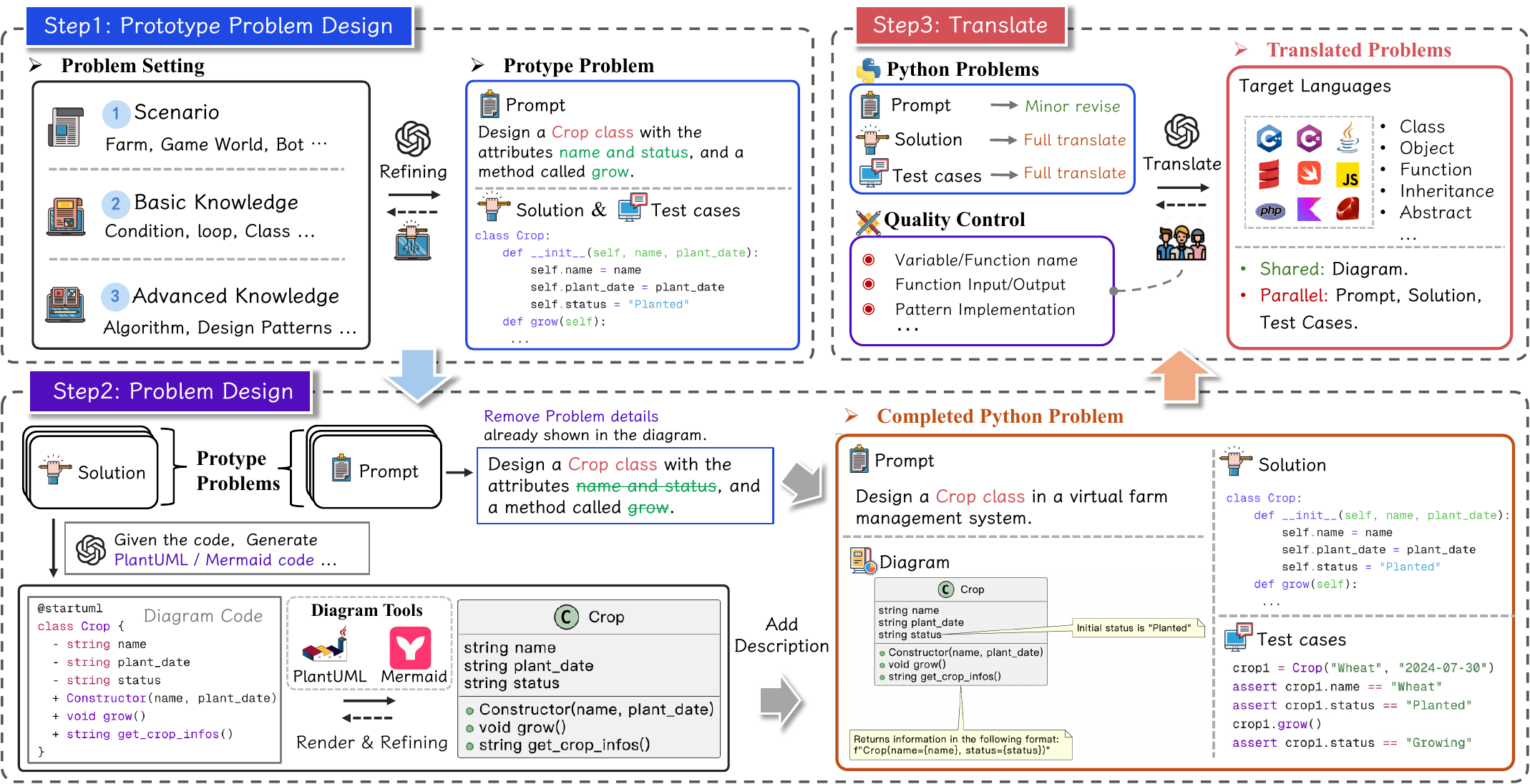}
    \caption{The curation process for \benchmark{}: (1) designing Python prototype problems grounded in core programming concepts; (2) transforming these into multimodal problems by incorporating diagrams and refining prompts; and (3) translating the problems into multiple programming languages.}
    
    \label{fig:bench_curation}
    \vspace{-10pt}
\end{center}
\end{figure*}

As illustrated in \autoref{fig:bench_curation}, we follow a three-step process to build high-quality \benchmark{}: (1) design prototype problems in Python based on common programming concepts; (2) convert them into multimodal problems by adding diagrams and removing redundant text; and (3) translate them into multiple programming languages. A general overview is presented here, with further details in \autoref{supp:bench}.

\textbf{Prototype Problem Design.}\quad
We design prototype Python problems based on scenarios and varying programming knowledge levels. An LLM generates initial prompts, solutions, and test cases. These are then manually refined through multiple iterations with the LLM to ensure accuracy, alignment with design goals, and comprehensive test coverage.

\textbf{Problem Design.}\quad
Next, for each prototype problem, we generate diagrams (\texttt{PlantUML/Mermaid}) from the solution code using an LLM, then manually refine these diagrams for structural and semantic accuracy. We revise the problem prompts by removing information redundant with the diagrams, making the prompt alone insufficient for a correct solution. Essential details are added directly to the diagrams to ensure the problem remains solvable when both the prompt and diagram are provided. This yields Python problems, each with a prompt, diagram, solution, and test cases.

\textbf{Translate to other PLs.}\quad
Finally, the Python problems are translated into nine other programming languages. Prompts are adapted (e.g., by modifying function names), and solutions and test cases are fully translated. This task is completed by nine volunteers, who use LLMs for assistance and manually verify the translations for accuracy, ensuring consistency in variables and input-output alignment.

\subsection{Data Statistics}
\begin{figure*}[t]
\centering
\begin{minipage}[l]{0.28\textwidth}
\small
\centering
\captionof{table}{\benchmark{} statistics.}
\begin{adjustbox}{width=\linewidth}
 \begin{tabular}{lr}
 \toprule
 \textbf{Statistic} & \textbf{Number} \\
 \midrule
    Total problems & 300 \\
    ~- PLs & 10 \\
    ~- Problems per PLs & 30 \\   
    Max. length & 388 \\
    Avg. length & 89 \\
  \midrule
    Images & 30 \\
    ~- Max. height & 1978 \\
    ~- Min. height & 159 \\
    ~- Max. width & 2081 \\
    ~- Min. width & 338 \\ 
  \midrule
  Solutions & 300 \\
  ~- Avg. Test Cases & 9 \\
  ~- Max. length & 826 \\
  ~- Avg. length & 326 \\
 \bottomrule
  \label{tab:bench_statistics}
 \end{tabular}

 \end{adjustbox}
\end{minipage}
\begin{minipage}[l]{0.35\textwidth}
\small
\centering
\captionof{table}{\instruct{} statistics.}
\centering
\fontsize{5.5pt}{\baselineskip}\selectfont 
 \renewcommand\arraystretch{0.54} 
\begin{adjustbox}{width=\linewidth}

 \begin{tabular}{lr}
 \toprule
 \textbf{Statistic} & \textbf{Number} \\
 \midrule
    Total problems & 13.1M  \\
    \midrule
    Stage 1 problems & 12.9M \\
    ~- PLs  & 50+ \\ 
    ~- Images & 42.3M \\
    ~- Image Height & 24-13345 \\ 
    ~- Image Width  & 10-57246 \\ 
    ~- Response Max. length & 5362 \\
    ~- Response Avg. length & 994 \\
    \midrule
    Stage 2 problems & 168K \\
    ~- PLs  & 20+ \\ 
    ~- Images & 168K \\
    ~- Image Height & 26-9694 \\ 
    ~- Image Width  & 125-1484 \\ 
    ~- Response Max. length & 1978 \\
    ~- Response Avg. length & 400 \\
    

 \bottomrule
  \label{tab:instruct_stats}
 \end{tabular}

 \end{adjustbox}
\end{minipage}
\begin{minipage}[c]{0.33\textwidth}
\small
\centering
\begin{adjustbox}{width=\linewidth}
\includegraphics[width=1.0\linewidth]{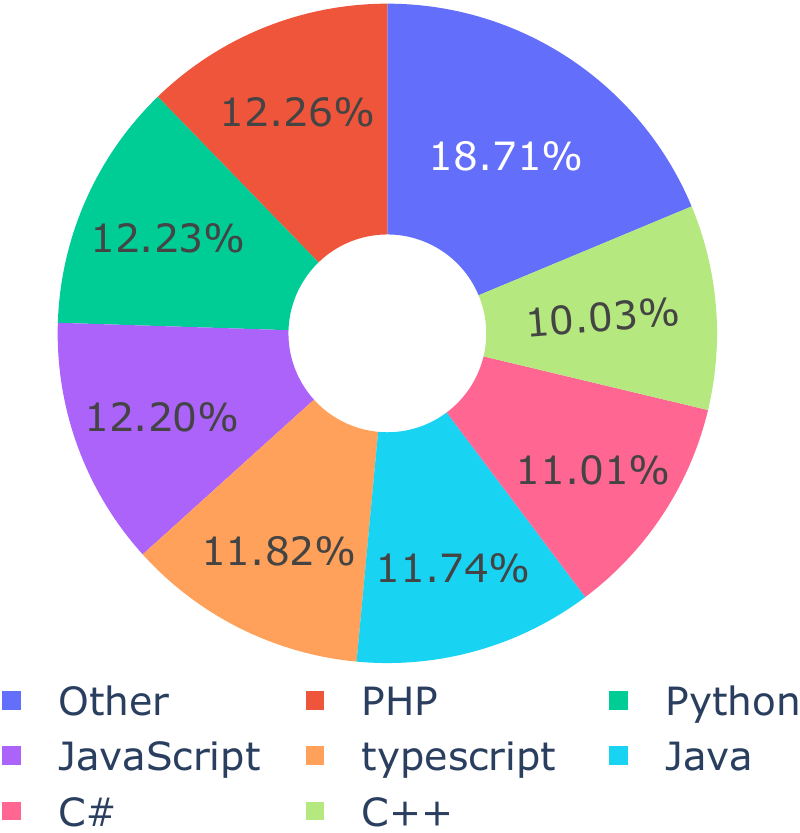}
 \end{adjustbox}
  \caption{PL distribution of \instruct{}. }
  \label{fig:instruct_data}
\end{minipage}
\vspace{-15pt}
\end{figure*}

Key statistics for the \benchmark{} benchmark are in Table~\ref{tab:bench_statistics}.
The benchmark contains 300 problems: 30 unique concepts, each in 10 Programming Languages (PLs), yielding 30 parallel instances per PL.
Problem descriptions average 89 tokens (max 388), tokenized with \texttt{Qwen2.5-Coder}.
It includes 30 distinct images, one per concept, shared across all 10 PL versions.
Image dimensions range from 159-1978px (height) and 338-2081px (width).
All 300 problems have solutions, averaging 326 tokens (max 826). Each solution is evaluated against an average of 9 test cases.

In \autoref{tab:bench_compare}, we compare \benchmark{} with other multimodal code datasets. Notably, \benchmark{} provides a significant supplement by supporting 10 programming languages (PLs), providing 13.1M samples for instruction tuning, and introducing a novel 'Visual Workflow' task type.

\begin{table}[h]
\vspace{-1em}
\caption{A comparison of our \benchmark{} to other multimodal code datasets. }
\label{tab:bench_compare}
    \centering
    \small
    \resizebox{0.95\textwidth}{!}{
    \begin{tabular}{l|ccccccc}
    \toprule
        Benchmarks & Languages & Eval & Instruct & Sources & \#Test & \#Train & Task Type \\ 
        \midrule
        Design2Code~\cite{design2code} & HTML & \rightsym & \wrongsym & Real  & 484 & - & Frontend \\ 
        Web2Code~\cite{web2code} & HTML & \rightsym & \rightsym & Synthetic & 5990 & 828K & Frontend \\ 
        \midrule
        MatPlotBench~\cite{yang2024matplotagent}  & Python & \rightsym & \wrongsym & Human Curated & 100 & - & Chart-to-Code \\ 
        Plot2Code~\cite{wu2024plot2code} & Python & \rightsym & \wrongsym & Human Curated & 132 & - & Chart-to-Code \\ 
        ChartMimic~\cite{shi2024chartmimic} & Python & \rightsym & \wrongsym  & Human Curated & 1K & - & Chart-to-Code \\ 
        ChartCoder~\cite{zhao2025chartcoder} & Python & \wrongsym & \rightsym  & Synthetic & - & 160K & Chart-to-Code \\ 
        \midrule
        SWEbench M~\cite{swe_multimodal} & JS & \rightsym & \wrongsym & Human Curated & 619 & - & Issue Resolving \\ 
        Visual SWEbench~\cite{zhang2024codev} & Python & \rightsym & \wrongsym & Human Curated & 128 & - & Chart Issue Resolving \\ 
        \midrule
        MMCode~\cite{li2024mmcode}  & Python & \rightsym & \wrongsym & Crawl & 263 & - & Algorithm \\ 
        HumanEval-V~\cite{humanevalv} & Python & \rightsym & \wrongsym & Crawl & 253 & - & Algorithm \\ 
        Code-Vision~\cite{codevision} & Python & \rightsym & \wrongsym & Human Curated & 438 & - & Algorithm \\ 
        \midrule
        \bf{\benchmark{}~(Ours)} & 10 PLs & \rightsym & \rightsym & Human Curated & 300 & 13.1M & Visual Workflow \\ 
        \bottomrule
    \end{tabular}
    }
    \vspace{-1em}
\end{table}

\section{\instruct{} and \coder{}}

\subsection{Data Overview}
Shown in Table~\ref{tab:instruct_stats} and Figure~\ref{fig:instruct_data}, \instruct{} comprises a comprehensive collection of 13.1M total problems, divided into two stages, which are used for the two-stage fine-tuning of \coder{} respectively. Stage-1 data contains 12.9M problems spanning over 50 programming languages, accompanied by 42.3M images. Stage-2 data features 168K problems across more than 20 programming languages, with 168K associated images. Stage-2 data responses are generally shorter. For more detailed information on the data statistics and construction process, please refer to the Appendix~\ref{supp:instruct}.

\subsection{Data Construction}
\begin{figure*}[t]
\centering
    \includegraphics[width=0.95\textwidth]{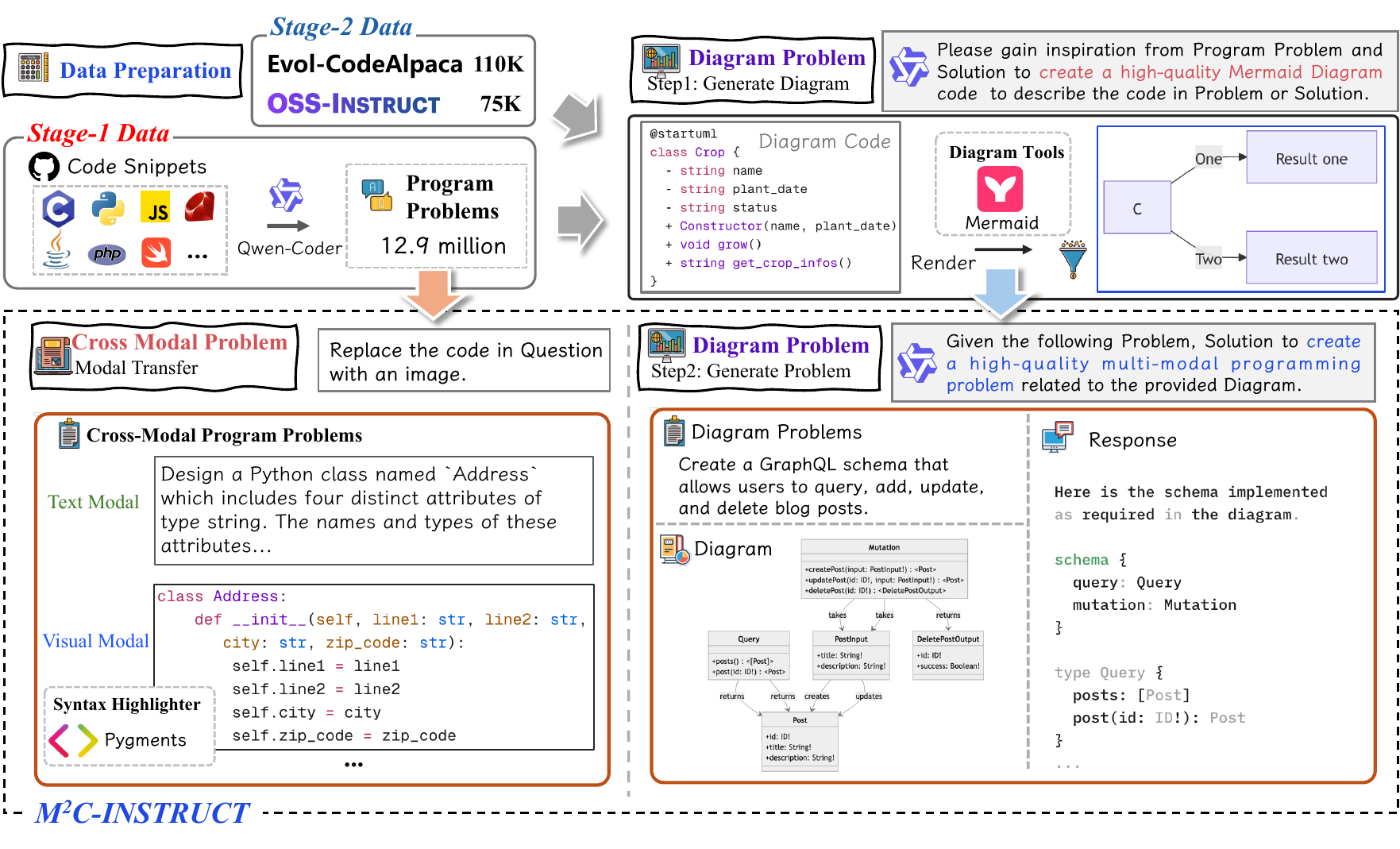}
    \caption{The \instruct{} data construction pipeline. Stage-1 data generates Cross-Modal Problems by converting code to images. Stage-2 data generates Diagram Problems via a two-step process: (1) creating diagrams from problems/solutions, and (2) formulating multimodal problems that necessitate these diagrams.}
    \label{fig:instruct_curation}
    \vspace{-15pt}
\end{figure*}

\textbf{Source Data Preparation.}\quad
As illustrated in \autoref{fig:instruct_curation}, \instruct{} comprises two stages of data preparation for \coder{} fine-tuning.
In the first stage, we collected a large-scale, multilingual code dataset from GitHub. Leveraging Qwen2.5-Coder, we generated 12.9 million question-answer pairs, which serve as the foundation for synthesizing multimodal fine-tuning data.
In the second stage, following the Magicoder~\citep{magicoder}, we employed two widely used datasets, Evol-CodeAlpaca~\citep{wizardcoder} and OSS-Instruct~\citep{magicoder}, to further synthesize multimodal diagram problems.

\textbf{Cross Modal Problem.}\quad 
The Cross-Modal problem refers to converting the code snippets within questions into the visual modality to enhance the code model's capabilities in visual code understanding and Optical Character Recognition (OCR). We utilize the code syntax highlighting tool \texttt{Pygments} to render the code within the questions.

\textbf{Diagram Problem.}\quad
The Diagram-related problem refers to generating corresponding code to solve a problem based on both the textual description and an associated diagram. We propose a two-step synthesis approach to achieve high-quality problem generation. In Step 1, we first prompt Qwen2.5-Coder to generate a diagram based on a standard code problem. We then attempt to render the diagram using Mermaid, filtering out cases where rendering is unsuccessful. In Step 2, we prompt Qwen2.5-Coder to create a multimodal problem based on the original problem and the diagram generated in Step 1, instructing it to ensure that key information necessary for solving the coding problem is preserved only within the diagram. This guarantees the necessity of the diagram during the problem-solving process.

\textbf{Stage-1 Data Synthesis.}\quad 
Based on the 12.9 million data samples prepared in the first stage, we split the dataset at a 9:1 ratio. Nine parts of the data are used to synthesize Cross-Modal problems, while one part is used to synthesize Diagram-related problems.

\textbf{Stage-2 Data Synthesis.}\quad
The data from the second stage is entirely used to synthesize Diagram-related problems. A total of 168K problems are synthesized. 

\subsection{\coder{} Training}
Based on our synthesized \instruct{} dataset, we propose a two-stage training framework to enhance the model's multimodal code generation capabilities.

\textbf{In the first stage}, the model is fine-tuned on extensive \instruct{}-stage-1 data. This dataset comprises a large volume of code-related images and textual information, a substantial amount of multilingual code, and a smaller subset of diagram-type data. The primary goal of this stage is to establish strong foundational visual understanding and information extraction capabilities, align code with critical multimodal information, and bolster the model's capacity for multilingual code comprehension and generation.

\textbf{In the second stage}, building upon the model from the first stage, we further fine-tune it using high-quality \instruct{}-stage-2 data. This stage focuses on enhancing the model’s image comprehension and instruction-following abilities, particularly for code-related tasks.



\section{Experiments}

\subsection{Experiment Setup}

\textbf{Evaluated Models.}\quad
In addition to evaluating our \coder{}, we evaluate 25+ widely used models, including both proprietary and open-source ones. For proprietary models, we assess GPT-4o~\citep{gpt4}, Claude3~\citep{claude3}, Gemini-2.5~\citep{gemini25}, etc. For open-source models, we test
Qwen-VL~\citep{qwen_vl2}, InternVL~\citep{internvl3}, Gemma3~\citep{gemma3}, Llama3/4~\citep{llama3, llama4}, and Phi-3~\citep{phi_3},etc.
Moreover, to validate the necessity of including diagrams in \benchmark{}, we also evaluated the text-only models Deepseek~\citep{deepseekv3} and Qwen-Coder~\citep{qwen25coder}, which are particularly strong in code-related tasks. (See full model list in \autoref{tab:model_list} in Appendix~\ref{supp:experiment}.)  

\textbf{\coder{} Training Setup.}\quad
\textit{Stage 1:} We use Qwen2-VL-7B-base to perform full-parameter fine-tuning on the \instruct{}-stage-1 dataset (max length 2048 tokens). Training with 1 epoch using AdamW (LR $5e-5$, batch size 1024), a cosine scheduler (0.1 warmup).
\textit{Stage 2:} Initializing from the Stage 1 checkpoint, we conduct fine-tuning only on the LLM (vision tower and projector frozen) using the \instruct{}-stage-2 dataset (max length 6000 tokens) for 2 epochs. Other hyperparameters remain identical to Stage 1. (Details can be found in the \autoref{supp:experiment}.)

\textbf{Evaluation Metrics.}\quad
Similar to text-only benchmarks such as HumanEval and MBPP, we adopt the \textbf{Pass@k} metric~\cite{chen2021pass_k}, which evaluates reliability based on execution results. In this study, we report the greedy \textbf{Pass@1} score for all ordinary LLMs using greedy decoding. For thinking type LLM (including Doubao-think, QvQ, and Kimi, etc), we use the corresponding official recommended inference temperature and sampling strategy to achieve the best performance.

\begin{table*}[h]
\small
\centering

\caption{Pass@1 (\%) scores of models for multimodal code generation task on \benchmark{}. In each model group, the best scores are in bold, and underlined numbers represent the second place.
``Avg.'' represents the average scores of all programming languages.
}
\label{tab:main_results}
\resizebox{\textwidth}{!}{\begin{tabular}{l|c|cccccccccc|c}
\toprule
\multirow{2.4}{*}{\bf Model} & \multirow{2.4}{*}{\bf Size} & \multicolumn{10}{c|}{\bf Program Languages} & \multirow{2.4}{*}{\bf Avg.}\\ 
\cmidrule{3-12} 
&  & C\# & CPP & Java & JS & Kotlin & PHP & Python & Ruby & Scala & Swift &  \\
\midrule
\multicolumn{12}{c}{\textcolor{red}{\textit{\textbf{LLMs w/o Diagram}}}} \\ \midrule
DeepSeek-V3 & 671B & 0.0 &  0.0 &  0.0 &  0.0 &  0.0 &  0.0 &  0.0 &  0.0 &  0.0 &  0.0 &  0.0 \\ 
Qwen2.5-Coder & 32B & 0.0 &  0.0 &  0.0 &  0.0 &  0.0 &  0.0 &  0.0 &  0.0 &  0.0 &  0.0 &  0.0 \\ 

\midrule
\multicolumn{12}{c}{\textit{\textbf{Proprietary LMMs}}} \\ \midrule

GPT-4o & \faLock{} & 40.0 & \bf 46.7 & \bf 56.7 & \underline{50.0} & \bf 56.7 & 46.7 & \underline{60.0} & \bf 56.7 & \underline{40.0} & 43.3 & \bf 49.7 \\ 
        Gemini-2.5-Flash-preview & \faLock{} & \bf 56.7 & \underline{36.7} & \underline{53.3} & \bf 53.3 & 43.3 & \bf 63.3 & \underline{60.0} & \underline{46.7} & 30.0 & 43.3 & \underline{48.7} \\
        Doubao1.5-thinking-pro& \faLock{} & 40.0 & 33.3 & 43.3 & 46.7 & 43.3 & 50.0 & \bf 66.7 & 43.3 & \underline{40.0} & 36.7 & 44.3 \\ 
        Claude-3.7-Sonnet & \faLock{} & 30.0 & 30.0 & 43.3 & 40.0 & \underline{50.0} & 50.0 & 43.3 & \underline{46.7} & \bf 43.3 & \underline{46.7} & 42.3 \\ 
        Claude-3.5-Sonnet& \faLock{} & \underline{46.7} & 26.7 & 40.0 & 40.0 & 26.7 & \underline{53.3} & 53.3 & \underline{46.7} & \underline{40.0} & \bf 50.0 & 42.3 \\ 
        Doubao1.5-vision-pro& \faLock{} & 33.3 & \bf 46.7 & 46.7 & 40.0 & 46.7 & 50.0 & 50.0 & 36.7 & \underline{40.0} & 26.7 & 41.7 \\ 
        GPT-4o-mini& \faLock{} & 40.0 & 26.7 & 46.7 & \underline{50.0} & 33.3 & 43.3 & 50.0 & 40.0 & 36.7 & 33.3 & 40.0 \\ 
        Qwen-VL-Max& \faLock{} & 13.3 & 23.3 & 26.7 & 43.3 & 33.3 & 33.3 & 40.0 & 36.7 & 16.7 & 23.3 & 29.0 \\ 
        Claude-3-Opus& \faLock{} & 0.0 & 10.0 & 10.0 & 20.0 & 13.3 & 23.3 & 16.7 & 20.0 & 10.0 & 10.0 & 13.3 \\
 
\midrule
\multicolumn{12}{c}{\textit{\textbf{70B+~~Open-Weight LMMs}}} \\
\midrule
        
Llama-4-Maverick & 400B & \bf{43.3} & \bf{33.3} & \bf{50.0} & \bf{46.7} & \underline{20.0} & \bf{50.0} & \bf{56.7} & \bf{50.0} & \bf{16.7} & \bf{40.0} & \bf{40.7} \\ 
Llama-4-Scout & 109B & \underline{26.7} & \underline{26.7} & \bf{50.0} & 30.0 & \bf{23.3} & \underline{36.7} & \underline{53.3} & \underline{30.0} & \bf{16.7} & 16.7 & \underline{31.0} \\ 
Qwen2-VL-Instruct & 72B & 16.7 & \underline{26.7} & \underline{23.3} & \underline{43.3} & \underline{20.0} & \underline{36.7} & 50.0 & \underline{30.0} & \underline{13.3} & \underline{26.7} & 28.7 \\ 
QVQ-Preview & 72B & 0.0 & 0.0 & 3.3 & 10.0 & 6.7 & 23.3 & 20.0 & 10.0 & 10.0 & 16.7 & 10.0 \\ 
Qwen2.5-VL-Instruct & 72B & 16.7 & \underline{26.7} & \underline{23.3} & 26.7 & \bf{23.3} & 33.3 & 30.0 & \underline{30.0} & \bf{16.7} & 20.0 & 24.7 \\ 
InternVL2-Llama3 & 76B & 13.3 & 13.3 & \underline{23.3} & 23.3 & \underline{20.0} & 26.7 & 23.3 & 23.3 & 10.0 & 16.7 & 19.3 \\ 
Llama3.2-vision & 90B  & 0.0 & 3.3 & 0.0 & 3.3 & 3.3 & 3.3 & 3.3 & 3.3 & 3.3 & 3.3 & 2.7 \\ 

\midrule
\multicolumn{12}{c}{\textit{\textbf{16B~-~32B~~Open-Weight LMMs}}} \\
\midrule

Gemma-3-it & 27B & \bf{23.3} & \bf{20.0} & \bf{20.0} & 16.7 & \bf{20.0} & \bf{30.0} & \bf{33.3} & \underline{13.3} & \underline{6.7} & \bf{20.0} & \bf{20.3} \\
Qwen2.5-VL-Instruct & 32B & 10.0 & \underline{16.7} & \bf{20.0} & \bf{26.7} & \underline{13.3} & \underline{20.0} & \underline{26.7} & \bf{23.3} & \bf{16.7} & \underline{13.3} & \underline{18.7} \\
Kimi-VL-Thinking & 16B & 0.0 & 13.3 & \underline{13.3} & \underline{20.0} & 6.7 & \bf{30.0} & \underline{26.7} & \bf{23.3} & \underline{6.7} & 3.3 & 14.3 \\
InternVL2 & 26B & \underline{13.3} & 3.3 & 6.7 & \underline{20.0} & 3.3 & 10.0 & 10.0 & 10.0 & 0.0 & 6.7 & 8.3 \\
DeepSeek-VL2 & 27B & 6.7 & 0.0 & 3.3 & 10.0 & 3.3 & 6.7 & 10.0 & 6.7 & 3.3 & 3.3 & 5.3 \\

\midrule
\multicolumn{12}{c}{\textit{\textbf{4B~-~8B~~Open-Weight LMMs}}} \\
\midrule
InternVL3 & 8B & 13.3 & \bf 20.0 & 3.3 & 20.0 & \underline{13.3} & \underline{16.7} & \underline{26.7} & \underline{26.7} & 3.3 & \underline{10.0} & \underline{15.3} \\
Qwen2-VL-Instruct & 7B & \underline{16.7} & 3.3 & \underline{10.0} & \underline{23.3} & \underline{13.3} & 10.0 & 20.0 & 20.0 & 3.3 & 0.0 & 12.0 \\
Qwen2.5-VL-Instruct & 7B & 6.7 & \underline{6.7} & 3.3 & 6.7 & 3.3 & 6.7 & 10.0 & 3.3 & \underline{6.7} & 0.0 & 5.3 \\
Phi3-vision & 4B  & 0.0 & 3.3 & 3.3 & 6.7 & 0.0 & 13.3 & 13.3 & 3.3 & 0.0 & 0.0 & 4.3 \\
MiniCPM-V-2.6 & 8B & 0.0 & \underline{6.7} & 3.3 & 6.7 & 3.3 & 3.3 & 3.3 & 3.3 & 0.0 & 3.3 & 3.3 \\
\rowcolor[HTML]{E8E8FF}
\textbf{\coder{}~(Ours)} & 7B & \bf 26.7 & \bf 20.0 & \bf 20.0 & \bf 30.0 & \bf 16.7 & \bf 36.7 & \bf 36.7 & \bf 33.3 & \bf 16.7 & \bf 16.7 & \bf 25.3 \\

\bottomrule
\end{tabular}}
\end{table*}

\subsection{Main Results}
\autoref{tab:main_results} shows the performance of models on \benchmark{} for multilingual multimodal code generation task. The key findings are summarized as follows:

\textbf{Multilingual multimodal code generation is challenging.}
Even for the strongest model, which also only reached about 50\% in the evaluation. This indicates that though most models are trained with text-rich images, the strong text recognition abilities do not guarantee high performance on \benchmark{}. 
Because \benchmark{} requires comprehensive OCR, visual logic understanding, and code generation capabilities. In addition, most models perform better in Python, JS, etc., but poorly in C\#, Scala, etc. This highlights the great potential for innovation in models, data, or training objectives to improve the performance of LMMs in multilingual multimodal programming tasks.

\textbf{There is still a gap between open and proprietary LMMs.}
The results show a clear performance gap between proprietary and open-source models across most programming languages, where models like GPT-4o, Gemini-2.5, Claude-3, and Doubao-1.5 lead the benchmark.

\textbf{LLM could not solve problems without diagrams.}
We also test the models without diagrams. The results show that DeepSeek-V3 and Qwen2.5-Coder cannot solve the problems without diagrams. This proves that diagrams are necessary in \benchmark{}(we paid special attention to this during annotation).

\textbf{\coder{} outperforms similarly sized models.}
It is noteworthy that \coder{}-7B outperforms other large language models (LLMs) of similar or even larger sizes (like Qwen2.5-VL-72B). 
This also validates the effectiveness of the \instruct{} data we have constructed, which can significantly improve the performance of models in multimodal programming tasks.

\begin{wrapfigure}{r}{0.38\textwidth}
\vspace{-12mm}
    \centering
    \includegraphics[width=0.38\textwidth]{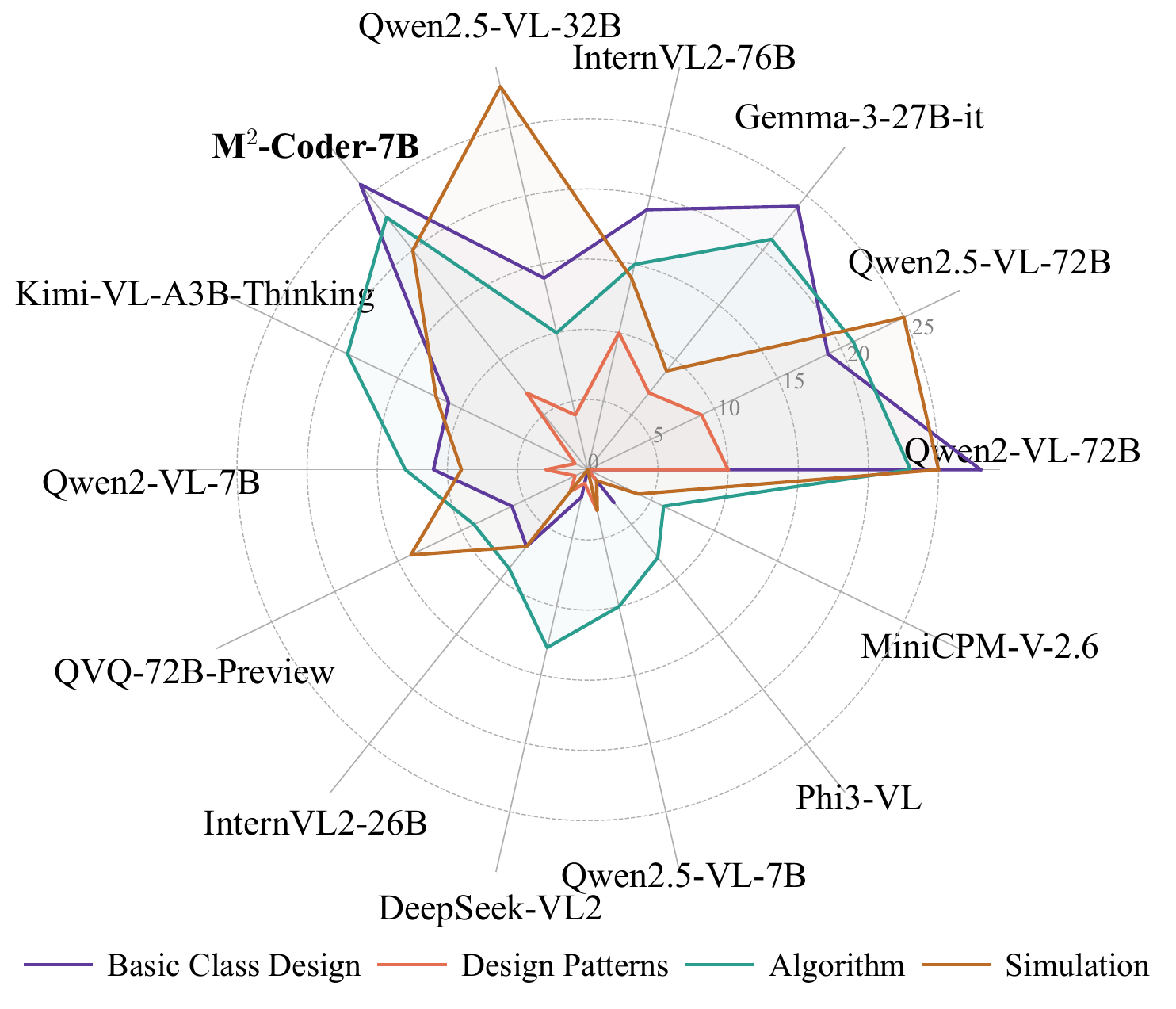}
    \caption{Performance comparison of models across task types.}
    \label{fig:radar_by_task}
    \vspace{-4mm}
\end{wrapfigure}

\subsection{Further Analysis}

To further dissect the experimental results, we conduct additional analyses examining performance from multiple perspectives. 

\textbf{Performance across different tasks}\quad
We have classified the problems in \benchmark{} into four categories based on their core knowledge points: Class Design, Design Patterns, Algorithms, and Simulation. (Details of this classification can be found in \autoref{tab:bench_data_category} in Appendix~\ref{supp:bench}.) Our model demonstrably outperforms other models of comparable size across all these task categories. Tasks involving Design Patterns are particularly challenging, as they require models to understand the design patterns in diagrams. Consequently, all models exhibited relatively poor performance on such tasks.
%



\begin{wrapfigure}{r}{0.38\textwidth}
\vspace{-5mm}
    \centering
    \includegraphics[width=0.38\textwidth]{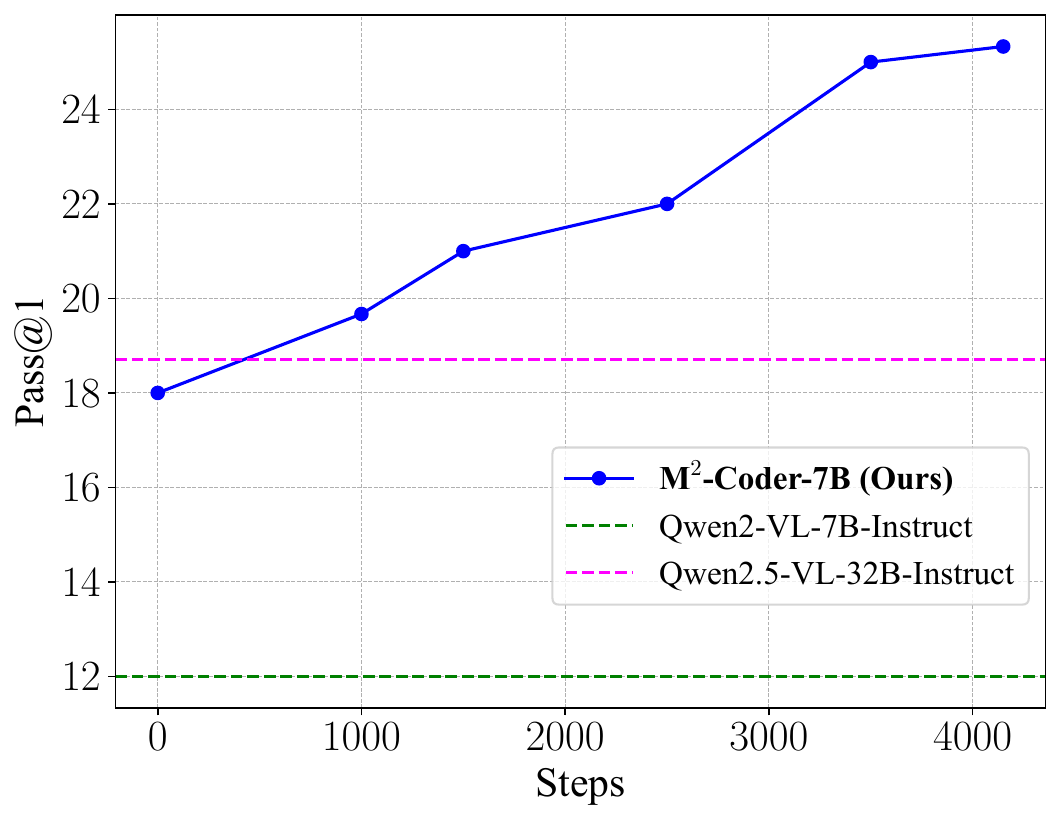}
    \vspace{-5mm}
    \caption{Ablation study on stage-1 data efficiency.}
    \label{fig:train_data_efficiency}
    \vspace{-5mm}
\end{wrapfigure}
\textbf{Ablation experiment for \coder{} fine-tuning.}\quad
We perform an ablation study on \coder{} and present the results in \autoref{tab:ablation} and \autoref{fig:train_data_efficiency}. Key findings include:

\vspace{-0.15cm}
\begin{itemize}[leftmargin=*]
    \item \textit{\textbf{Full Two-Stage SFT is Optimal:}} Fine-tuning Qwen2-VL-Base with both stage1 and stage2 SFT (Model \ding{178}) achieved the highest average Pass@1 score of 25.3.\vspace{-0.12cm}
    \item \textit{\textbf{Stage 2 SFT Efficacy:}} Applying stage2 SFT alone significantly improved performance. For instance, Qwen2-VL-Instruct (Model \ding{172}, Avg. 12.0) + stage2 SFT (Model \ding{173}) increased the average to 16.7. Similarly, Qwen2-VL-Base + stage2 SFT (Model \ding{175}) reached an average of 18.0.\vspace{-0.10cm}
    \item \textit{\textbf{Stage-1 SFT Contribution:}} Stage1 SFT on Qwen2-VL-Base (Model \ding{177}) yielded an average of 10.0. Fine-tuning the vision encoder during stage 1 proved beneficial, as freezing it (Model \ding{176}) resulted in a lower score of 8.7.\vspace{-0.10cm}
\end{itemize}

\begin{itemize}[leftmargin=*]
    \item \textit{\textbf{Combined Strength:}} The results underscore the effectiveness of both SFT stages, with the complete two-stage approach on the base model (Model \ding{178}) outperforming partial configurations.\vspace{-0.10cm}
    \item \textit{\textbf{Stage-1 Data Efficiency:}} 
     We select model weights (Model \ding{177})  preserved at different steps during the stage-1 SFT (1 epoch in total). These checkpoints correspond to varying amounts of stage-1 training data. These selected weights are then used for stage-2 training. It can be observed in \autoref{fig:train_data_efficiency} that the more data the model is trained on during stage-1, the better its performance after the completion of stage-2 training. This again demonstrates the effectiveness of our stage-1 training.
    \vspace{-0.10cm}
\end{itemize}

\begin{table*}[ht]
\centering
\vspace{-5mm}
\caption{Ablation study of our two-stage SFT dataset, \instruct{}. Results show Pass@1 on \benchmark{} for Qwen2-VL-7B fine-tuned with different parts of \instruct{}.}

\label{tab:ablation}

\resizebox{0.95\textwidth}{!}{\begin{tabular}{l|l|cccccccccc|c}
\toprule
\multirow{2.4}{*}{\bf ID} & \multirow{2.4}{*}{\bf Model} & \multicolumn{10}{c|}{\bf Program Languages} & \multirow{2.4}{*}{\bf Avg.}\\ 
\cmidrule{3-12} 
& & C\# & CPP & Java & JS & Kotlin & PHP & Python & Ruby & Scala & Swift &  \\
\midrule
\large{{\ding{172}}} & Qwen2-VL-Instruct  & 16.7 & 3.3 & 10.0 & 23.3 & 13.3 & 10.0 & 20.0 & 20.0 & 3.3 & 0.0 & 12.0 \\ 
\large{{\ding{173}}} &\large{{\ding{172}}}~+~stage2 SFT & 10.0 & 13.3 & 10.0 & 20.0 & 10.0 & 23.3 & 23.3 & 23.3 & \bf 20.0 & 13.3 & 16.7 \\
\midrule
\large{{\ding{174}}}& Qwen2-VL-Base & - & - & - & - & - & - & - & - & - & - & - \\ 
\large{{\ding{175}}} &\large{{\ding{174}}}~+~stage2 SFT & 13.3 & 10.0 & 13.3 & \bf 30.0 & \bf 16.7 & 26.7 & 20.0 & 20.0 & 13.3 & \bf 16.7 & 18.0 \\ 
\large{{\ding{176}}} &\large{{\ding{174}}}~+~stage1 SFT (freeze) & 10.0 & 6.7 & 13.3 & 16.7 & 3.3 & 3.3 & 16.7 & 13.3 & 3.3 & 0.0 & 8.7 \\ 
\large{{\ding{177}}} &\large{{\ding{174}}}~+~stage1 SFT & 20.0 & 3.3 & 13.3 & 23.3 & 3.3 & 6.7 & 16.7 & 10.0 & 3.3 & 0.0 & 10.0 \\ 
\large{{\ding{178}}} &\large{{\ding{174}}}~+~stage1 \& stage2 SFT & \bf 26.7 & \bf 20.0 & \bf 20.0 & \bf 30.0 & \bf 16.7 & \bf 36.7 & \bf 36.7 & \bf 33.3 & 16.7 & \bf 16.7 & \textbf{25.3} \\

\bottomrule
\end{tabular}}
\end{table*}



\begin{wrapfigure}{r}{0.3\textwidth}
\vspace{-8mm}
    \centering
    \includegraphics[width=0.3\textwidth]{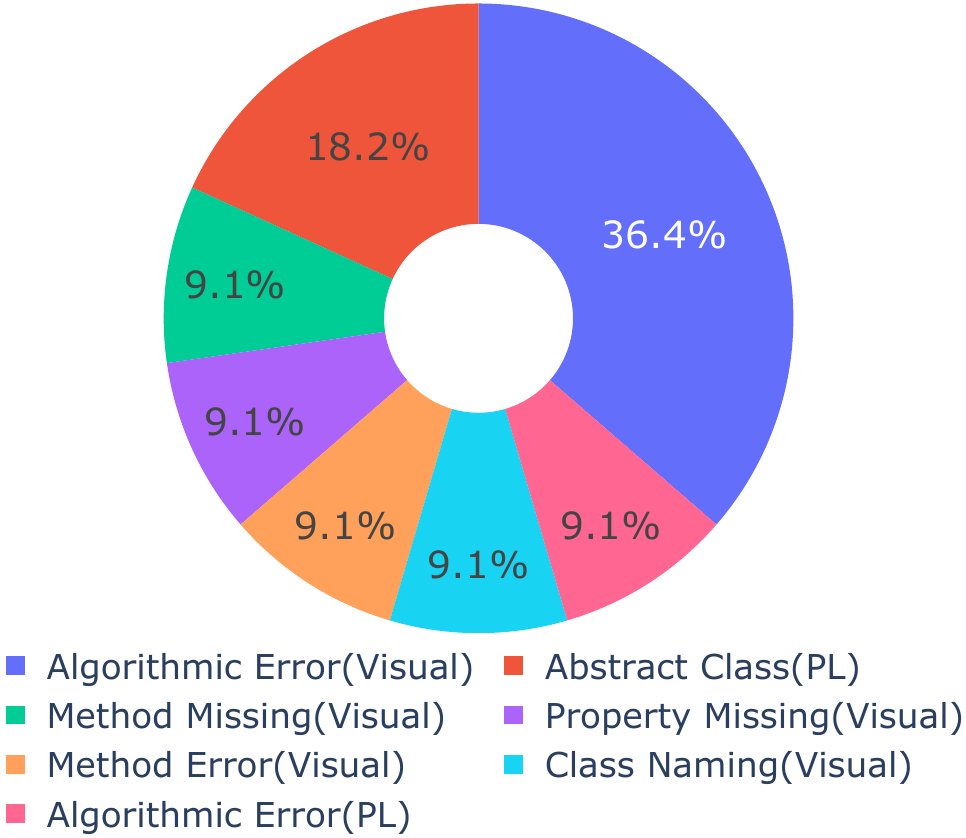}
    \vspace{-5mm}
    \caption{Distribution of GPT-4o's Errors in Python.}
    \label{fig:error_analysis}
    \vspace{-8mm}
\end{wrapfigure}

\textbf{Error Analysis.}\quad
\label{sec:error_analysis}
To gain deeper insights into model capabilities, we conducted an error analysis on GPT-4o's responses to 11 problems in Python. As shown in \autoref{fig:error_analysis}, our results underscore the limitations of current models in multimodal code generation, as revealed by our benchmark. Errors in Python are frequently linked to deficient visual reasoning, demonstrating an inability to fully leverage visual cues for accurate code generation. 
We further observe that these types of errors are more pronounced in smaller models and when generating code for less common programming languages. 
For more error case analysis, please refer to Appendix~\ref{supp:error_analysis}.

\begin{figure*}[h]
\begin{center}
    \includegraphics[width=1.0\textwidth]{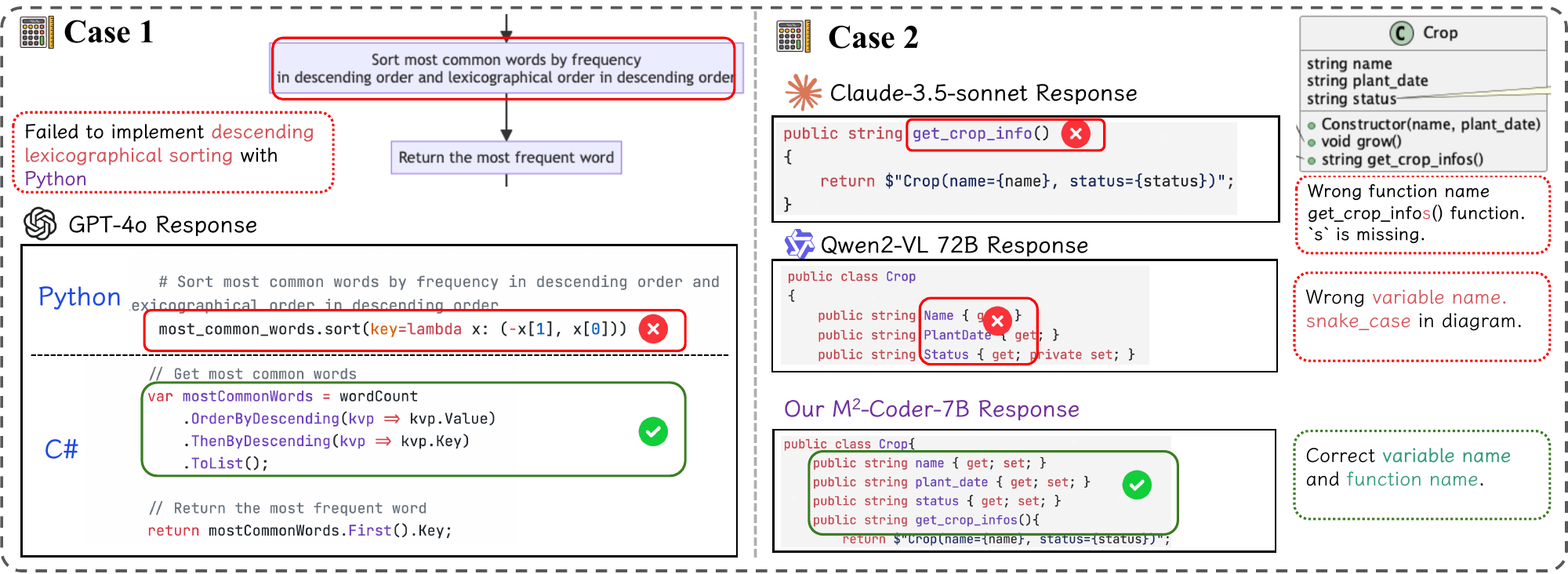}
    \caption{A case study for \benchmark{} showcasing common error patterns in multimodal code generation. It highlights failures on specific challenges (complex sorting, diagram-to-code with naming conventions).}
    \label{fig:case_study}
    \vspace{-10pt}
\end{center}
\end{figure*}

\vspace{0.35cm}
\textbf{Case Study.}\quad
Furthermore, we illustrate several error cases in \autoref{fig:case_study}. In Case 1, when tasked with the same sorting problem, GPT-4o is unable to produce a solution when responding in Python, yet it succeeded when using C\#. In Case 2, concerning class implementation, GPT-4o's Java response is missing a function. Qwen-VL does not follow the textual instructions regarding variable visibility requirements, and the code generated by Intern-VL exhibits attribute and function naming inconsistent with the provided diagram. Multimodal code generation tasks demand that models concurrently process both textual and visual information, which requires models to have not only strong visual comprehension skills but also robust code generation capabilities.

\begin{figure}[htb]
	\centering
     \subfigure[Scores on each PLs]
    {\includegraphics[width=0.45\textwidth]{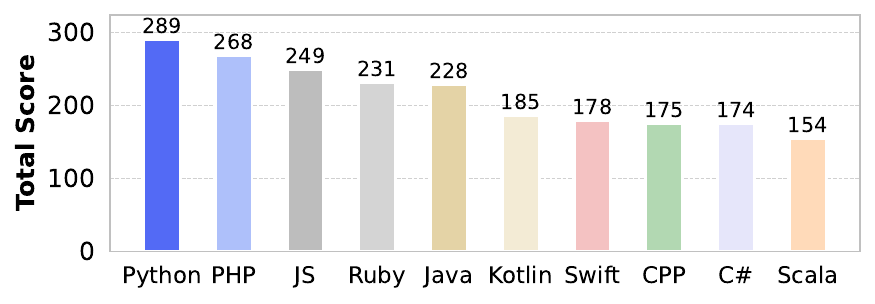}}
    \hspace{0.03\textwidth} 
    \subfigure[Scores on each problem in each languages]{\includegraphics[width=0.5\textwidth]{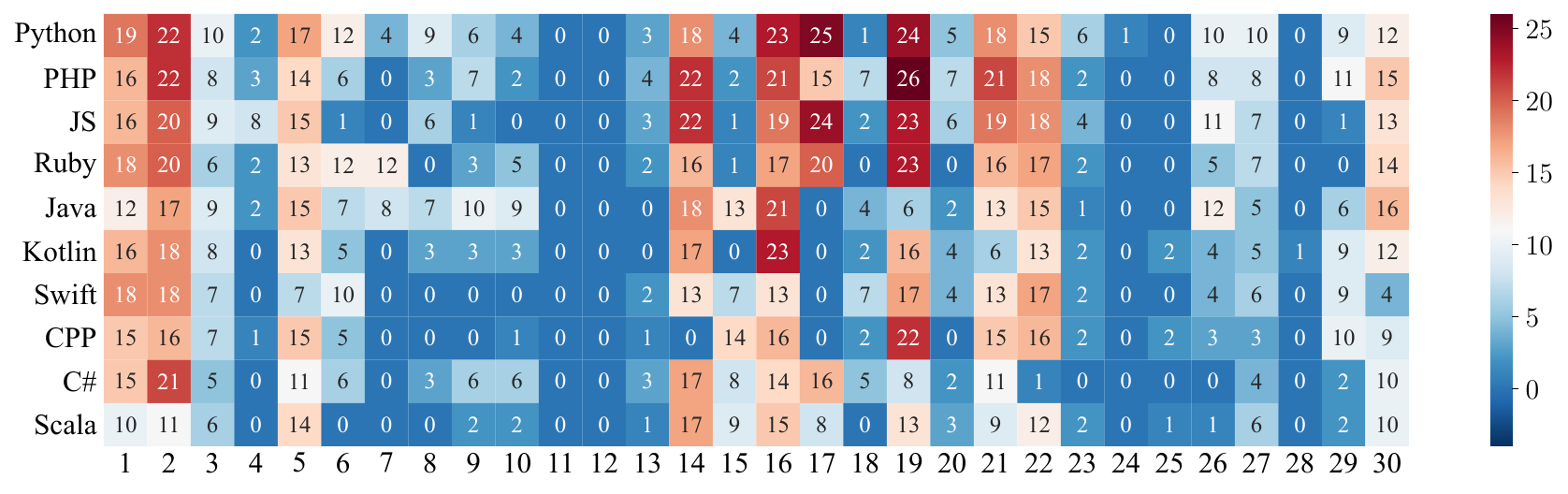}}
    
    \vspace{-5pt}
	\caption{Performance differences between each PLs on \benchmark{}.}
	\label{fig:pl_performance_stat}
    \vspace{-10pt}
\end{figure}

\textbf{Programming Language Matters.}\quad
Extending the analysis of language-specific performance in~\textit{Error Analysis} section, we present a broader analysis. \autoref{fig:pl_performance_stat}~(a) shows the aggregated correct answers from all evaluated models, grouped by programming language. Scripting languages such as Python, PHP, and JS demonstrated superior performance, contrasting with weaker results from languages like Swift, C\#, and Scala. This suggests, firstly, an imbalance in LLM programming proficiency, likely due to benchmark-driven optimizations favoring Python. Secondly, models struggle with languages enforcing strict type checking and rules (e.g., variable and method visibility), which demand more precise instructions and diagram following.
\autoref{fig:pl_performance_stat}~(b) details the correct solutions per problem across different languages. On simpler tasks (e.g., problem IDs 1, 2, 14, 16), inter-language performance variance is minimal, with most languages achieving good results. Conversely, for medium to difficult problems (e.g., problem IDs 8, 9, 26, 27), these performance disparities are significantly more pronounced.

\section{Related Work}
\textbf{Code Large Language Models and Evaluation.}\quad The rapid progress of large language models (LLMs)\citep{gpt4,claude3,llama3,Qwen} has enabled significant advances in code-related tasks. Early models like BERT\citep{bert} and GPT~\citep{gpt} were adapted for code understanding and generation~\citep{chen2021pass_k, code_bert, bloom}. More recent Code LLMs benefit from domain-specific pre-training and instruction tuning on large code corpora~\citep{code_llama, wizardcoder, deepseek_coder, magicoder, starcoder2, qwen25coder}, achieving strong performance in tasks like code completion and synthesis.
To evaluate these models, diverse benchmarks have emerged, ranging from basic coding tasks~\citep{chen2021pass_k, mbpp, bigcodebench} to more complex settings such as code repair~\citep{quixbugs, debugbench, EvalGPTFix}, multilingual code~\citep{multipl_e, mbxp, mceval}, repository-level assessments~\citep{repobench, r2c2coder, codeplan}, and agent-based evaluations~\citep{swe_bench, pybench}.

\textbf{Visual Reasoning and Code Generation.}\quad Large Multimodal Models(LMMs)\cite{minigpt4,llava,Qwen_VL,mplug,llava_next,internlm_xcomposer,llavaonevision,qwen_vl2, zong2024mova} incorporate visual information into LLMs through visual encoders\citep{clip}, extending the capabilities of LLMs to address visual tasks. 
Prior studies, such as VQA\citep{vqa,vinvqa,mtvqa}, evaluated the basic visual semantic capabilities of models. 
With the emergence of increasingly powerful visual and semantic capabilities in LMMs and LLMs, many recent works have shifted focus toward more complex multi-modal tasks, such as  mathematical reasoning\citep{trinh2024solvingolympiad, shao2024visual, huang2024olympicarena, mavis}, 
chart understanding\citep{chartllama,flowchartqa,flowvqa,wang2024charxiv,li-towards-real,multimodalselfinstruct}, code generation\citep{design2code,li2024mmcode,logomotion,wu2024plot2code,shi2024chartmimic,robocodex}, and agent-driven interactions\citep{webarena,osworld,visualagentbench}. 
Recent multimodal code works focused on visual algorithmic problems\citep{li2024mmcode, codevision}, chart code generation\cite{wu2024plot2code,shi2024chartmimic}, and UI design\citep{design2code, logomotion}. Our \benchmark{} explores the task of code generation based on code diagrams. Compared with previous work\citep{liu2022flow2code} that converts flowcharts into node information for complex processing, we focus on more practical scenarios that directly perform semantic understanding based on images. 

\section{Conclusion}
\label{sec:conclusion}

In this work, we introduce \coder{}, a novel multilingual multimodal software developer integrating visual design inputs with textual instructions to improve code generation accuracy and architectural alignment.
To facilitate its development and evaluation, we create \instruct{}, a large-scale instruction-tuning dataset \textit{enhanced with rendered cross-modal data for improved code visual understanding and OCR}, and \benchmark{}, a new comprehensive benchmark for multimodal code generation.
Our experiments demonstrate \coder{}'s promising performance, comparable to larger models, and underscore \textit{the effectiveness of \instruct{} in boosting task performance}.
However, significant challenges remain: models still struggle with \textit{precise visual information utilization, robust instruction adherence, and mastery of advanced programming concepts}.
Despite these complexities, by enabling LLMs to process multimodal specifications, this research is a significant step towards more effective AI-assisted software automation and future advancements in multimodal AI for software engineering.



\bibliography{neurips}
\bibliographystyle{plain}
\input{appendix}



\end{document}

%% file: appendix.tex
\clearpage
\appendix
\section*{Appendix}
\startcontents[sections]
\printcontents[sections]{l}{1}{\setcounter{tocdepth}{3}}
\clearpage

\input{showcase_format} 

\section{Limitations}
\label{supp:limitation}
Our work has the following limitations:
\begin{itemize}[leftmargin=*]
\item \textbf{Scope of Tasks:} Our method was implemented and validated exclusively on multimodal code generation tasks. Future work could extend its application to a broader range of code-related tasks, such as multimodal code understanding, code completion, and code debugging. 
\item \textbf{Benchmark:} Due to constraints related to team size and the inherent complexity of annotating multimodal code data, our constructed benchmark currently comprises 10 programming languages. Future efforts could focus on expanding this benchmark to include a significantly larger volume of data and a more diverse set of scenarios. 
\item \textbf{Training Methodology:} We employed instruction fine-tuning for model training. Future research could explore alternative or complementary training paradigms, such as reinforcement learning or techniques like "long thinking," to potentially enhance model performance or capabilities.
\end{itemize}

\section{Potential societal impacts}
\label{supp:impacts}
The introduction of \coder{} and its associated ecosystem (\instruct{}, \benchmark{}) presents a significant advancement in AI-assisted software engineering, with several potential societal impacts, both positive and negative.
\subsection{Positive impacts}
The development of \coder{} promises to enhance software developer productivity and improve code quality by ensuring better alignment with architectural intent through multimodal input. Its multilingual capabilities and the ability to interpret visual designs could broaden participation in software development, making it more accessible globally and to individuals with diverse learning preferences. Furthermore, the creation of resources like \instruct{} and \benchmark{} will likely stimulate further research and innovation in multimodal AI for software engineering. By demonstrating strong performance, even compared to larger models, \coder{} could democratize access to advanced AI-driven development tools, fostering innovation across the industry and accelerating the development of more sophisticated software systems. This work represents a tangible step towards AI systems that can more effectively assist in and automate complex industrial programming tasks.

\subsection{Negative impacts}
While promising, the broader adoption of advanced AI tools like \coder{} in software engineering necessitates careful consideration of potential societal adjustments. The evolution of software development roles may occur as automation handles more routine coding tasks, emphasizing the need for ongoing skill development. Furthermore, as with all AI systems reliant on large datasets, continued vigilance will be important to ensure responsible development and deployment, addressing general concerns common to AI applications regarding fairness and robust performance in diverse scenarios. The long-term implications of widespread AI-assisted code generation on the software ecosystem will also require ongoing observation and discussion.

\clearpage

\section{Details of \benchmark{}}
\label{supp:bench}
In this section, we provide a detailed presentation of the \benchmark{} statistics, construction process, annotation details, and annotator payment.

\subsection{Detailed Statistics}

In \autoref{tab:bench_data_category}, we list the task categories, the programming knowledge points/concepts tested, the difficulty level, and the corresponding number of test cases for the 30 problems in the dataset (which have parallel versions across 10 programming languages).

\begin{table}[htbp]
\centering
\caption{Overview of Test Categories, Difficulties, and Case Counts}
\label{tab:bench_data_category}
\resizebox{\textwidth}{!}{\begin{tabular}{l|lp{7cm}lc} 
\toprule
No. & Task & Category & Difficulty & Number of Test Cases \\
\midrule
1  & Basic Class Design          & Class Design                                                                                                & Easy   & 4  \\
2  & Basic Class Design          & Class Design                                                                                                & Easy   & 4  \\
3  & Basic Class Design          & Class Design                                                                                                & Medium & 11 \\
4  & Basic Class Design          & Class Design                                                                                                & Hard   & 16 \\
5  & Abstraction and Inheritance & Abstract, Inheritance                                                                                       & Easy   & 8  \\
6  & Abstraction and Inheritance & Interface, Inheritance                                                                                      & Medium & 10 \\
7  & Creational Patterns         & Inheritance, Singleton Pattern                                                                              & Hard   & 8  \\
8  & Creational Patterns         & Abstract, Singleton, Factory Pattern                                                                        & Hard   & 8  \\
9  & Creational Patterns         & Abstract, Singleton, Builder Pattern                                                                        & Hard   & 9  \\
10 & Structural Pattern          & Abstract, Singleton, Adapter Pattern                                                                        & Hard   & 9  \\
11 & Behavioral Pattern          & Abstract, Strategy Pattern, Composite Pattern, Template Method Pattern                                        & Hard   & 12 \\
12 & Creational Patterns         & Abstract, Strategy Pattern, Composite Pattern, Template Method Pattern, Abstract Factory Pattern            & Hard   & 14 \\
13 & Behavioral Pattern          & Abstract, Strategy Pattern, Composite Pattern, Chain of Responsibility Pattern                                & Hard   & 11 \\
14 & Structural Pattern          & Abstract, Proxy Pattern                                                                                     & Easy   & 5  \\
15 & Behavioral Pattern          & Abstract, Observer Pattern                                                                                  & Medium & 6  \\
16 & Algorithm                   & Size Comparison                                                                                             & Easy   & 10 \\
17 & Algorithm                   & Binary Search                                                                                               & Easy   & 9  \\
18 & Algorithm                   & Complex Sorting                                                                                             & Hard   & 8  \\
19 & Algorithm                   & Sorting                                                                                                     & Easy   & 10 \\
20 & Algorithm                   & Sorting                                                                                                     & Hard   & 7  \\
21 & Simulation                  & Simulation                                                                                                  & Easy   & 6  \\
22 & Simulation                  & Simulation                                                                                                  & Easy   & 8  \\
23 & Algorithm                   & Dynamic Programming, Graph                                                                                  & Hard   & 10 \\
24 & Algorithm                   & Dynamic Programming                                                                                         & Hard   & 12 \\
25 & Simulation                  & Simulation, Sorting, Hashing, Collection                                                                    & Hard   & 6  \\
26 & Simulation                  & Simulation, Heap, Greedy                                                                                    & Medium & 13 \\
27 & Simulation                  & Data Aggregation                                                                                            & Medium & 2  \\
28 & Simulation                  & Simulation, Hashing, String Processing                                                                      & Hard   & 9  \\
29 & Simulation                  & Rule Simulation                                                                                             & Medium & 6  \\
30 & Simulation                  & Sorting, Recommendation                                                                                     & Medium & 19 \\
\bottomrule
\end{tabular}}
\end{table}

\subsection{\benchmark{} Curation Process}
\textbf{Prototype Problem Design.}\quad
To comprehensively evaluate the multimodal coding capabilities of LMMs within a limited set of problems, we prepared the problem design from three dimensions: scenarios, basic programming knowledge, and advanced programming knowledge. First, we designed corresponding scenarios for each problem. Based on these settings, we used an LLM to generate prototype problems in Python, each of which includes a problem prompt, the corresponding solution, and test cases. During this stage, we manually interact with the LLM in multiple rounds to ensure that the prototype questions align with the problem settings, that the provided solutions are accurate, and that the test cases cover the key evaluation points of each problem.

\textbf{Problem Design.}\quad
In the second step, based on the identified prototype problems, we further process the Prompt and Solution. Specifically, given the class or function associated with the Solution, we leverage an LLM to generate diagram code in PlantUML\footnote{https://pygments.org/}  or Mermaid\footnote{https://mermaid.js.org/}. We then manually refine and adjust the generated diagrams to ensure structural and semantic correctness. Subsequently, we revise the Prompt by removing redundant information that overlaps with the diagram content, such that the Prompt alone is insufficient for correct code synthesis. To ensure that the problem remains solvable when both the Prompt and the diagram are provided, we add necessary descriptions directly within the diagram. Through this process, we complete the annotation of Python problems, where each instance consists of a prompt, a diagram, a solution, and corresponding test cases.

\textbf{Problem Translate.}\quad
Finally, we extend the problems to nine other programming languages. For each problem, the prompt needs to be adapted accordingly (e.g., modifying function names), and both the solution and test cases must be fully translated. We recruited nine volunteers to complete this task. The volunteers conducted the translations with the assistance of LLMs, and during annotation, they were required to verify aspects such as variable consistency and input-output alignment to ensure the accuracy of the translations.

\subsection{Data Annotation}
In this section, we detail our data annotation guidelines, the specific division of labor for each step, and the specific remuneration information we pay to the annotators in accordance with relevant regulations.

\subsubsection{Data annotation guidelines}
Before data annotation, we first designed the following data annotation guidelines to ensure the quality of the data we annotated.
The guidelines primarily cover the following aspects:
\begin{itemize}[leftmargin=*]
\item \textbf{Accessibility:} The reference data we use for annotations comes from materials with permissive licenses. This means it can be freely used and shared for research purposes.
\item \textbf{Standardized Format:} We provide a sample annotation for 10 programming languages. Annotators are expected to follow this established format for all their annotation work.
\item \textbf{Difficulty Classification:} Annotators receive clear guidelines on how to categorize the difficulty level for each language. They must strictly apply these guidelines to label problems based on their algorithmic complexity and the features involved.
\item \textbf{Self-Contained Problems:} Annotators must thoroughly check their annotated problems. This ensures that the problem and visual inputs contain all the necessary information to be solved clearly. The example inputs and outputs must be accurate, the reference solutions should run correctly, and the test cases created must fully assess the correctness of the functions.
\end{itemize}

\subsubsection{Annotation of Prototype Problem and Python Problem.}
The annotation process was conducted collaboratively by two annotators. They jointly investigated and organized the scenarios and knowledge points relevant to each question, and completed the annotation of diagrams, solutions, and test cases with the assistance of LLM. To ensure data quality, mutual cross-checking was performed to verify that each question was answerable and that the intended questions required joint reasoning over both textual and visual inputs. Furthermore, a comprehensive set of test cases was constructed for evaluation purposes. For each question, the annotators also provided a reference solution to further validate the correctness of both the questions and the associated test cases.

\subsubsection{Annotation of Problem Translation.}
\paragraph{Employ Annotators.}
To translate the annotated Python problems into 9 other programming languages, we employed nine paid annotators for this task. These annotators were Master's or PhD candidates in Computer Science. During the recruitment process, we comprehensively assessed their programming knowledge and skills (e.g., object-oriented programming (OOP) concepts, code compilation, unit testing), as well as their proficiency in using Large Language Models (LLMs). 

\paragraph{Annotation Training.}
Each annotator was assigned a programming language with which they were familiar, ensuring they could successfully complete the annotation tasks for their respective language.
Furthermore, prior to commencing annotation, all annotators received training. This training detailed the specific annotation requirements and guidelines, and provided them with illustrative examples. 

\paragraph{Data Annotation Content.}
In the actual annotation process, with the assistance of LLMs, they were required to translate the functions and parameter descriptions from the original problems into syntactically valid forms for the target language. They also translated the solutions and test cases into the corresponding language. Critically, to ensure the quality of the annotations, the annotators were required to verify that the solutions they translated could successfully pass the test cases they also translated.

\subsubsection{Quality Control}
Subsequent to the completion of their initial annotation tasks, the annotators performed a final data verification phase to ensure the integrity of the data.
To facilitate this, each annotator was allocated annotation data produced by two of their peers. Annotators were responsible for examining the correctness of the problems, which involved executing the associated solutions and test cases to validate them, and subsequently correcting any discovered inaccuracies. In instances where errors in the data were contentious, discussions were first initiated with the original annotator. Should an agreement not be reached, a third annotator was introduced to facilitate a joint resolution. The root causes of such errors were then disseminated to the entire annotation team to alert them to analogous potential problems.
Ultimately, the canonical solution (labeled by the annotator) was required to pass all corresponding test cases, thereby ensuring the correctness of every problem created (with a 100\% pass rate). This meticulous validation process guaranteed the development of a high-quality, multilingual programming benchmark that supports comprehensive analysis and comprehension of code examples across a variety of programming languages.

\subsubsection{Annotation Payments}
\label{supp:annotate_pay}
Annotators were compensated at about \$3 per problem, with all work performed online. We also provided the necessary LLM API for the annotation task. In total, approximately 300 problems were annotated. The entire process, including quality assurance for problems~(two problems per annotators), incurred a total cost of approximately \$2,700.

\clearpage
\section{Details of \instruct{}}
\label{supp:instruct}
In this section, we provide a detailed presentation of the \instruct{} statistics, construction process, and quality control.
\vspace{-2mm}
\subsection{Detailed Statistics}

\begin{figure}[htb]
	\centering

     \subfigure[cross-model subset]
    {\includegraphics[width=0.45\textwidth]{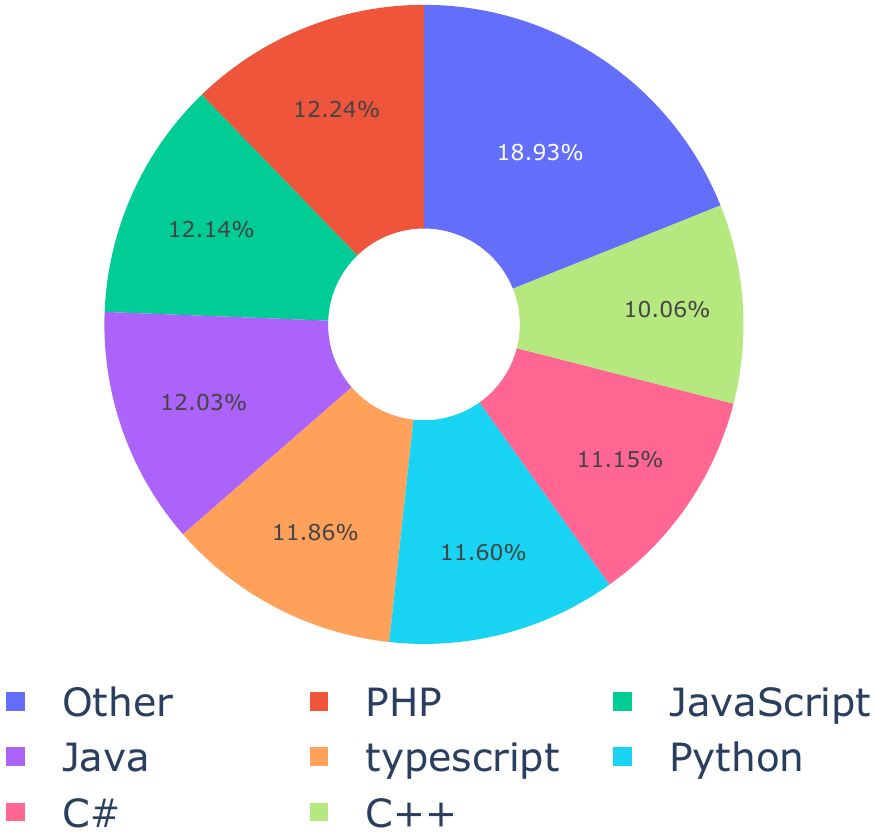}
    }
    \hspace{0.03\textwidth} 
    \subfigure[diagram subset]{\includegraphics[width=0.45\textwidth]{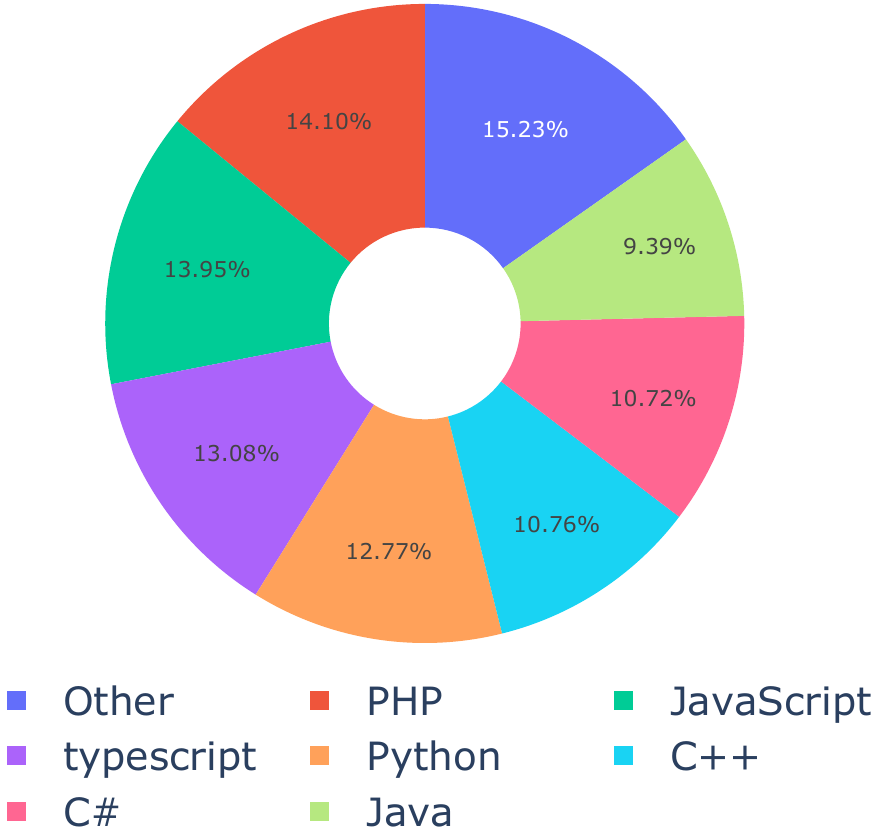}}
    
    \vspace{-5pt}
	\caption{Programming language distribution in \instruct{} stage 1 data}
	\label{fig:inst_stage1_dist_stat}
    \vspace{-10pt}
\end{figure}

\noindent \textbf{\instruct{} Stage 1.}\quad
\autoref{tab:stg1_instruct_statistics} provides a detailed breakdown of the overall \instruct{} stage 1.
In addition, specific statistics for the cross-modal and diagram subsets are presented in \autoref{tab:stg1_cross_modal_instruct_statistics} and \autoref{tab:stg1_diagram_instruct_statistics}, respectively. Futhermore, \autoref{fig:inst_stage1_dist_stat} shows the programming language distribution in \instruct{} stage 1 data.

\noindent \textbf{\instruct{} Stage 1 Lightweight Version.}\quad
Furthermore, we constructed a lightweight subset to facilitate more convenient and low-cost model training. We selected approximately 4.25M data instances with shorter total input-output lengths, the majority of which have a total input-output length of less than 2048 (tokenized with Qwen2-VL~\citep{qwen_vl2}). Detailed statistics can be found in \autoref{tab:stg1_light_statistics}, \autoref{tab:stg1_light_cross_modal_statistics}, and \autoref{tab:stg1_light_diagram_statistics}.

\begin{figure*}[h]
\centering
\begin{minipage}[l]{0.32\textwidth}
\small
\centering
\captionof{table}{\instruct{} stage 1 statistics.}
\centering
 \renewcommand\arraystretch{0.96} 
\begin{adjustbox}{width=\linewidth}

\begin{tabular}{lr}
  \toprule
  \textbf{Statistic} & \textbf{Number} \\
  \midrule
  Total problems & 12962781 \\
  Total images & 42360470 \\
  \midrule
  Max. Problem length & 4538 \\
  Min. Problem length & 3 \\
  Avg. Problem length & 452\\
  \midrule
  Max. Response length & 5362 \\
  Min. Response length & 5 \\
  Avg. Response length & 995 \\
  \midrule
  Max. Image height & 13345 \\
  Min. Image height & 24 \\
  Avg. Image height & 315\\
  \midrule
  Max. Image width & 57246 \\
  Min. Image width & 10 \\
  Avg. Image width & 619 \\
\bottomrule
  \label{tab:stg1_instruct_statistics}
 \end{tabular}

 \end{adjustbox}
\end{minipage}
\quad
\begin{minipage}[l]{0.31\textwidth}
\small
\centering
\captionof{table}{\instruct{} stage 1 cross-modal statistics.}
\begin{adjustbox}{width=\linewidth}

\begin{tabular}{lr}
  \toprule
  \textbf{Statistic} & \textbf{Number} \\
  \midrule
  Total problems & 11976588 \\
  Total images & 41374277 \\
  \midrule
  Max. Problem length & 4538 \\
  Min. Problem length & 3 \\
  Avg. Problem length & 426 \\
  \midrule
  Max. Response length & 5362 \\
  Min. Response length & 8 \\
  Avg. Response length & 1011 \\
  \midrule
  Max. Image height & 13345 \\
  Min. Image height & 24 \\
  Avg. Image height & 291 \\
  \midrule
  Max. Image width & 57246 \\
  Min. Image width & 10 \\
  Avg. Image width & 618 \\
\bottomrule
  \label{tab:stg1_cross_modal_instruct_statistics}
 \end{tabular}
 \end{adjustbox}
\end{minipage}
\quad
\begin{minipage}[l]{0.3\textwidth}
\small
\centering
\captionof{table}{\instruct{} stage 1 diagram statistics.}
\begin{adjustbox}{width=\linewidth}

\begin{tabular}{lr}
  \toprule
  \textbf{Statistic} & \textbf{Number} \\
  \midrule
  Total problems & 986193 \\
  Total images & 986193 \\
  \midrule
  Max. Problem length & 2041 \\
  Min. Problem length & 7 \\
  Avg. Problem length & 762 \\
  \midrule
  Max. Response length & 1828 \\
  Min. Response length & 5 \\
  Avg. Response length & 796 \\ 
  \midrule
  Max. Image height & 8018 \\
  Min. Image height & 37 \\
  Avg. Image height & 1332 \\
  \midrule
  Max. Image width & 784 \\
  Min. Image width & 153 \\
  Avg. Image width & 675 \\
\bottomrule
  \label{tab:stg1_diagram_instruct_statistics}
 \end{tabular}

 \end{adjustbox}
\end{minipage}
\vspace{-10mm}
\end{figure*}


\begin{figure*}[h]
\centering
\begin{minipage}[l]{0.32\textwidth}
\small
\centering
\captionof{table}{\instruct{} lightweight version stage 1 statistics.}
\centering
 \renewcommand\arraystretch{0.96} 
\begin{adjustbox}{width=\linewidth}

\begin{tabular}{lr}
  \toprule
  \textbf{Statistic} & \textbf{Number} \\
  \midrule
  Total problems & 4249102 \\
  Total images & 15582159 \\  
  \midrule
  Max. Problem length & 1527 \\
  Min. Problem length & 3 \\
  Avg. Problem length & 420 \\
  \midrule
  Max. Response length & 1580 \\  
  Min. Response length & 50 \\
  Avg. Response length & 819 \\
  \midrule
  Max. Image height & 8078 \\
  Min. Image height & 24 \\
  Avg. Image height & 180 \\
  \midrule
  Max. Image width & 21859 \\
  Min. Image width & 10 \\
  Avg. Image width & 509 \\
\bottomrule
  \label{tab:stg1_light_statistics}
 \end{tabular}

 \end{adjustbox}
\end{minipage}
\quad
\begin{minipage}[l]{0.31\textwidth}
\small
\centering
\captionof{table}{\instruct{} lightweight version stage 1 cross-modal statistics.}
\begin{adjustbox}{width=\linewidth}

\begin{tabular}{lr}
  \toprule
  \textbf{Statistic} & \textbf{Number} \\
  \midrule
  Total problems & 4112529 \\
  Total images & 15445586 \\  
  \midrule
  Max. Problem length & 1445 \\
  Min. Problem length & 3 \\
  Avg. Problem length & 413 \\
  \midrule
  Max. Response length & 1580 \\
  Min. Response length & 50 \\
  Avg. Response length & 827 \\
  \midrule
  Max. Image height & 8078 \\
  Min. Image height & 24 \\
  Avg. Image height & 174 \\
  \midrule
  Max. Image width & 21859 \\
  Min. Image width & 10 \\
  Avg. Image width & 509 \\
\bottomrule
  \label{tab:stg1_light_cross_modal_statistics}
 \end{tabular}

 \end{adjustbox}
\end{minipage}
\quad
\begin{minipage}[l]{0.3\textwidth}
\small
\centering
\captionof{table}{\instruct{} lightweight version stage 1 diagram statistics.}
\begin{adjustbox}{width=\linewidth}

\begin{tabular}{lr}
  \toprule
  \textbf{Statistic} & \textbf{Number} \\
  \midrule
  Total problems & 136573 \\
  Total images & 136573 \\
  \midrule  
  Max. Problem length & 1527 \\
  Min. Problem length & 156 \\
  Avg. Problem length & 628 \\
  \midrule
  Max. Response length & 1429 \\
  Min. Response length & 84 \\
  Avg. Response length & 578 \\
  \midrule
  Max. Image height & 4084 \\
  Min. Image height & 37 \\
  Avg. Image height & 806 \\
  \midrule
  Max. Image width & 784 \\
  Min. Image width & 153 \\
  Avg. Image width & 602 \\
\bottomrule
  \label{tab:stg1_light_diagram_statistics}
 \end{tabular}

 \end{adjustbox}
\end{minipage}

\end{figure*}

\paragraph{\instruct{} Stage 2.}

\begin{figure*}[h]
\centering
\begin{minipage}[l]{0.32\textwidth}
\small
\centering
\captionof{table}{\instruct{} stage 2 statistics.}
\centering
 \renewcommand\arraystretch{0.96} 
\begin{adjustbox}{width=\linewidth}

\begin{tabular}{lr}
  \toprule
  \textbf{Statistic} & \textbf{Number} \\
  \midrule
  Total problems & 168178 \\
  Total images & 168178 \\
  \midrule
  Max. Problem length & 2032 \\
  Min. Problem length & 31 \\
  Avg. Problem length & 277\\
  \midrule
  Max. Response length & 1978 \\
  Min. Response length & 8 \\
  Avg. Response length & 400 \\
  \midrule
  Max. Image height & 9694 \\
  Min. Image height & 26 \\
  Avg. Image height & 1325\\
  \midrule
  Max. Image width & 1484 \\
  Min. Image width & 125 \\
  Avg. Image width & 631 \\
\bottomrule
  \label{tab:stg2_instruct_statistics}
 \end{tabular}

 \end{adjustbox}
\end{minipage}
\hspace{0.15\textwidth}
\begin{minipage}[l]{0.5\textwidth}
\small
\centering

\begin{adjustbox}{width=\linewidth}
\includegraphics[width=1.0\linewidth]{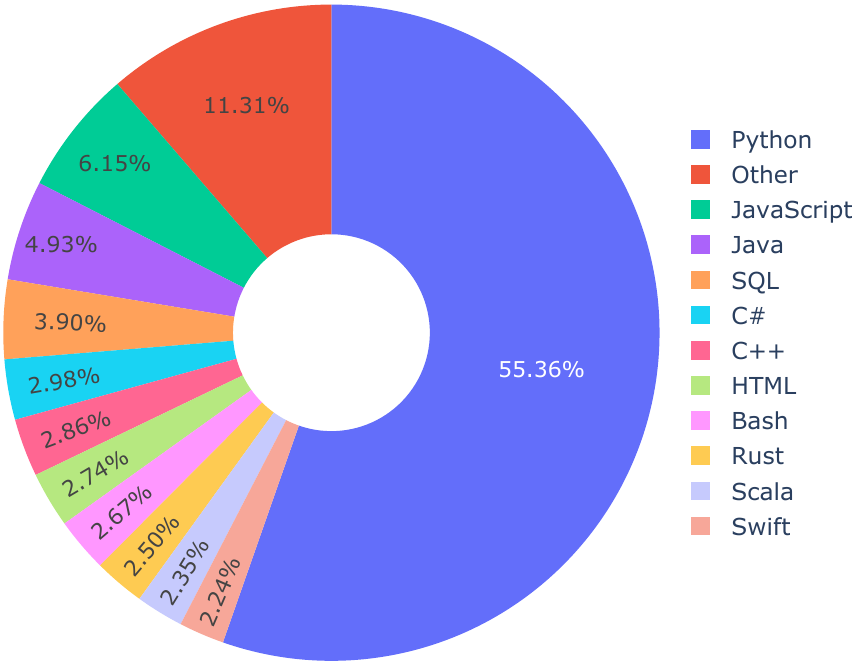}
 \end{adjustbox}
  \caption{Programming language distribution of \instruct{} Stage 2.}
  \label{fig:instruct_stage2_lang_dist_data}
\end{minipage}

\end{figure*}


\subsection{\instruct{} Construction Process}
This section provides a detailed account of the specific methodologies and procedures used for data synthesis and quality control during the construction of \instruct{}.

\noindent \textbf{Source Data Preparation.}\quad

As illustrated in \autoref{fig:instruct_curation}, \instruct{} comprises two stages of data preparation for fine-tuning.
For Stage 1 of \instruct{}, we first collected a large-scale, multilingual code dataset from GitHub, and then sampled and processed the data according to the following steps:
\begin{itemize}[leftmargin=*]
\item We preprocessed the data following the StarCoder~\citep{starcoder} data processing pipeline and employed rule-based methods to filter out invalid data (e.g., duplicates, garbled text, illegal information, etc.).
\item By prompting Qwen2.5-Coder-32B, we converted the data into a question-answer pair format, generating 12.9 million question-answer pairs. These pairs serve as the foundation for synthesizing multimodal instruction-tuning data.
\end{itemize}

For the \instruct{} stage 2, following the Magicoder~\citep{magicoder}, we employed two widely used datasets, Evol-CodeAlpaca\citep{wizardcoder} and OSS-Instruct\citep{magicoder}, to further synthesize multimodal diagram problems.

\begin{figure}[ht]
    \centering
    \includegraphics[width=0.9\linewidth]{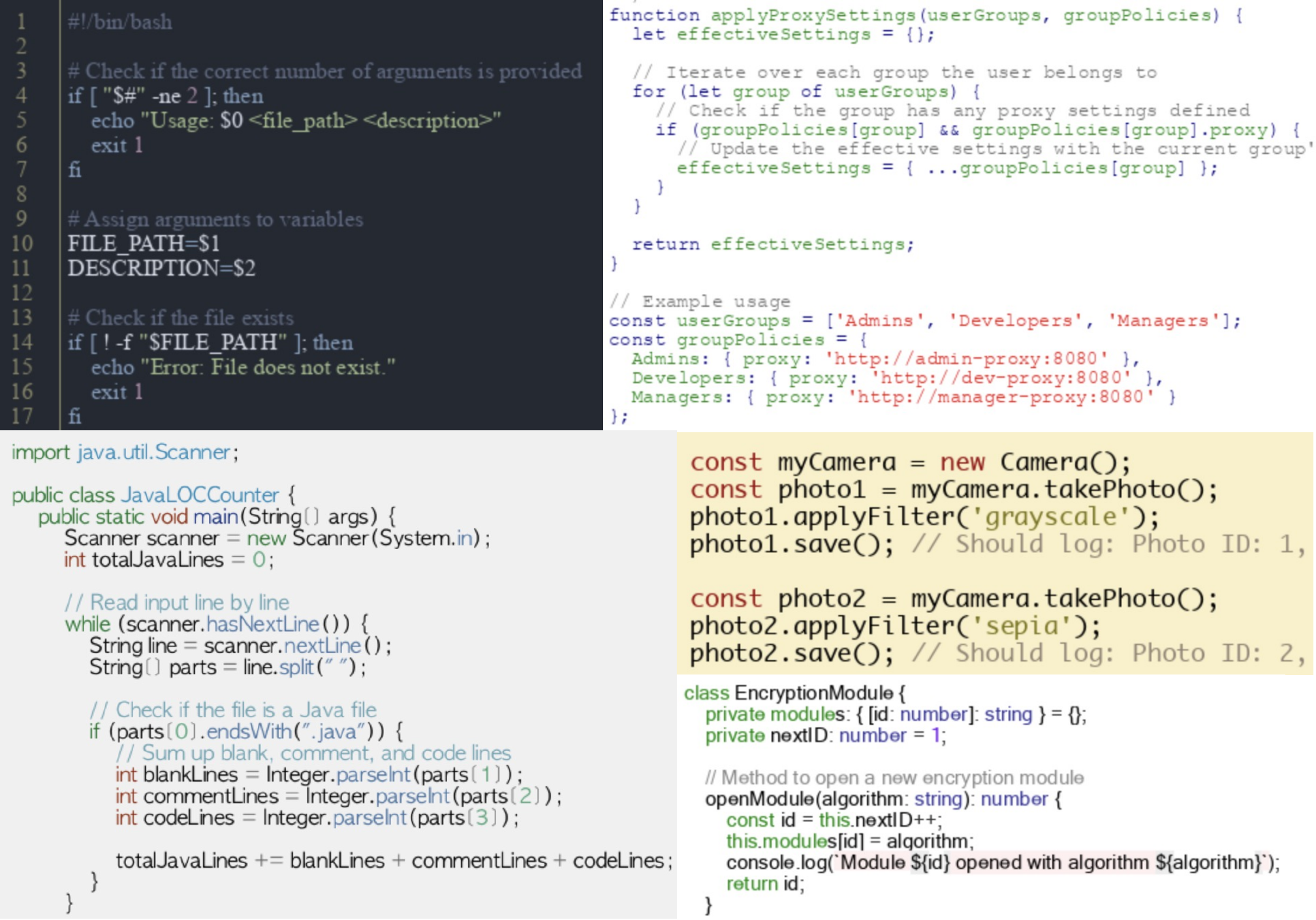}
    \caption{Examples of rendered code images via pygments.}
    \label{fig:render_code_image_examples}
\end{figure}

\paragraph{Cross-Modal Problem.}
The Cross-Modal problem refers to converting the code segments within questions into the image modality to enhance the code model's capabilities in visual code understanding and Optical Character Recognition (OCR). We utilize the code syntax highlighting tool \texttt{Pygments}\footnote{https://pygments.org/} to render the code within the questions.
Examples of rendered code images are illustrated in \autoref{fig:render_code_image_examples}. The complete cross-modal examples can be found in \autoref{supp:examples}.

\paragraph{Diagram Problem.}

\begin{figure}[ht]
    \centering
    \includegraphics[width=0.9\linewidth]{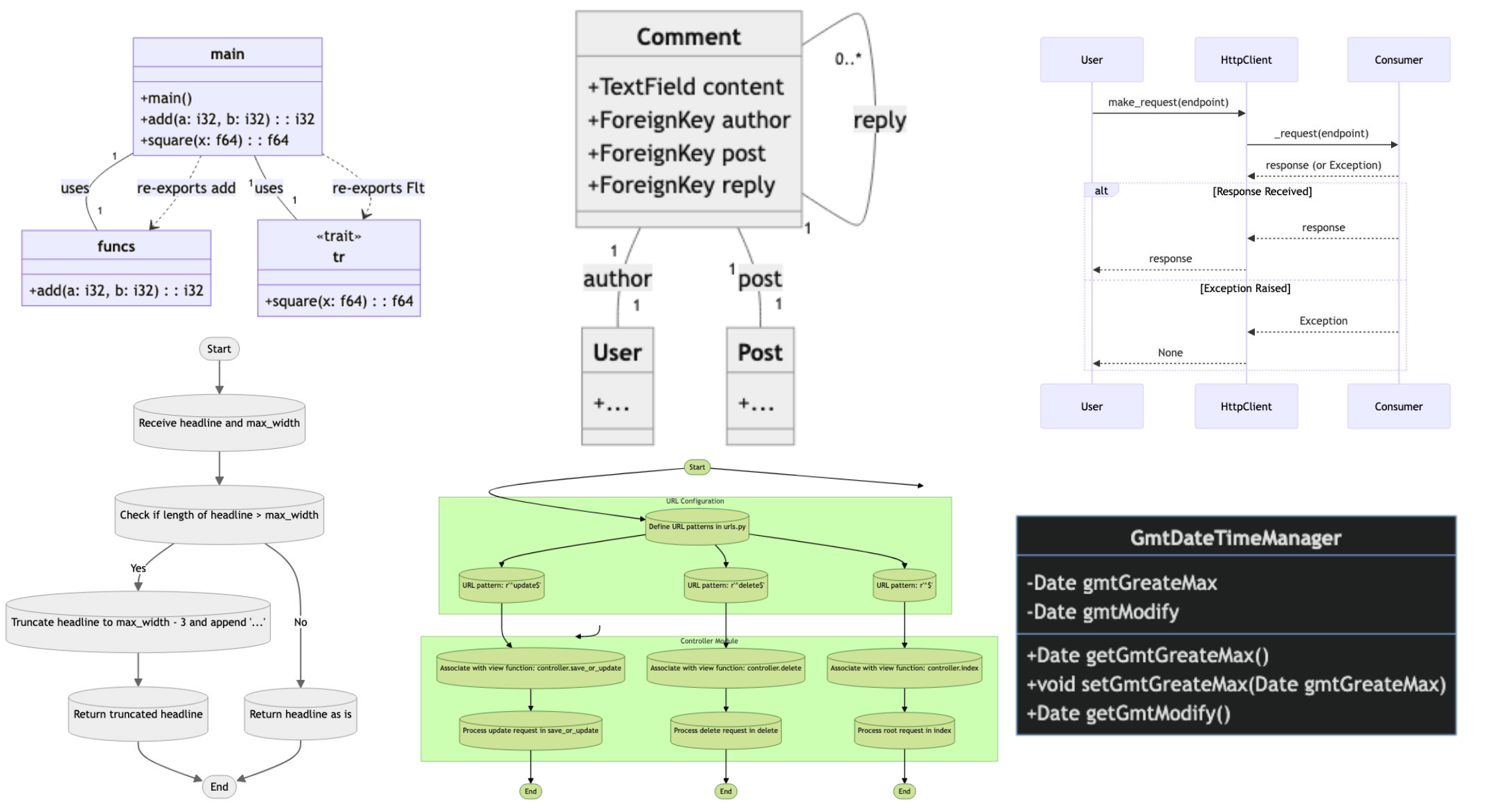}
    \caption{Examples of diagram code images via mermaid.}
    \label{fig:diagram_code_image_examples}
\end{figure}

The concept of a 'Diagram-related problem' encompasses tasks that necessitate generating executable code to address a problem, drawing information from both a textual description and an accompanying diagram. We introduce a three-step synthesis methodology designed to produce high-fidelity problems of this nature.
In the first step (Step 1), illustrated in \autoref{fig:diagram_syn_step1}, Qwen2.5-Coder-32B is initially prompted to generate a diagrammatic representation derived from a standard code problem. Following this, an attempt is made to render the generated diagram using Mermaid, with instances of unsuccessful rendering being systematically excluded.
In the second step (Step 2), as shown in \autoref{fig:diagram_syn_step2}, Qwen2.5-Coder is again prompted, this time to formulate a multimodal problem. This involves synthesizing the original problem statement with the diagram produced in Step 1. A critical instruction is given to ensure that vital information, indispensable for solving the coding challenge, is exclusively contained within the diagram. This constraint effectively mandates the use of the diagram during the problem-solving phase.
In the third step (Step 3), the Mermaid code resulting from Step 2 is employed to render the final visual diagrams with mermaid-cli tools\footnote{https://github.com/mermaid-js/mermaid-cli}. Illustrative examples of these rendered diagrams are presented in \autoref{fig:diagram_code_image_examples}. A complete set of examples can be found in \autoref{supp:examples}.

\begin{figure}[h]
\begin{center}
        \begin{tcolorbox}[title=Prompt Template For Step 1, instruction]
\textbf{ \#\#\# Please gain inspiration from Program Problem and Solution to create a high-quality **Mermaid Diagram code** to describe the code in Problem or Solution. } \\

\textbf{\#\#\#\# Problem for inspiration: } \\
\textcolor{orange}{\bf\{problem\}} \\

\textbf{\#\#\#\# Solution: }\\
\textcolor{orange}{\bf\{solution\}} \\

\textbf{\#\#\#\# Guidelines for creating mermaid diagram code:}\\
1. Identify and highlight the key information from the problem and solution to include in the Mermaid diagram, such as main processes, detail args, steps, decision points, and outcomes. \\
2. Use Mermaid syntax to visually represent the flow of the problem and solution. For example, you can use **flowcharts, sequence diagrams, or class diagrams .etc** depending on the complexity and nature of the problem and solution. \\
3. Place the Mermaid code inside \`{}\`{}\`{}mermaid\`{}\`{}\`{}. \\
4. Description in the diagram should be natural, here is an example: \\
\begin{minted}{python}
flowchart TD
    Start([Start]) --> Input[("Receive arrays A and B of length n")]
    Input --> Init[("Initialize result array C with size 2n-1")]
    Init --> Loop["Iterate i from 0 to 2n-2"]
    Loop --> InnerLoop["For each i, iterate j from max(0, i-n+1) to min(i, n-1)"]
    InnerLoop --> Compute[("Add A[j] * B[i-j] to C[i]")]
    Compute --> InnerLoop
    InnerLoop -- End --> Loop
    Loop -- End --> Output[("Output array C")]
    Output --> End([End]) 

\end{minted}

5. Most Importantly, **ensuring all text is described within quotes** to avoid syntax errors. \\

    \end{tcolorbox}
\end{center}
    \caption{Prompt template for step 1 of constructing diagram-type data in \instruct{}.}
    \label{fig:diagram_syn_step1}
\end{figure}

\begin{figure}[h]
\begin{center}
        \begin{tcolorbox}[title=Prompt Template For Step 2, instructionnext]
\textbf{\#\#\# Please gain inspiration from the following Problem, Solution to create a high-quality **multi-modal** programming problem related to the **provided Diagram**. Present your output in two distinct sections: \textcolor{red}{[Incomplete Problem]} and \textcolor{red}{[Solution]}.}\\

\textbf{\#\#\#\# Problem for inspiration:} \\
\textcolor{orange}{\bf\{problem\}} \\

\textbf{\#\#\#\# Solution for inspiration:}\\
\textcolor{orange}{\bf\{solution\}} \\

\textbf{\#\#\#\# Diagram:} \\
\textcolor{orange}{\bf\{mermaid\_code\}} \\

\textbf{\#\#\#\# Guidelines for each section:} \\
\textbf{1. \textcolor{red}{[Incomplete Problem]}:} \\
- This problem is incomplete, and can not be solved with only your generated [Incomplete Problem], some key information is only provided by the diagram. \\
    - Replace some steps or details with "xxx is detailed in the provided diagram" or "xxx could be found in the diagram". \\
    - Ensure [Incomplete Problem] is incomplete, must refer to the diagram for supplementary information. \\
    - Don't need to include the diagram code in the problem, the diagram will be provided as an image. \\

\textbf{2. \textcolor{red}{[Solution]}}: \\
    - Offer a comprehensive, **correct** solution that accurately addresses the [Problem] you provided. \\
    - Don't generate the main or check function. \\

    \end{tcolorbox}
\end{center}
    \caption{Prompt template for step 2 of constructing diagram-type data in \instruct{}.}
    \label{fig:diagram_syn_step2}
\end{figure}



\clearpage
\section{Experimental Details}
\label{supp:experiment}

\subsection{Model Source}

In \autoref{tab:model_source}, we list the names, release time, and source links of all the models we used in data synthesis, evaluation, and training.

\begin{table*}[h]
    \centering
    \small
     \fontsize{7.8pt}{\baselineskip}\selectfont 
     \renewcommand\tabcolsep{1.0pt} 
     \renewcommand\arraystretch{1.0} 
    \caption{The Release Time and Model Source of LLMs used in our work.}
    \label{tab:model_list}

    \resizebox{1.0\textwidth}{!}{ 
    \begin{tabular}{l@{\hspace{0.3cm}}c@{\hspace{0.3cm}}l}
    \toprule
    \textbf{Model} & \textbf{Release Time} & \textbf{\makecell[c]{Source}} \\
    \midrule
    \multicolumn{3}{l}{\textit{\textbf{LLMs}}} \\
    DeepSeek-V3~\cite{deepseekv3}     &   2025-03    & \url{https://console.volcengine.com/ark/} \\
    Qwen2.5-Coder-32B-Instruct~\cite{qwen25coder} &   2024-12    & \url{https://huggingface.co/Qwen/Qwen2.5-Coder-32B-Instruct} \\
    \midrule
    \multicolumn{3}{l}{\textit{\textbf{Proprietary LMMs}}} \\
    GPT-4o~\cite{gpt4o} & 2024-08  & \url{https://platform.openai.com/} \\
    GPT-4o-mini~\cite{gpt4o} & 2024-08  & \url{https://platform.openai.com/} \\
    Claude-3.7-Sonnet~\cite{claude3} & 2025-02   & \url{https://console.anthropic.com/} \\
    Claude-3.5-Sonnet~\cite{claude3} & 2024-06   & \url{https://console.anthropic.com/} \\
    Claude-3-Opus~\cite{claude3}  & 2024-02  & \url{https://console.anthropic.com/} \\
    Gemini-2.5-Flash-preview~\cite{gemini25} & 2025-04  & \url{https://aistudio.google.com/} \\
    Doubao1.5-thinking-pro~\cite{doubao_thinking} & 2025-04    & \url{https://console.volcengine.com/ark/} \\
    Doubao1.5-vision-pro~\cite{doubao_thinking} & 2025-03   & \url{https://console.volcengine.com/ark/} \\
    Qwen-VL-Max~\cite{qwen_vl2} & 2023-11  & \url{https://bailian.console.aliyun.com/} \\

    \midrule
    \multicolumn{3}{l}{\textit{\textbf{Open-Weight LMMs}}} \\
    Qwen2-VL-Instruct~\cite{qwen_vl2} & 2025-03 & \url{https://huggingface.co/Qwen/Qwen2-VL-7B-Instruct} \\
    Qwen2.5-VL-Instruct~\cite{Qwen2.5-VL} & 2025-02 & \url{https://huggingface.co/Qwen/Qwen2.5-VL-32B-Instruct} \\
    QVQ-Preview~\cite{qvq-72b-preview} & 2025-01 & \url{https://huggingface.co/Qwen/QVQ-72B-Preview} \\
    InternVL2~\cite{internvl25} & 2024-07 & \url{https://huggingface.co/OpenGVLab/InternVL2-Llama3-76B} \\
    InternVL3~\cite{internvl3} & 2025-04 & \url{https://huggingface.co/OpenGVLab/InternVL3-8B} \\
    Llama3.2-vision~\cite{llama3} & 2025-09 & \url{https://huggingface.co/meta-llama/Llama-3.2-90B-Vision-Instruct} \\      
    LLama-4~\cite{llama4} & 2025-04 & \url{https://huggingface.co/meta-llama/Llama-4-Maverick-17B-128E-Instruct} \\
    Kimi-VL-Thinking~\cite{kimivl} & 2025-04 & \url{https://huggingface.co/moonshotai/Kimi-VL-A3B-Thinking} \\
    Gemma-3-it~\cite{gemma3} & 2025-03 & \url{https://huggingface.co/google/gemma-3-27b-it} \\
    DeepSeek-VL2~\cite{deepseekvl2} & 2024-12 & \url{https://huggingface.co/deepseek-ai/deepseek-vl2} \\
    Phi3-vision~\cite{phi_3} & 2024-05 & \url{https://huggingface.co/microsoft/Phi-3-vision-128k-instruct} \\
    MiniCPM-V-2.6~\cite{minicpm_v} & 2024-08 & \url{https://huggingface.co/openbmb/MiniCPM-V-2_6} \\
    \bottomrule
    \end{tabular}
    }
    \label{tab:model_source}
\end{table*}

\subsection{\coder{} Training}
In this section, we list in detail the hyperparameters we set during the \coder{} training. Our training tasks utilized 8 NVIDIA A800 GPUs to complete all experiments. All training works are done with the open-source training framework LlamaFactory\citep{llamafactory}.

\paragraph{Stage 1 fine-tuning}
We initialize our model from the Qwen2-VL-7B-base. Stage 1 fine-tuning is performed using a full fine-tuning approach, where the vision tower, multi-modal projector, and language model are all unfrozen and trained.
Training is conducted on the \instruct{}-stage-1. Input data is formatted using the qwen2\_vl template, with sequences truncated or padded to a maximum length of 2048 tokens. 
For optimization,  we employ AdamW~\citep{adamw} as the optimizer, with a batch size of 1024 and a maximum sequence length of 2048. The model is trained for 1.0 epoch with a learning rate of \(5 \times 10^{-5}\). A cosine learning rate scheduler is employed with a warmup ratio of 0.1. Training is performed with bfloat16 (bf16) mixed precision and accelerated using DeepSpeed with a ZeRO Stage 2 configuration.

\paragraph{Stage 2 fine-tuning}
In Stage 2, we initialize our model from the Stage 1 checkpoint and continue with SFT. During this stage, the vision tower and multi-modal projector are kept frozen, and only the language model parameters are fine-tuned. Training is conducted on the \instruct{}-stage-2. Input data is formatted using the qwen2\_vl template, with sequences truncated or padded to a maximum length of 6000 tokens.
For optimization, we employ AdamW as the optimizer, with a global batch size of 1024 and a maximum sequence length of 6000. The model is trained for 2.0 epochs with a learning rate of 5e-5. A cosine learning rate scheduler with a warmup ratio of 0.1 is utilized. Training is performed with bfloat16 mixed precision and accelerated using DeepSpeed with a ZeRO Stage 2 configuration.

\subsection{Evaluation Setup}
\textbf{Evaluation Environment.}\quad
\autoref{tab:runtime_env}~presents the code sandbox environment used in our evaluation, including version information for 10 programming languages. 
The construction of the environment refers to MultiPL-E\citep{multipl_e}.
We will release this environment as an open-source Dockerfile to facilitate future evaluation and testing.

\textbf{Model Inference.}\quad
To test proprietary and large-scale models, we utilized the APIs accessible via the links in \autoref{tab:model_source} for inference. For the majority of open-weight LMMs, model weights were downloaded from the links in \autoref{tab:model_source}, and inference was conducted using the vLLM (v0.8.5) framework~\citep{vllm}. Additionally, for regular models, greedy decoding was applied with temperature=0 and a maximum output length of 2048. For "thinking" models, we configured the temperature to 0.6, top\_p to 0.95, and the maximum length to 4096.

\begin{table}[h]
    \centering
    \small
        \caption{Runtime environments for 10 programming languages.}
        \label{tab:runtime_env}
    \begin{tabular}{l|l}
    \toprule
    Language & Runtime Environments\\
    \midrule 
        C\# & Mono C\# compiler version 6.12.0.200 \\ 
        CPP & g++ (Ubuntu 11.4.0-1ubuntu1~22.04) 11.4.0 \\ 
        Java & openjdk version "11.0.26" 2025-01-21 \\ 
        JavaScript & Node.js v23.10.0 \\ 
        Kotlin & kotlinc-jvm 2.1.10 (JRE 11.0.26+4-post-Ubuntu-1ubuntu122.04) \\ 
        PHP & PHP 8.1.2-1ubuntu2.20 (cli) (built: Dec  3 2024 20:14:35) (NTS) \\ 
        Python & Python 3.10.12 \\ 
        Ruby & ruby 3.0.2p107 (2021-07-07 revision) [x86\_64-linux-gnu] \\ 
        Scala & Scala code runner version 3.3.5 \\ 
        Swift & Swift version 6.0.3 (swift-6.0.3-RELEASE)\\ 
    \bottomrule
    \end{tabular}
    
\end{table}

\subsection{Additional Analysis}

\begin{figure}[htb]
	\centering

    {\includegraphics[width=0.6\textwidth]{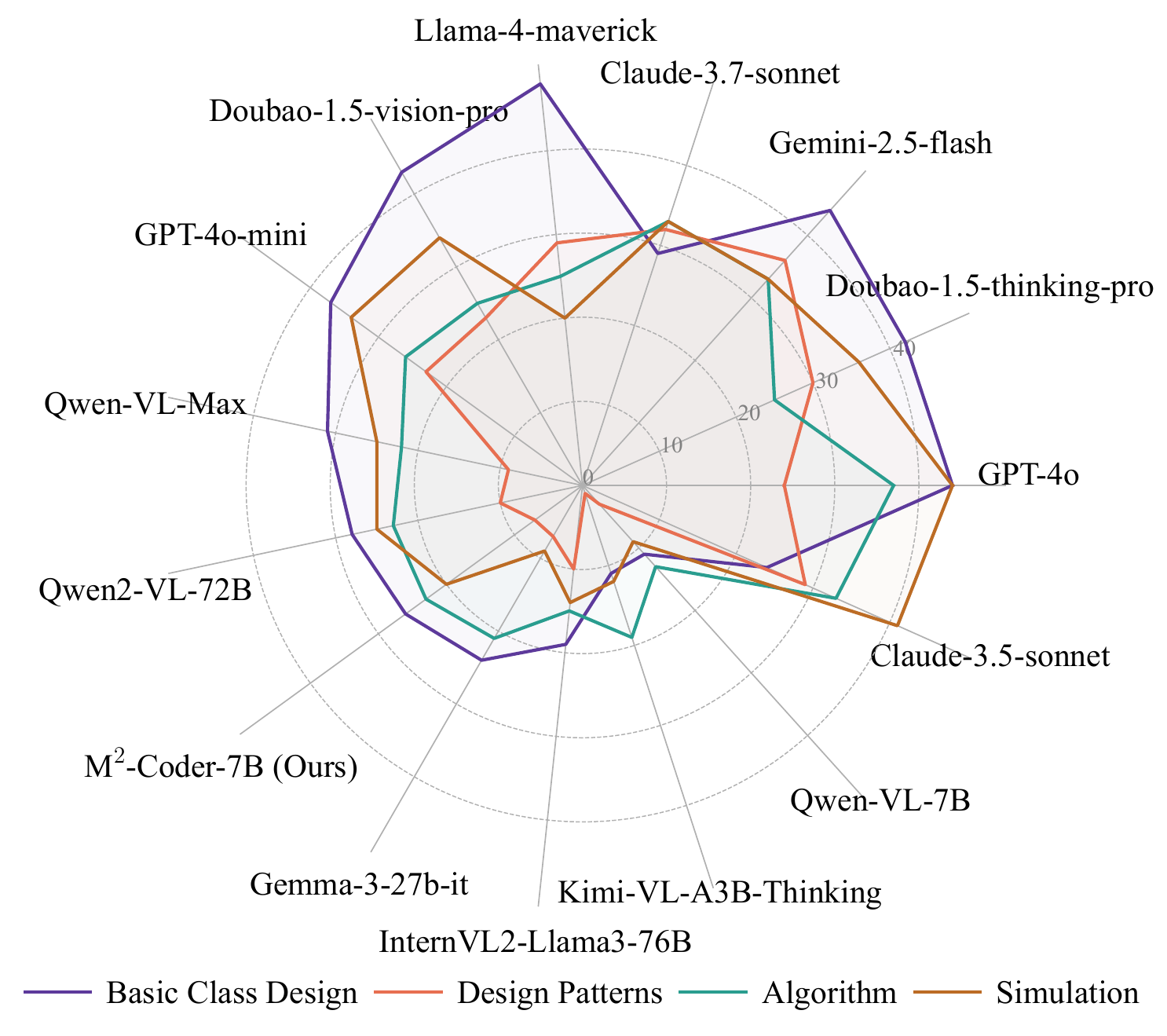}}
	\caption{Performance comparison of models across task types.}
	\label{fig:supp_pl_performance_stat}
    \vspace{-10pt}
\end{figure}

\autoref{fig:supp_pl_performance_stat} reveals that our \coder{} model demonstrates competitive performance, with its 7B parameters achieving results comparable to some models around the 70B scale, particularly in Basic Class Design tasks. Concurrently, leading models such as GPT-4o and Gemini-2.5-flash exhibit a significant lead in Design Patterns tasks. This disparity underscores the critical importance of underlying programming knowledge and reasoning for tasks involving diagrams, as proficiency in areas like Design Patterns, which often rely on diagrammatic representation, is a key differentiator and highlights a crucial aspect for understanding and generating code from visual specifications.

\begin{figure*}[ht]
\begin{center}
    \includegraphics[width=1.0\textwidth]{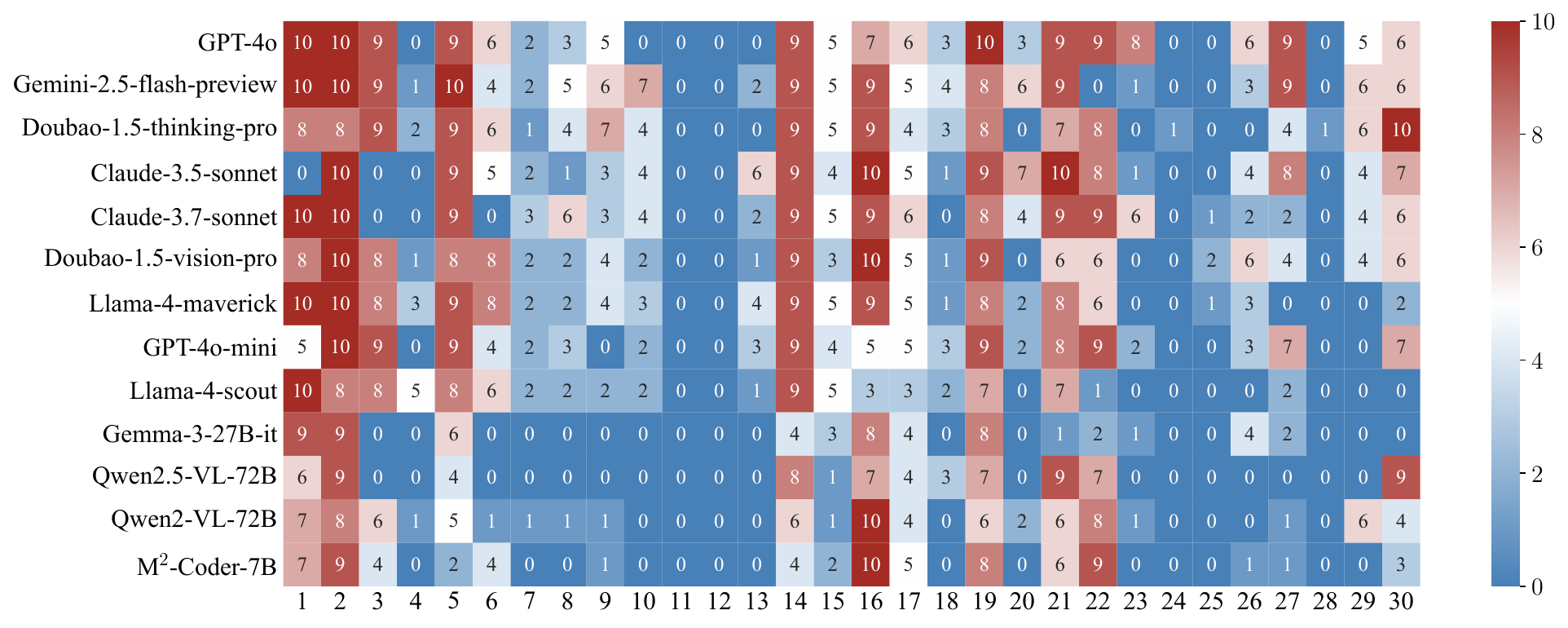}
    \caption{Scores of models on each problem.}
    \label{fig:heatmap_agg_by_model}
    \vspace{-10pt}
\end{center}
\end{figure*}

\autoref{fig:heatmap_agg_by_model} shows the model's performance on different problems (a score of 10 indicates that the code in all languages passes the tests). It can be seen that on some simple problems (e.g., problems 1, 2, 16, and 19), most models perform well in all languages. However, on some complex problems, the model's performance is relatively poor. This indicates that even for the same problem, the choice of programming language for the solution significantly impacts accuracy.

\subsection{Cases for Error Analysis}
\label{supp:error_analysis}

\begin{figure*}[h]
\begin{center}
    \includegraphics[width=1.0\textwidth]{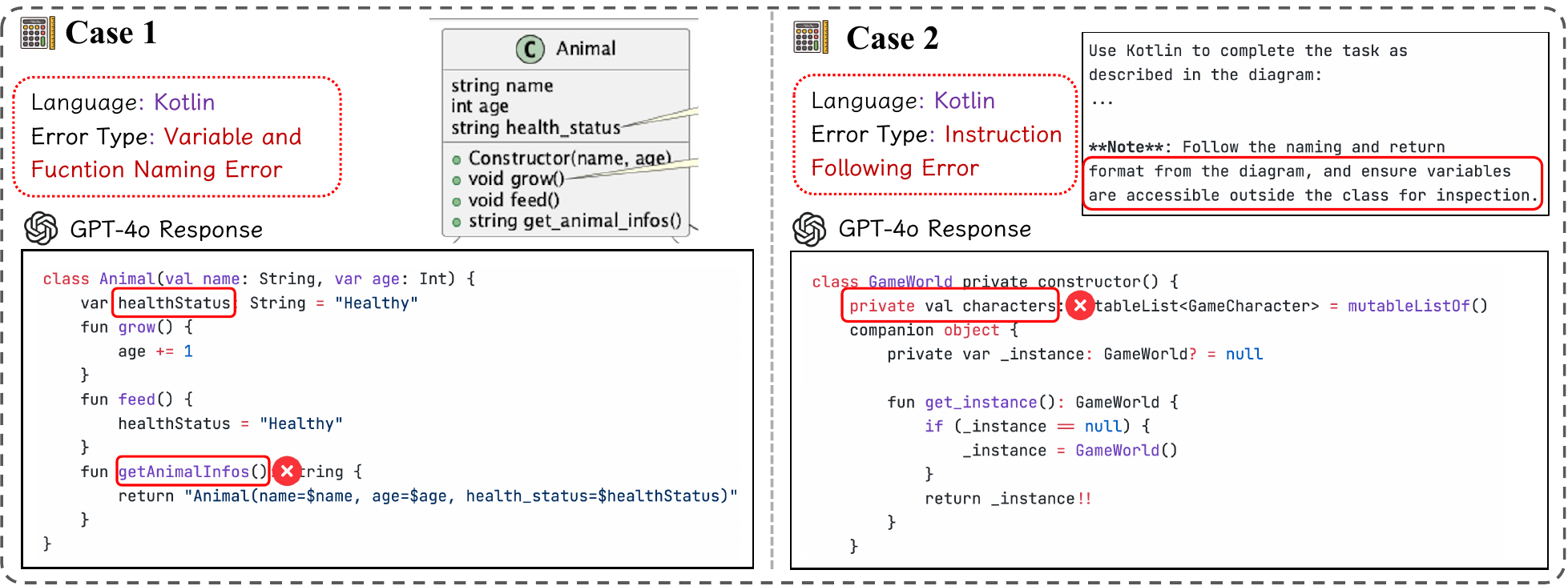}
    \caption{Kotlin error cases from GPT-4o on \benchmark{}.}
    \label{fig:supp_error_case_visual}
    \vspace{-10pt}
\end{center}
\end{figure*}

\begin{figure*}[h]
\begin{center}
    \includegraphics[width=1.0\textwidth]{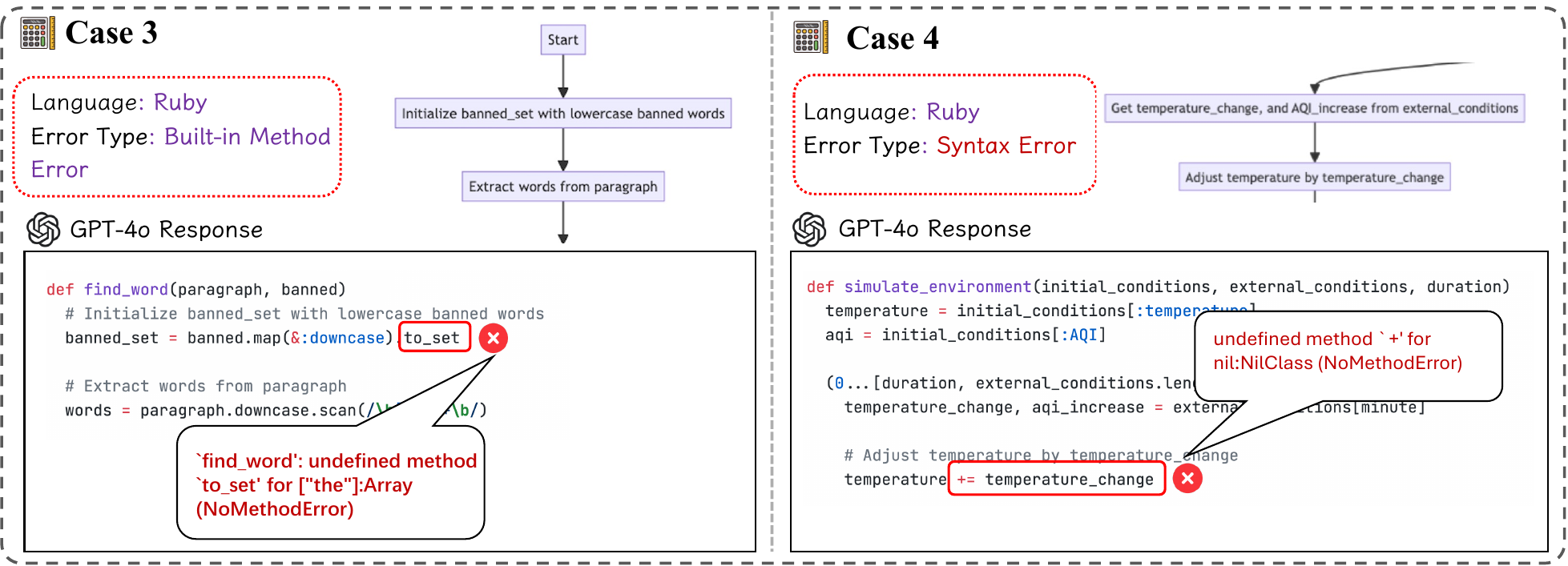}
    \caption{Ruby error cases from GPT-4o on \benchmark{}.}
    \label{fig:supp_error_case_pl}
    \vspace{-10pt}
\end{center}
\end{figure*}

In \autoref{fig:supp_error_case_visual} and \autoref{fig:supp_error_case_pl}, we present illustrative error cases pertaining to visual understanding and programming language generation by GPT-4o, one of today's leading Large Multimodal Models (LMMs). Despite its advanced capabilities, even this model makes some fundamental mistakes. For instance, Case 1 involves errors in variable and function naming; Case 2 exhibits instruction non-compliance; and Cases 3 and 4 reveal intrinsic language-specific errors. These examples underscore the multifaceted complexity of multimodal programming, highlighting persistent and significant challenges in areas such as the precise capture of visual information, robust instruction following, and fundamental programming language proficiency.

\clearpage
\section{Examples}
\label{supp:examples}

\subsection{\instruct{} Examples}
\input{examples/instruct_1}
\input{examples/instruct_2}
\input{examples/instruct_3}
\input{examples/instruct_4}
\input{examples/instruct_5}
\input{examples/instruct_6}
\input{examples/instruct_7}
\input{examples/instruct_8}

\clearpage
\subsection{\benchmark{} Examples}

\input{examples/bench_1}
\input{examples/bench_2}
\input{examples/bench_3}
\input{examples/bench_4}
\input{examples/bench_5}
\input{examples/bench_6}
\input{examples/bench_7}
\input{examples/bench_8}

\clearpage
\section{Detailed Related Work}
\textbf{Code Large Language Model.}~~With the rapid advancement of large language models(LLMs)\citep{gpt4,llama2,llama3,Qwen,qwen2}, solving complex code-related tasks has become increasingly feasible, leading to the emergence of numerous Code LLMs.
Early studies utilized models like BERT\citep{bert} or GPT\citep{gpt} as backbones, trained on billions of code snippets to enable tasks involving code understanding and generation\citep{chen2021pass_k, code_bert, bloom, AlphaCode,codet5,santacoder}. 
Recently, advancements in domain-specific pre-training and instruction fine-tuning techniques have led to extensive efforts in fine-tuning models on large-scale code corpora and crafting code-related task instructions\citep{code_llama,codegeex,wizardcoder,octopack,codegemma,opencodeinterpreter,deepseek_coder,magicoder,unicoder,starcoder2,mistral,qwen25coder}. These models demonstrate remarkable performance in tasks like code completion, synthesis, and interpretation.

\textbf{Code Evaluation.}~~To assess the code capabilities of LLMs, a wide range of benchmarks have been developed to evaluate code quality, functionality, and efficiency. Early efforts \citep{chen2021pass_k, mbpp, evalplus, bigcodebench} concentrated on the fundamental coding abilities of LLMs. However, given the complexity and multifaceted nature of code tasks, subsequent evaluations have expanded to include code repair~\citep{quixbugs, debugbench, EvalGPTFix, runbugrun, he2022distribution,wang2023delving}, multilingual code assessments~\citep{codegeex, multipl_e, odex, mbxp, mceval}, repository-level evaluations~\citep{repobench, r2c2coder, codeplan}, agent-based code evaluations~\citep{swe_bench, pybench}, and more. 

\textbf{Visual Reasoning and Code Synthesis.}~~Large Language Multmodal Models (LMMs)\cite{minigpt4,llava,Qwen_VL,mplug,llava_next,internlm_xcomposer,llavaonevision,qwen_vl2, zong2024mova}  incorporate visual information into LLMs through visual encoders\citep{clip}, extending the capabilities of LLMs to address visual tasks. 
Prior studies such as VQA\citep{vqa,vinvqa,mtvqa} evaluated the basic visual semantic capabilities of models. 
With the emergence of increasingly powerful visual and semantic capabilities in LMMs and LLMs, many recent works have shifted focus toward more complex multi-modal tasks, such as  mathematical reasoning\citep{trinh2024solvingolympiad, shao2024visual, huang2024olympicarena, mavis}, 
chart understanding\citep{chartllama,flowchartqa,flowvqa,wang2024charxiv,li-towards-real,multimodalselfinstruct}, code generation\citep{design2code,li2024mmcode,logomotion,wu2024plot2code,shi2024chartmimic,robocodex}, and agent-driven interactions\citep{webarena,osworld,visualagentbench}. 

Recent multimodal code works focused on visual algorithmic problems\citep{li2024mmcode, codevision}, chart code generation\cite{wu2024plot2code,shi2024chartmimic}, and UI design\citep{design2code, logomotion}. Our \benchmark{} explores the task of code generation based on code diagrams. Compared with previous work\citep{liu2022flow2code} that converts flowcharts into node information for complex processing, we focus on more practical scenarios that directly perform semantic understanding based on images.

%% file: showcase_format.tex
\definecolor{bluecol}{HTML}{2948DA}
\definecolor{purplecol}{HTML}{4A13B7}
\definecolor{purplecollight}{HTML}{9966FF}
\definecolor{redlight}{HTML}{FF6666}
\definecolor{graylight}{HTML}{646464}
\definecolor{green}{HTML}{00CC66}
\definecolor{goldanswercol}{HTML}{FF9900}
\definecolor{querycol}{HTML}{7964E8}
\definecolor{chorange}{HTML}{FFB43B}
\definecolor{otherscol}{HTML}{FC5BCF}
\input{colors}
\tcbset{
    instruction/.style={
        fonttitle=\normalsize,
        colframe=bluecol, 
        colback=white, 
        coltitle=white, 
        left=2mm,  
        right=2mm,
        fontupper=\scriptsize,
        fontlower=\scriptsize,
    },
    instructionnext/.style={
        fonttitle=\normalsize,
        colframe=purplecol, 
        colback=white, 
        coltitle=white, 
        left=2mm,  
        right=2mm,
        fontupper=\scriptsize,
        fontlower=\scriptsize,
    },
    showcase/.style={
        fonttitle=\large,
        colback=white!20,  
        colframe=graylight,   
        coltitle=white,   
        boxrule=0.5mm,    
        arc=2mm,          
        outer arc=2mm,    
        left=0.5mm,         
        right=0.5mm,        
        top=0.5mm,          
        bottom=0.5mm,       
        width=\textwidth, 
        before skip=0.1pt,
        after skip=0.1pt,
    },
    context/.style={
        fontupper=\scriptsize,
        fonttitle=\large,
        colframe=querycol,     
        coltitle=white,   
        colback=white,    
        boxrule=0.3mm,    
        arc=2mm,          
        outer arc=2mm,    
        left=1mm,         
        right=1mm,        
        top=1mm,          
        bottom=1mm,       
        before skip=1pt,
        after skip=0.1pt, 
    },
    problem/.style={
        fontupper=\scriptsize,
        fontlower=\scriptsize,
        colframe=green,
        coltitle=white,   
        colback=white,    
        fonttitle=\normalsize, 
        boxrule=0.3mm,    
        arc=0mm,          
        outer arc=0mm,    
        left=1mm,         
        right=1mm,        
        top=1mm,          
        bottom=1mm,       
        before skip=1pt,
        after skip=0.1pt,
    },
    solution/.style={
        colframe=goldanswercol, 
        coltitle=white,   
        colback=white,    
        fonttitle=\normalsize, 
        boxrule=0.3mm,    
        arc=0mm,          
        outer arc=0mm,    
        left=1mm,         
        right=1mm,        
        top=1mm,          
        bottom=1mm,       
        before skip=1pt,
        after skip=0.1pt,
        fontupper=\scriptsize,
        fontlower=\scriptsize,
    },
    query/.style={
        fontupper=\scriptsize,
        fontlower=\scriptsize,
        colframe=querycol,     
        coltitle=white,   
        colback=white,    
        boxrule=0.1mm,    
        arc=2mm,          
        outer arc=2mm,    
        left=1mm,         
        right=1mm,        
        top=1mm,          
        bottom=1mm,       
        before skip=1pt,
        after skip=0.1pt,
    },
    abab/.style={
        fontupper=\scriptsize,
        fonttitle=,
        colframe=ababcol, 
        coltitle=white,   
        boxrule=0.5mm,    
        arc=2mm,          
        outer arc=2mm,    
        left=1mm,         
        right=1mm,        
        top=1mm,          
        bottom=1mm,       
        width=0.33\textwidth, 
        before skip=0.1pt,
        after skip=0.1pt, 
    },
    others/.style={
        fontupper=\scriptsize,
        colframe=otherscol,     
        coltitle=white,
        boxrule=0.5mm,    
        arc=2mm,          
        outer arc=2mm,    
        left=1mm,         
        right=1mm,        
        top=1mm,          
        bottom=1mm,       
        width=0.33\textwidth, 
        before skip=0.1pt,
        after skip=0.1pt, 
    },
    goldanswer/.style={
        fontupper=\scriptsize,
        colframe=goldanswercol,     
        coltitle=white,   
        boxrule=0.5mm,    
        arc=2mm,          
        outer arc=2mm,    
        left=1mm,         
        right=1mm,        
        top=1mm,          
        bottom=1mm,       
        width=0.33\textwidth, 
        before skip=0.1pt,
        after skip=0.1pt, 
    },
}

%% file: colors.tex
\definecolor{myhailuo1dark}{HTML}{FC8900}
\definecolor{myhailuo2dark}{HTML}{F14738}
\definecolor{myhailuo3dark}{HTML}{D12AAA}
\definecolor{myhailuo4dark}{HTML}{4C4DC2}

\definecolor{myhailuo1}{HTML}{FFB43B}
\definecolor{myhailuo2}{HTML}{F97669}
\definecolor{myhailuo3}{HTML}{FC5BCF}
\definecolor{myhailuo4}{HTML}{7964E8}

\definecolor{myhailuo1light}{HTML}{FFD085}
\definecolor{myhailuo2light}{HTML}{FFA19F}
\definecolor{myhailuo3light}{HTML}{FFA9FA}
\definecolor{myhailuo4light}{HTML}{BDACFB}

\colorlet{myorange}{Orange!20}
\colorlet{mygreen}{LimeGreen!25}
\colorlet{myyellow}{Yellow!30}
\colorlet{myblue}{CornflowerBlue!25}
\colorlet{mybrown}{RawSienna!25}
\colorlet{mypurple}{Orchid!25}
\colorlet{myred}{Red!60}
\colorlet{myorangefull}{YellowOrange!60}
\colorlet{mybrownfull}{RawSienna!60}

\colorlet{myorangethick}{Orange!40}
\colorlet{mygreenthick}{LimeGreen!50}
\colorlet{myyellowthick}{Yellow!60}
\colorlet{mybluethick}{CornflowerBlue!50}

%% file: examples/instruct_1.tex
\begin{figure}[h]

\begin{center}
    \begin{tcolorbox}[title=\instruct{} Cross-Modal~Example 1, showcase] 
    \begin{tcolorbox}[title=Problem, problem]

You are tasked with writing a TypeScript program that simulates a simple file system on a hard drive. The file system should support basic operations such as creating files, deleting files, and listing files. Each file will have a name and a size. The hard drive has a fixed capacity, and the program should handle cases where the hard drive is full.

\#\#\# Requirements

1. **Create File**: Add a file to the file system. If the hard drive is full, return an error message.

2. **Delete File**: Remove a file from the file system.

3. **List Files**: Return a list of all files in the file system along with their sizes.

4. **Check Capacity**: Return the remaining capacity of the hard drive.

\#\#\# Example

       \fbox{ \includegraphics[width=0.4\linewidth]{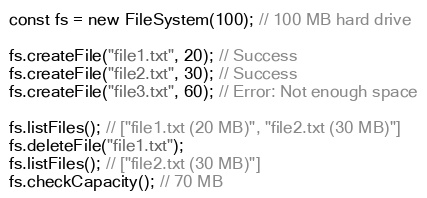}}
    \end{tcolorbox}

\begin{tcolorbox}[title=Response, solution]
Here is an improved version of your code with these considerations:

\begin{minted}{typescript}
class File {
    constructor(public name: string, public size: number) {
        if (size <= 0) {
            throw new Error("File size must be a positive number");
        }
    }
}
class FileSystem {
    private files: File[] = [];
    private totalCapacity: number;
    private usedCapacity: number = 0;

    constructor(capacity: number) {
...
\end{minted}
\end{tcolorbox}
\end{tcolorbox}
\end{center}

\caption{Cross-modal example 1 from \instruct{}.}
\label{fig:inst_exmaple1}
\end{figure}

%% file: examples/instruct_2.tex
\begin{figure}[h]

\begin{center}
    \begin{tcolorbox}[title=\instruct{} Cross-Modal~Example 2, showcase] 
    \begin{tcolorbox}[title=Problem, problem]

You are tasked with simulating the behavior of an Infrared LED (IR LED) in a security system. The IR LED has a specific power consumption and brightness level, which changes based on the ambient temperature. The goal is to create a JavaScript function that calculates the brightness of the IR LED given the ambient temperature and the power consumption. The brightness is calculated using the formula:

\[ \text{Brightness} = \text{Power Consumption} \times (1 - \frac{\text{Ambient Temperature} - 25}{100}) \]

Where:
- Power Consumption is in milliwatts (mW).
- Ambient Temperature is in degrees Celsius (°C).
- Brightness is in arbitrary units (AU).
...

       \fbox{ \includegraphics[width=0.8\linewidth]{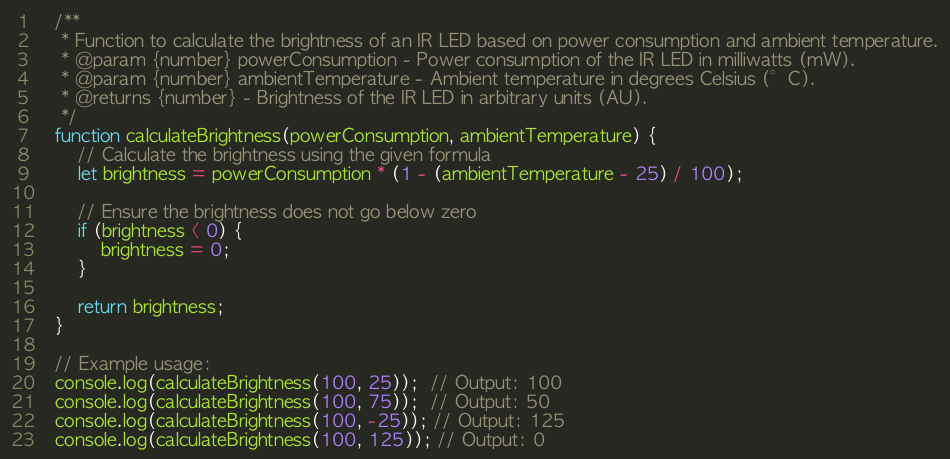}}
    \end{tcolorbox}

\begin{tcolorbox}[title=Response, solution]
The task is to simulate the behavior of an Infrared LED (IR LED) in a security system by calculating its brightness based on the power consumption and ambient temperature. The brightness is calculated using a specific formula, and we need to ensure that the brightness does not go below zero.

Here's the JavaScript function `calculateBrightness` that performs this calculation:

\begin{minted}{typescript}
/**
 * Function to calculate the brightness of an IR LED 
 based on power consumption and ambient temperature.
 * @param {number} powerConsumption - Power consumption of the IR LED in milliwatts (mW).
 * @param {number} ambientTemperature - Ambient temperature in degrees Celsius (°C).
 * @returns {number} - Brightness of the IR LED in arbitrary units (AU).
 */
function calculateBrightness(powerConsumption, ambientTemperature) {
    // Calculate the brightness using the given formula
    let brightness = powerConsumption * (1 - (ambientTemperature - 25) / 100);
    
    // Ensure the brightness does not go below zero
    if (brightness < 0) {
        brightness = 0;
    }
    
    return brightness;
}

// Example usage:
console.log(calculateBrightness(100, 25));  // Output: 100
console.log(calculateBrightness(100, 75));  // Output: 50
console.log(calculateBrightness(100, -25)); // Output: 125
console.log(calculateBrightness(100, 125)); // Output: 0    
...
\end{minted}
\end{tcolorbox}
\end{tcolorbox}
\end{center}

\caption{Cross-modal example 2 from \instruct{}.}
\label{fig:inst_exmaple2}
\end{figure}

%% file: examples/instruct_3.tex
\begin{figure}[h]

\begin{center}
    \begin{tcolorbox}[title=\instruct{} Cross-Modal~Example 3, showcase] 
    \begin{tcolorbox}[title=Problem, problem]
You are tasked with creating a JavaScript factory function that generates objects representing database records. This factory function should allow for the creation of objects with specific properties and methods, similar to how a plugin might extend a class in Perl's Class::DBI::Plugin::Factory. The factory should be able to handle different types of database records, such as `User` and `Product`, each with their own set of properties and methods.

\#\#\# Requirements

1. **Factory Function**: Create a factory function `createRecordFactory` that takes a `type` parameter and returns a constructor function for that type of record.

2. **Constructor Function**: The constructor function should initialize the object with properties based on the type.

3. **Methods**: Each type of record should have specific methods. For example, `User` records should have a `greet` method, and `Product` records should have a `displayPrice` method.

4. **Edge Cases**: Handle cases where the type is not recognized by returning a generic `Record` type with a default method.

\#\#\# Example Usage

\fbox{ \includegraphics[width=0.6\linewidth]{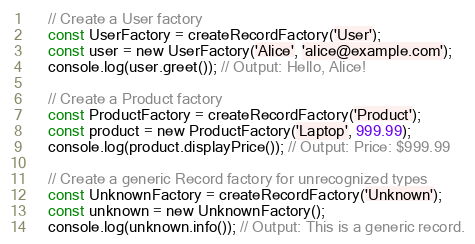}}
...
    \end{tcolorbox}

\begin{tcolorbox}[title=Response, solution]

\begin{minted}{markdown}
### Explanation and Code Review

The provided solution effectively meets the requirements by creating a factory 
function that generates constructor functions for different types of database records. 
Here's a detailed breakdown of the solution and some suggestions for improvement:

#### Key Components

1. **Factory Function (`createRecordFactory`)**:
   - This function takes a `type` parameter and returns a constructor function for the specified type.
   - It uses a switch statement to determine which constructor to return.

2. **Constructor Functions**:
   - **User**: Initializes with `name` and `email` properties and has a `greet` method.
   - **Product**: Initializes with `name` and `price` properties and has a `displayPrice` method.
   - **Record**: A generic constructor for unrecognized types with an `info` method.

3. **Prototypes**:
   - Methods are added to the prototypes of the constructors to ensure they are shared across 
   all instances, which is a good practice for memory efficiency.

4. **Edge Cases**:
   - If the `type` is not recognized, the factory returns a `Record` constructor, providing 
   a default method.

\end{minted}
\end{tcolorbox}
\end{tcolorbox}
\end{center}

\caption{Cross-modal example 3 from \instruct{}.}
\label{fig:inst_exmaple3}
\end{figure}

%% file: examples/instruct_4.tex
\begin{figure}[h]

\begin{center}
    \begin{tcolorbox}[title=\instruct{} Diagram~Example 1, showcase] 
    \begin{tcolorbox}[title=Problem, problem]

You are tasked with creating a TypeScript class method that fetches and processes data from a server using the provided code snippet as a reference. Your task is to implement the `jsonMutationDataFetcher` function and ensure that the `fetchData` method correctly handles the asynchronous data retrieval and processing.

The `fetchData` method is a private method of a TypeScript class and is decorated with `@autobind`. It takes a `filter` parameter of type `TFilter` and returns a `Promise` of type `TData`. The `jsonMutationDataFetcher` function is used to fetch and process the data. It takes three parameters: `mutationName` of type `string`, `filter` of type `TFilter` or its formatted version, and `formatter` function.

Your task is to implement the `jsonMutationDataFetcher` function and ensure that the `fetchData` method correctly calls it with the appropriate parameters. Additionally, you need to handle the asynchronous nature of data fetching and processing.
...

\fbox{ \includegraphics[width=0.95\linewidth]{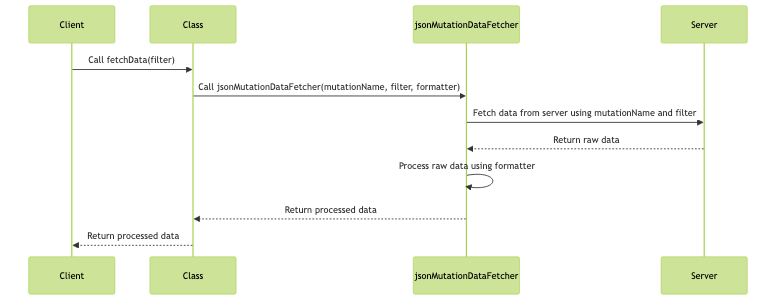}}

    \end{tcolorbox}

\begin{tcolorbox}[title=Response, solution]

\begin{minted}{typescript}
// Solution for jsonMutationDataFetcher function
async function jsonMutationDataFetcher<TRaw, TData>(
  mutationName: string,
  filter: TFilter,
  formatter: (data: TRaw) => TData
): Promise<TData> {
  // Fetch data from the server using the mutationName and filter
  const rawData = await fetchDataFromServer(mutationName, filter);
  // Process the raw data using the provided formatter function
  const processedData = formatter(rawData);
  return processedData;
}

// Solution for fetchData method
@autobind
private async fetchData(filter: TFilter): Promise<TData> {
  return jsonMutationDataFetcher<TRaw, TData>(
    this.props.mutationName,
    this.props.filterFormatter ? this.props.filterFormatter(filter) : filter,
    this.props.formatter
  );
}
...
\end{minted}
In the solution, the `jsonMutationDataFetcher` function is implemented to handle the data fetching and processing. It takes the `mutationName`, `filter`, and `formatter` as parameters and returns a `Promise` of type `TData`.
...

\end{tcolorbox}
\end{tcolorbox}
\end{center}

\caption{Diagram example 1 from \instruct{}.}
\label{fig:inst_example4}
\end{figure}

%% file: examples/instruct_5.tex
\begin{figure}[h]

\begin{center}
    \begin{tcolorbox}[title=\instruct{} Diagram~Example 2, showcase] 
    \begin{tcolorbox}[title=Problem, problem]
You are given a list of software applications installed on a user's computer. The list contains the names of the applications, but there are some errors in the names. The errors include misspellings, incorrect capitalization, and missing characters. Your task is to create a Python function that takes the list of applications as input and returns a corrected list with the errors fixed.

The function signature is:
```python
def fix\_application\_names(apps: list) -> list:
    pass
```
...

\fbox{ \includegraphics[width=0.6\linewidth]{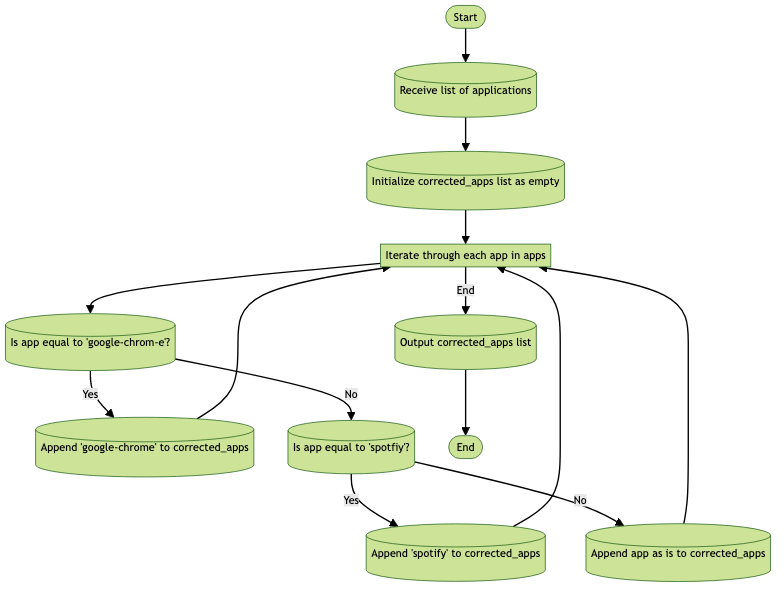}}

    \end{tcolorbox}

\begin{tcolorbox}[title=Response, solution]

\begin{minted}{python}
def fix_application_names(apps: list) -> list:
    corrected_apps = []
    for app in apps:
        if app == "google-chrom-e":
            corrected_apps.append("google-chrome")
        elif app == "spotfiy":
            corrected_apps.append("spotify")
        else:
            corrected_apps.append(app)
    return corrected_apps
\end{minted}
The `fix\_application\_names` function iterates through the input list of applications and corrects the errors by replacing the incorrect names with the correct ones. Specifically, it handles the corrections for "google-chrom-e" and "spotfiy" and appends the corrected names to a new list, which is then returned as the output. The detailed flow of this process can be found in the provided diagram.

\end{tcolorbox}
\end{tcolorbox}
\end{center}

\caption{Diagram example 2 from \instruct{}.}
\label{fig:inst_example5}
\end{figure}

%% file: examples/instruct_6.tex
\begin{figure}[h]

\begin{center}
    \begin{tcolorbox}[title=\instruct{} Diagram~Example 3, showcase] 
    \begin{tcolorbox}[title=Problem, problem]
Your task is to implement the `DebugContextHolder` interface and provide a concrete implementation of the `getDebugContext` method. Additionally, you need to create a sample usage scenario to demonstrate the functionality of the debugging context holder.

Your implementation should adhere to the following requirements:
- Implement the `DebugContextHolder` interface with the `getDebugContext` method.
- Create a concrete implementation of the `DebugContextHolder` interface.
- Demonstrate the usage of the debugging context holder in a multi-threaded environment.

You can use the given code snippet as a reference for the interface definition and method signature. The class structure and relationships are detailed in the provided diagram.
...

\fbox{ \includegraphics[width=0.5\linewidth]{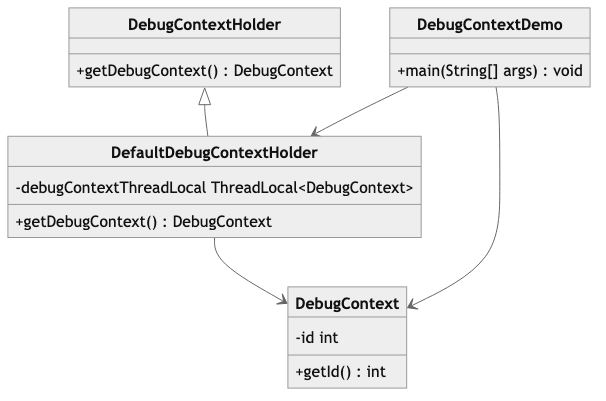}}

    \end{tcolorbox}

\begin{tcolorbox}[title=Response, solution]

\begin{minted}{java}
import java.util.concurrent.atomic.AtomicInteger;

/**
 * Debug context holder interface. By default debugging context stores in ThreadLocal 
 variable {@link DefaultDebugContextHolder}

public interface DebugContextHolder {

    /**
     * Get debug context.
     *
     * @return DebugContext
     */
    DebugContext getDebugContext();
}

/**
 * Concrete implementation of DebugContextHolder using ThreadLocal for storing debugging context.
 */
public class DefaultDebugContextHolder implements DebugContextHolder {

    private static final ThreadLocal<DebugContext> debugContextThreadLocal = 
    new ThreadLocal<DebugContext>() {
        @Override
        protected DebugContext initialValue() {
            return new DebugContext();
        }
    };
...
\end{minted}
In this solution, we first define the `DebugContextHolder` interface with the `getDebugContext` method. Then, we provide a concrete implementation, `DefaultDebugContextHolder`, which uses a `ThreadLocal` variable to store the debugging context. We also demonstrate the usage of the debugging context holder in a multi-threaded environment by creating and starting multiple threads to access the debugging context. Finally, we define the `DebugContext` class to represent the debugging context with a unique ID for each instance. The class structure and relationships are as detailed in the provided diagram.

\end{tcolorbox}
\end{tcolorbox}
\end{center}

\caption{Diagram example 3 from \instruct{}.}
\label{fig:inst_example6}
\end{figure}

%% file: examples/instruct_7.tex
\begin{figure}[h]

\begin{center}
    \begin{tcolorbox}[title=\instruct{} Diagram~Example 4, showcase] 
    \begin{tcolorbox}[title=Problem, problem]
You are tasked with implementing a multi-modal C++ class that encapsulates a set of standard functions and global variables for LLVM (Low Level Virtual Machine) operations. The provided code snippet is a part of a namespace `asllvm::detail` containing two structs: `StandardFunctions` and `GlobalVariables`. The `StandardFunctions` struct holds references to various LLVM function callees, while the `GlobalVariables` struct is incomplete in the given snippet.

...

\fbox{ \includegraphics[width=0.5\linewidth]{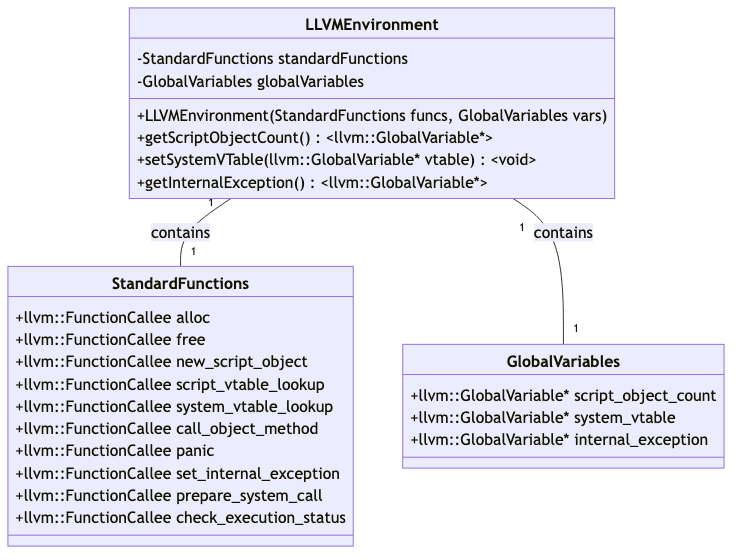}}

    \end{tcolorbox}

\begin{tcolorbox}[title=Response, solution]

\begin{minted}{cpp}
#include <llvm/IR/Function.h>
#include <llvm/IR/GlobalVariable.h>

namespace asllvm::detail{
    struct StandardFunctions{
        llvm::FunctionCallee alloc;
        llvm::FunctionCallee free;
        llvm::FunctionCallee new_script_object;
        llvm::FunctionCallee script_vtable_lookup;
        ...
    };
    struct GlobalVariables{
        llvm::GlobalVariable* script_object_count;
        ...
    };

    class LLVMEnvironment{
    private:
        StandardFunctions standardFunctions;
        GlobalVariables globalVariables;

    public:
        // Constructor to initialize standard functions and global variables
        LLVMEnvironment(StandardFunctions funcs, GlobalVariables vars)
            : standardFunctions(funcs), globalVariables(vars) {}

        // Method to get the script object count
        llvm::GlobalVariable* getScriptObjectCount() const{
            return globalVariables.script_object_count;
        }
        // Method to set the system vtable
        void setSystemVTable(llvm::GlobalVariable* vtable){
            globalVariables.system_vtable = vtable;
        }

...
\end{minted}
In this solution, the `GlobalVariables` struct is completed by adding member variables for `script\_object\_count`, `system\_vtable`, and `internal\_exception`. Additionally, a C++ class `LLVMEnvironment` is defined to encapsulate the `StandardFunctions` and `GlobalVariables` structs. The class provides methods to access and manipulate the standard functions and global variables, demonstrating encapsulation and object-oriented programming principles. The relationships and specific details of the classes and their methods are consistent with the provided diagram.

\end{tcolorbox}
\end{tcolorbox}
\end{center}

\caption{Diagram example 4 from \instruct{}.}
\label{fig:inst_example7}
\end{figure}

%% file: examples/instruct_8.tex
\begin{figure}[h]

\begin{center}
    \begin{tcolorbox}[title=\instruct{} Diagram~Example 5, showcase] 
    \begin{tcolorbox}[title=Problem, problem]
You are tasked with creating a multi-modal Python class that can serialize and deserialize data for a pet adoption application. The application needs to handle various attributes of pets, such as their name, age, breed, and availability for adoption. Your task is to implement a serializer class that can convert pet objects into JSON format for storage and transmission, as well as deserialize JSON data back into pet objects.

You are provided with a basic serializer class, `PetSerializer`, which needs to be extended and customized to handle the specific attributes of pets. The `PetSerializer` class should include methods for serializing and deserializing pet objects, ensuring that the data is accurately represented in JSON format and can be reconstructed into pet objects.

...

\fbox{ \includegraphics[width=0.7\linewidth]{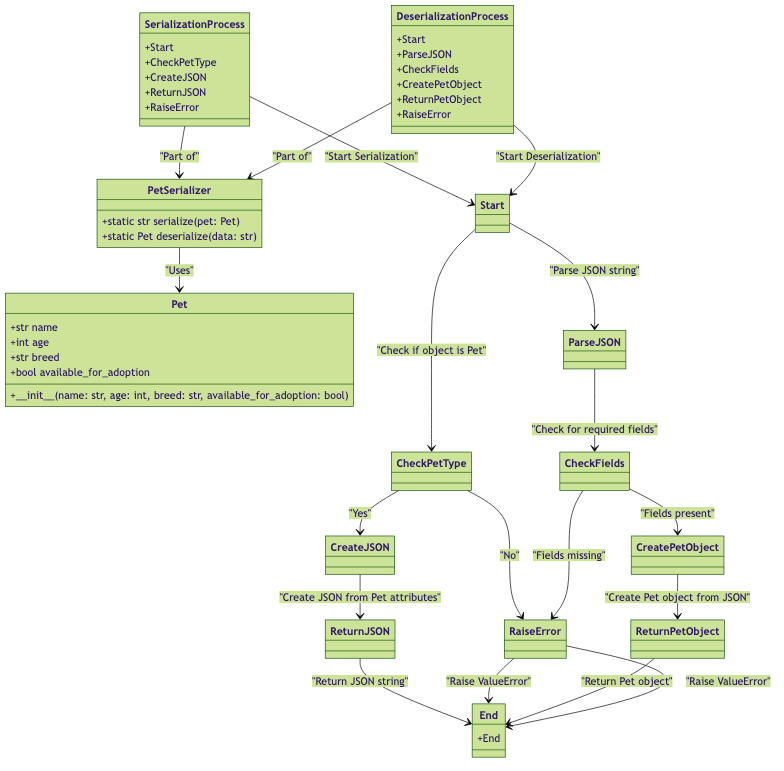}}

    \end{tcolorbox}

\begin{tcolorbox}[title=Response, solution]

\begin{minted}{python}
import json

class Pet:
    def __init__(self, name, age, breed, available_for_adoption):
        self.name = name
        self.age = age
        self.breed = breed
        self.available_for_adoption = available_for_adoption

class PetSerializer:
    @staticmethod
    def serialize(pet):
        # Check if the object is of type Pet
        if isinstance(pet, Pet):
            # Create JSON representation of the pet's attributes
            return json.dumps({
                "name": pet.name,
                "age": pet.age,
                "breed": pet.breed,
                "available_for_adoption": pet.available_for_adoption
            })
        else:
            # Raise ValueError if the object is not of type Pet
            raise ValueError("Object is not of type Pet")
...
\end{minted}
This solution follows the processes outlined in the diagram, ensuring that the `PetSerializer` class correctly serializes and deserializes `Pet` objects while handling errors appropriately.

\end{tcolorbox}
\end{tcolorbox}
\end{center}

\caption{Diagram example 5 from \instruct{}.}
\label{fig:inst_example8}
\end{figure}

%% file: examples/bench_1.tex
\begin{figure}[h]
\begin{center}
    \begin{tcolorbox}[title=\benchmark{}~Example 1, showcase] 
    \begin{tcolorbox}[title=Problem, problem]
Use \textcolor{redlight}{\bf C\#} to complete the task as described in the diagram:
\textcolor{purplecol}{Design a Crop class in a virtual farm management system.}

        \includegraphics[width=0.9\linewidth]{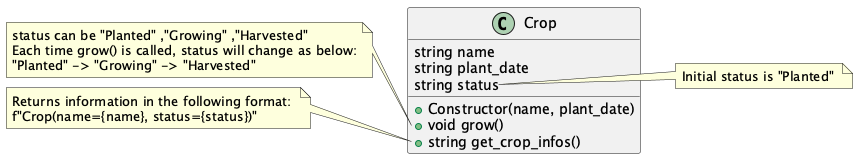}
    \end{tcolorbox}

\begin{tcolorbox}[title=Canonical Solution, solution]
\begin{minted}{csharp}
using System;
using System.Diagnostics;

public class Crop
{
    public string name { get; set; }
    public string plant_date { get; set; }
    public string status { get; set; } // status can be "Planted", "Growing", "Harvested"

    public Crop(string name, string plant_date)
    {
        this.name = name;
        this.plant_date = plant_date;
        this.status = "Planted";
    }
    public void grow()
    {
        if (this.status == "Planted")
        {
            this.status = "Growing";
        }
        else if (this.status == "Growing")
        {
            this.status = "Harvested";
        }
    }
    public string get_crop_infos()
    {
        return $"Crop(name={this.name}, status={this.status})";
    }
}
\end{minted}
\end{tcolorbox}
\end{tcolorbox}
\end{center}
\caption{Example 1 from \benchmark{}.}
\label{fig:bench_exmaple1}
\end{figure}

%% file: examples/bench_2.tex
\begin{figure}[h]
\begin{center}
    \begin{tcolorbox}[title=\benchmark{}~Example 2, showcase] 
    \begin{tcolorbox}[title=Problem, problem]
        Use \textcolor{redlight}{\bf Python} to complete the task as described in the diagram:
\textcolor{purplecol}{Design \textbf{Crop(abstract), Wheat and Corn class} in a virtual farm management system.}\\
        \includegraphics[width=0.9\linewidth]{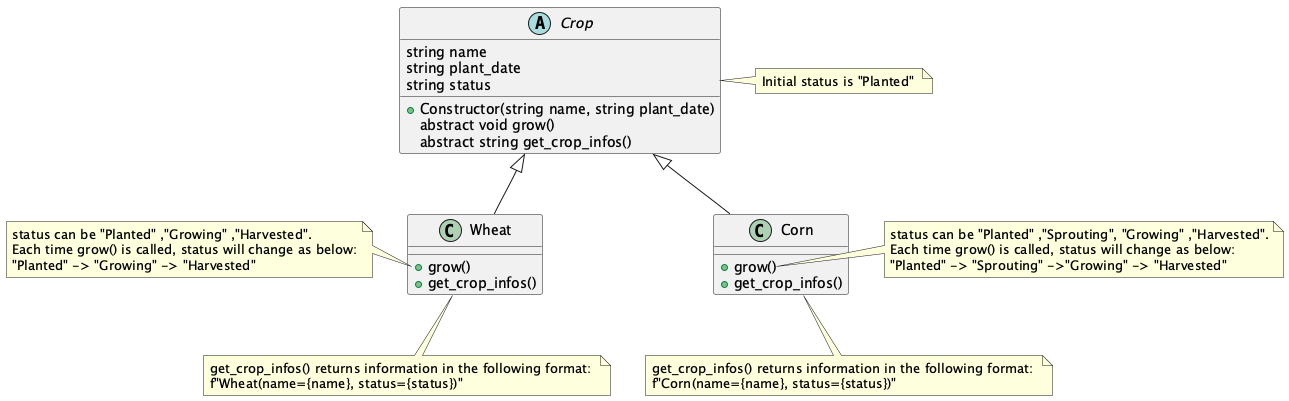}
    \end{tcolorbox}

\begin{tcolorbox}[title=Canonical Solution, solution]
\begin{minted}{python}
from abc import ABC, abstractmethod
class Crop(ABC):
    def __init__(self, name, plant_date):
        self.name = name
        self.plant_date = plant_date
        self.status = "Planted"
    @abstractmethod
    def grow(self):
        pass
    @abstractmethod
    def get_crop_infos(self):
        pass 
        
class Wheat(Crop):
    def grow(self):
        if self.status == "Planted":
            self.status = "Growing"
        elif self.status == "Growing":
            self.status = "Harvested"
    def get_crop_infos(self):      
        return f"Wheat(name={self.name}, status={self.status})"
        
class Corn(Crop):
    def grow(self):
        if self.status == "Planted":
            self.status = "Sprouting"
        elif self.status == "Sprouting":
            self.status = "Growing"
        elif self.status == "Growing":
            self.status = "Harvested"
    def get_crop_infos(self):      
        return f"Corn(name={self.name}, status={self.status})
\end{minted}
\end{tcolorbox}
\end{tcolorbox}
\end{center}
\caption{Example 2 from \benchmark{}.}
\label{fig:bench_exmaple2}
\end{figure}

%% file: examples/bench_3.tex
\begin{figure}[h]
\begin{center}
    \begin{tcolorbox}[title=\benchmark{}~Example 3, showcase] 
    \begin{tcolorbox}[title=Problem, problem]
Use \textcolor{redlight}{\bf Kotlin} to complete the task as described in the diagram:
\textcolor{purplecol}{Design GameCharacter(abstract), Warrior, Mage, GameWorld class and a CharacterFactory class} to create characters of type Warrior or Mage in a VR game world where users can create characters, explore the world, and interact with other characters.

        \includegraphics[width=0.9\linewidth]{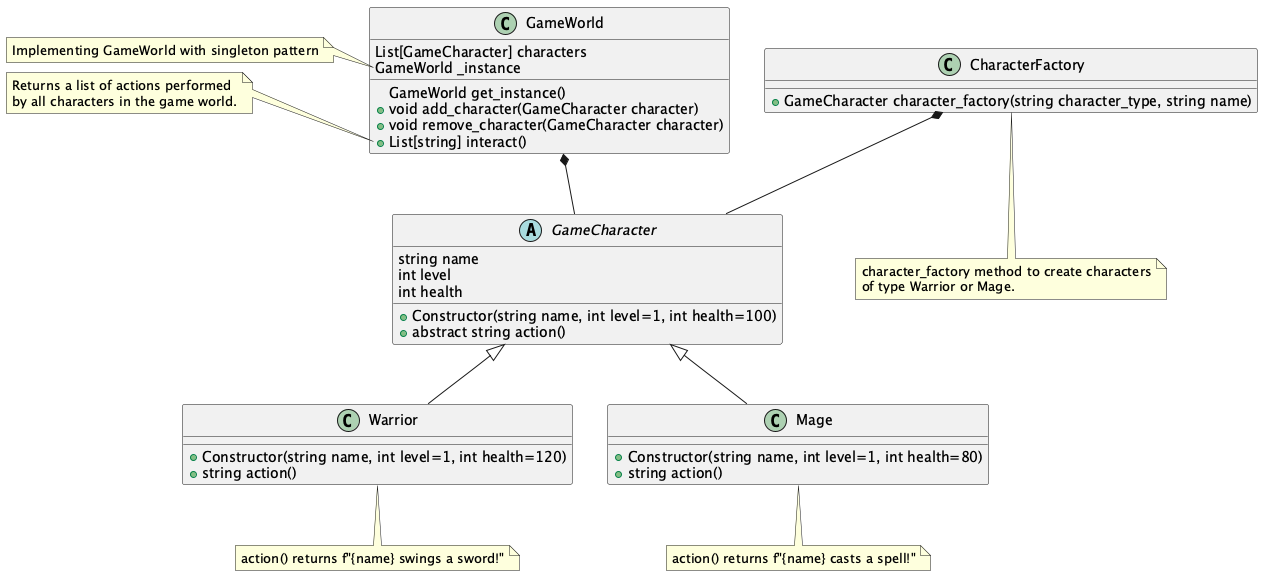}
    \end{tcolorbox}

\begin{tcolorbox}[title=Canonical Solution, solution]
\begin{minted}{kotlin}
abstract class GameCharacter(val name: String, var level: Int = 1, var health: Int = 100) {
    abstract fun action(): String
}
class Warrior(name: String, level: Int = 1, health: Int = 120) : GameCharacter(name, level, health) {
    override fun action(): String {
        return "$name swings a sword!"
    }
}
class Mage(name: String, level: Int = 1, health: Int = 80) : GameCharacter(name, level, health) {
    override fun action(): String {
        return "$name casts a spell!"
    }
}
class GameWorld private constructor() {
    val characters: MutableList<GameCharacter> = mutableListOf()

    fun add_character(character: GameCharacter) {
        characters.add(character)
    }
    fun remove_character(character: GameCharacter) {
        characters.remove(character)
    }
    fun interact(): List<String> {
        return characters.map { it.action() }
    }
    companion object {
        @Volatile
        private var instance: GameWorld? = null

        fun get_instance(): GameWorld =
            instance ?: synchronized(this) {
                instance ?: GameWorld().also { instance = it }
            }
    }
}
class CharacterFactory {
    fun character_factory(characterType: String, name: String): GameCharacter {
        return when (characterType) {
            "Warrior" -> Warrior(name)
            "Mage" -> Mage(name)
            else -> throw IllegalArgumentException("Unknown character type")
        }
    }
}
\end{minted}
\end{tcolorbox}
\end{tcolorbox}
\end{center}

\caption{Example 3 from \benchmark{}.}
\label{fig:bench_exmaple3}
\end{figure}

%% file: examples/bench_4.tex
\begin{figure}[h]
\begin{center}
    \begin{tcolorbox}[title=\benchmark{}~Example 4, showcase] 
    \begin{tcolorbox}[title=Problem, problem]
Use Use \textcolor{redlight}{\bf Swift} to complete the task as described in the diagram:
\textcolor{purplecol}{Design an AI assistant system} to manage different types of AI assistants and handle various user requests.
You need Design AIAssistant(abstract), WeatherAssistant, NewsAssistant and HealthAssistant classes to handle different types of AI assistants, and you need to design a User class to manage different types of AI assistants. 
        \includegraphics[width=0.9\linewidth]{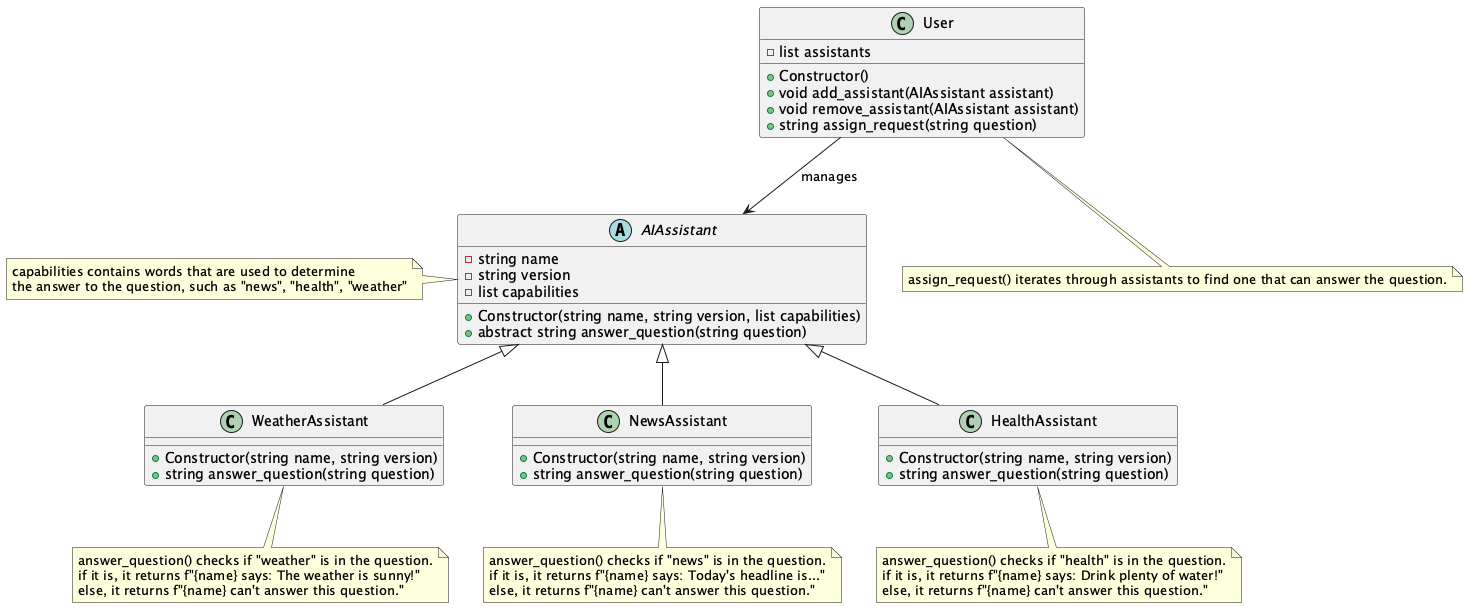}
    \end{tcolorbox}
   
\begin{tcolorbox}[title=Canonical Solution, solution]
\begin{minted}{ruby}
class AIAssistant
  attr_reader :name, :version, :capabilities

  def initialize(name, version, capabilities)
    @name = name
    @version = version
    @capabilities = capabilities
  end

  def answer_question(question)
    raise NotImplementedError, "Subclasses must implement the answer_question method"
  end
end

class WeatherAssistant < AIAssistant
  def initialize(name, version)
    super(name, version, ["weather"])
  end

  def answer_question(question)
    if question.downcase.include?("weather")
      "#{@name} says: The weather is sunny!"
    else
      "#{@name} can't answer this question."
    end
  end
end
...
class User
  def initialize
    @assistants = []
  end

  def add_assistant(assistant)
    @assistants << assistant
  end

  def remove_assistant(assistant)
    @assistants.delete(assistant)
  end

  def assign_request(question)
    @assistants.each do |assistant|
      response = assistant.answer_question(question)
      return response unless response.include?("can't answer")
    end
    "None of the assistants can answer this question."
  end
end
\end{minted}
\end{tcolorbox}
\end{tcolorbox}
\end{center}

\caption{Example 4 from \benchmark{}.}
\label{fig:bench_exmaple4}
\end{figure}

%% file: examples/bench_5.tex
\begin{figure}[h]
\begin{center}
    \begin{tcolorbox}[title=\benchmark{}~Example 5, showcase] 
    \begin{tcolorbox}[title=Problem, problem]

Use \textcolor{redlight}{\bf PHP} to complete the task as described in the diagram:
Write a function `function calculate\_number(int \$number): int ` to calculate the number.

        \includegraphics[width=0.4\linewidth]{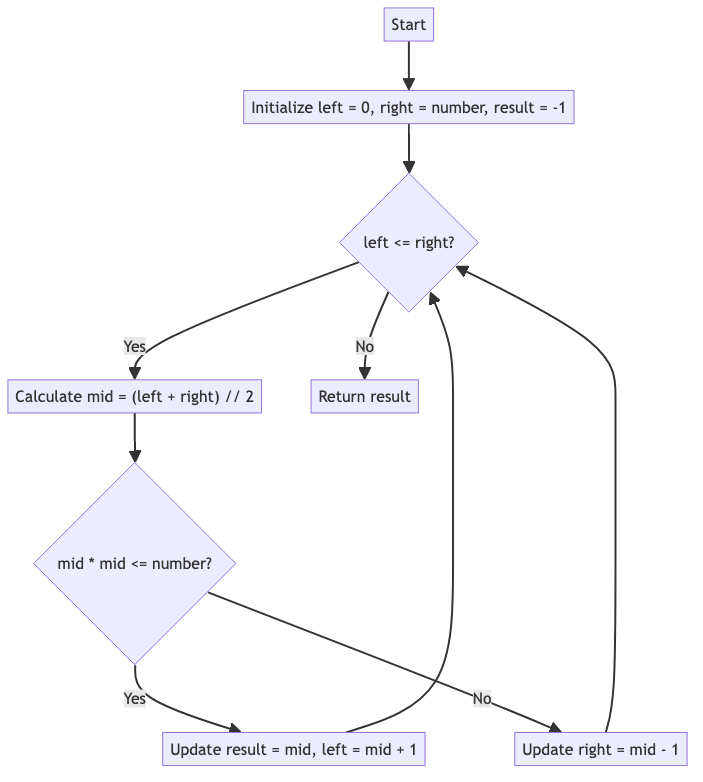}
    \end{tcolorbox}

\begin{tcolorbox}[title=Canonical Solution, solution]
\begin{minted}{php}
<?php

function calculate_number(int $number): int {
    $left = 0;
    $right = $number;
    $result = -1;
    
    while ($left <= $right) {
        $mid = intdiv($left + $right, 2);
        if ($mid * $mid <= $number) {
            $result = $mid;
            $left = $mid + 1;
        } else {
            $right = $mid - 1;
        }
    }  
    return $result;
}
?>
\end{minted}
\end{tcolorbox}
\end{tcolorbox}
\end{center}

\caption{Example 5 from \benchmark{}.}
\label{fig:bench_exmaple5}
\end{figure}

%% file: examples/bench_6.tex
\begin{figure}[h]
\begin{center}
    \begin{tcolorbox}[title=\benchmark{}~Example 6, showcase] 
    \begin{tcolorbox}[title=Problem, problem]
Use \textcolor{redlight}{\bf Scala} to complete the task as described in the diagram:
Write a function `def find\_word(paragraph: String, banned: List[String]): Option[String]` to find a special word in a paragraph.

        \includegraphics[width=0.4\linewidth]{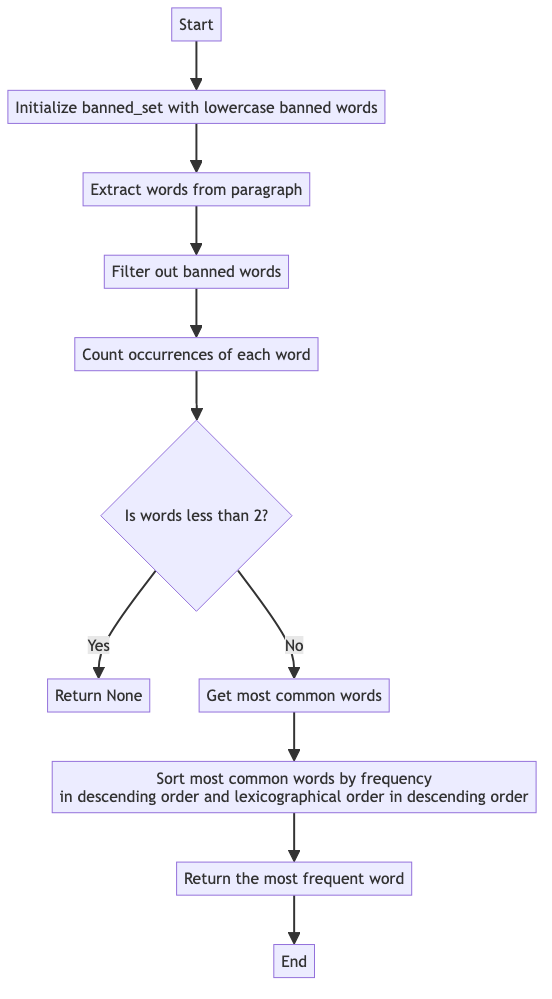}
    \end{tcolorbox}

\begin{tcolorbox}[title=Canonical Solution, solution]
\begin{minted}{scala}
import scala.collection.mutable
import scala.math.Numeric.IntIsIntegral
import scala.math.Ordering.Implicits._
import scala.util.matching.Regex

object Main {
  def find_word(paragraph: String, banned: List[String]): Option[String] = {
    val bannedSet = banned.map(_.toLowerCase).toSet
    val words = new Regex("\\w+").findAllIn(paragraph.toLowerCase).toList
    val filteredWords = words.filterNot(bannedSet.contains)
    val wordCounts = mutable.Map[String, Int]().withDefaultValue(0)

    filteredWords.foreach(word => wordCounts(word) += 1)

    if (wordCounts.size < 2) {
      None
    } else {
      val mostCommon = wordCounts.toList.sortBy { case (word, count) => (-count, word.map(-_.toInt)) }
      Some(mostCommon.head._1)
    }
  }
}
\end{minted}
\end{tcolorbox}
\end{tcolorbox}
\end{center}

\caption{Example 6 from \benchmark{}.}
\label{fig:bench_exmaple6}
\end{figure}

%% file: examples/bench_7.tex
\begin{figure}[h]
\begin{center}
    \begin{tcolorbox}[title=\benchmark{}~Example 7, showcase] 
    \begin{tcolorbox}[title=Problem, problem]
Use \textcolor{redlight}{\bf JavaScript} to complete the task as described in the diagram:
Write a function `function navigate\_complex\_road(road\_conditions)` to solve the following problem:

The function should analyze the sequence of road conditions and decide on the appropriate actions to ensure safe and efficient navigation. 

Args:
road\_conditions (List[str]): A list of strings representing the sequence of road conditions the vehicle will encounter.

Returns:
List[str]: A list of strings representing the actions the vehicle should take to navigate through the given road conditions.

        \includegraphics[width=0.9\linewidth]{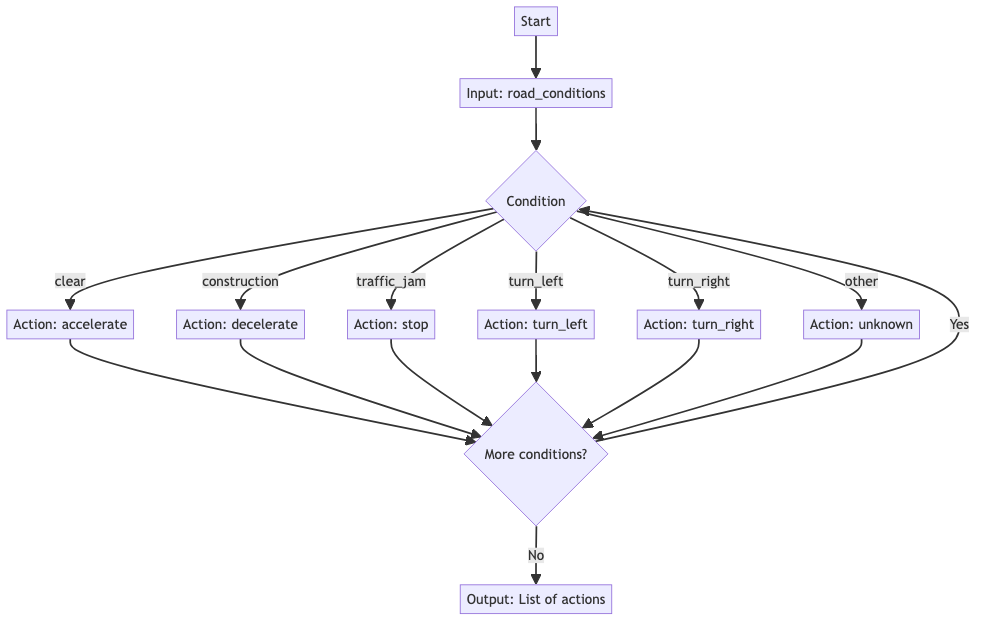}
    \end{tcolorbox}

\begin{tcolorbox}[title=Canonical Solution, solution]
\begin{minted}{javascript}
function navigate_complex_road(road_conditions) {
    const actions = [];
    for (let condition of road_conditions) {
        switch (condition) {
            case "clear":
                actions.push("accelerate");
                break;
            case "construction":
                actions.push("decelerate");
                break;
            case "traffic_jam":
                actions.push("stop");
                break;
            case "turn_left":
                actions.push("turn_left");
                break;
            case "turn_right":
                actions.push("turn_right");
                break;
            default:
                actions.push("unknown");
        }
    }
    return actions;
}
\end{minted}
\end{tcolorbox}
\end{tcolorbox}
\end{center}

\caption{Example 7 from \benchmark{}.}
\label{fig:bench_exmaple7}
\end{figure}

%% file: examples/bench_8.tex
\begin{figure}[h]
\begin{center}
    \begin{tcolorbox}[title=\benchmark{}~Example 8, showcase] 
    \begin{tcolorbox}[title=Problem, problem]
Use \textcolor{redlight}{\bf CPP} to complete the task as described in the diagram:

Write a function `unordered\_map<string, vector<string>> virtual\_meeting\_assistant(const string\& operation, const unordered\_map<string, string>\& data)` that performs different operations based on the given operation type and data. The function should manage the following tasks:
    - Schedule a Meeting
    - Add Participants
    - Record Meeting Minutes
    - Generate a Summary

Args:

    - operation (str): The type of operation to perform. It can be one of the following:
        ...

    - data (dict): A dictionary containing the necessary data for the operation.
        ...

Return:
    For "generate\_summary", return a dictionary with a key "summary" and a list of key points as the value. For other operations, return an empty dictionary.
    
        \includegraphics[width=0.6\linewidth]{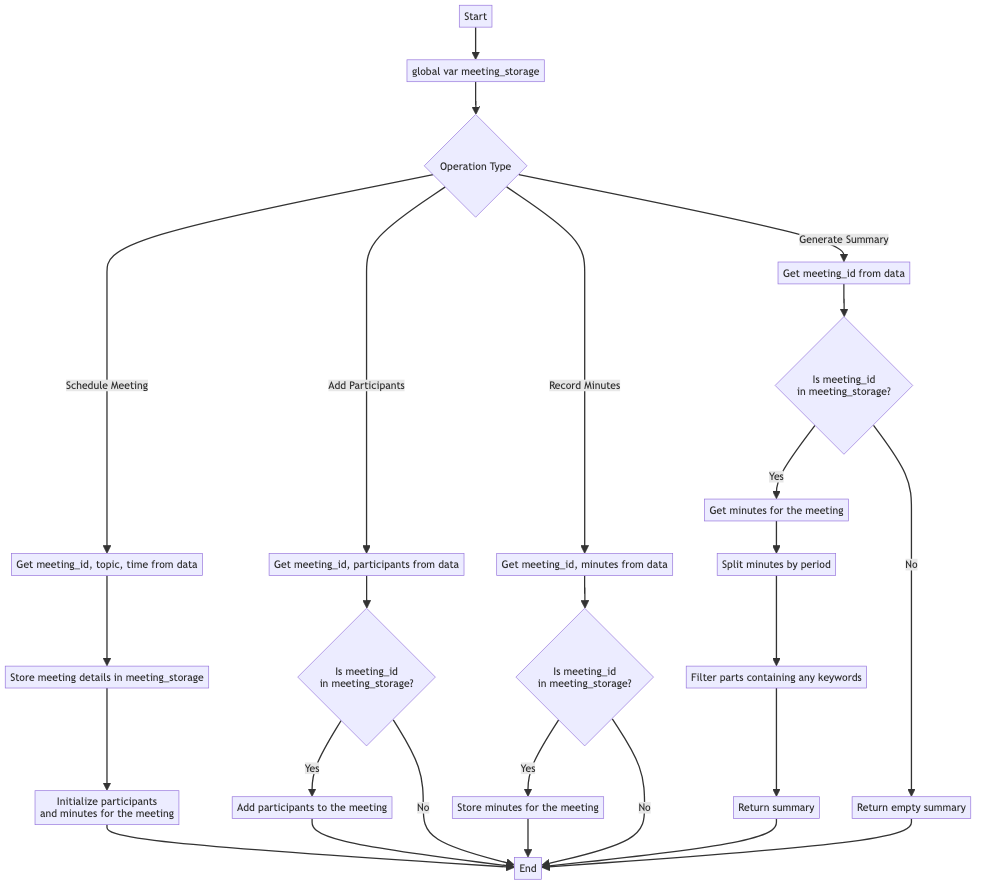}
    \end{tcolorbox}

\begin{tcolorbox}[title=Canonical Solution, solution]
\begin{minted}{cpp}
#include <iostream>
...
using namespace std;
struct MeetingStorage {
    unordered_map<int, pair<string, string>> meetings;
    unordered_map<int, vector<string>> participants;   
    unordered_map<int, string> minutes;                
};
MeetingStorage meeting_storage;

unordered_map<string, vector<string>> virtual_meeting_assistant(const string& operation, 
const unordered_map<string, string>& data) {
    if (operation == "schedule") {
        ...
    } 
    else if (operation == "add_participant") {
        int meeting_id = stoi(data.at("meeting_id"));
        if (meeting_storage.participants.find(meeting_id) != meeting_storage.participants.end()) {
            ...
        }
    } 
    else if (operation == "record_minutes") {
        int meeting_id = stoi(data.at("meeting_id"));
        string minutes = data.at("minutes");
        ...
    } 
    else if (operation == "generate_summary") {
        int meeting_id = stoi(data.at("meeting_id"));
        ...

        return {{"summary", key_points}};
    }

    return {};
}
\end{minted}
\end{tcolorbox}
\end{tcolorbox}
\end{center}

\caption{Example 8 from \benchmark{}.}
\label{fig:bench_exmaple8}
\end{figure}

%% file: neurips_2025.bbl
\begin{thebibliography}{100}

\bibitem{phi_3}
Marah Abdin, Sam~Ade Jacobs, Ammar~Ahmad Awan, Jyoti Aneja, Ahmed Awadallah, Hany Awadalla, Nguyen Bach, Amit Bahree, Arash Bakhtiari, Harkirat Behl, et~al.
\newblock Phi-3 technical report: A highly capable language model locally on your phone.
\newblock {\em arXiv preprint arXiv:2404.14219}, 2024.

\bibitem{llama3}
Meta AI.
\newblock Introducing meta llama 3: The most capable openly available llm to date.
\newblock \url{https://ai.meta.com/blog/meta-llama-3/}, apr 2024.

\bibitem{aider}
{Aider Team}.
\newblock Aider llm leaderboards.

\bibitem{santacoder}
Loubna~Ben Allal, Raymond Li, Denis Kocetkov, Chenghao Mou, Christopher Akiki, Carlos~Munoz Ferrandis, Niklas Muennighoff, Mayank Mishra, Alex Gu, Manan Dey, et~al.
\newblock {SantaCoder}: Don't reach for the stars!
\newblock {\em arXiv preprint arXiv:2301.03988}, 2023.

\bibitem{claude3}
Anthropic.
\newblock The claude 3 model family: Opus, sonnet, haiku.
\newblock Technical report, Anthropic, 2024.

\bibitem{vqa}
Stanislaw Antol, Aishwarya Agrawal, Jiasen Lu, Margaret Mitchell, Dhruv Batra, C~Lawrence Zitnick, and Devi Parikh.
\newblock Vqa: Visual question answering.
\newblock In {\em Proceedings of the IEEE international conference on computer vision}, pages 2425--2433, 2015.

\bibitem{mbxp}
Ben Athiwaratkun, Sanjay~Krishna Gouda, Zijian Wang, Xiaopeng Li, Yuchen Tian, Ming Tan, Wasi~Uddin Ahmad, Shiqi Wang, Qing Sun, Mingyue Shang, Sujan~Kumar Gonugondla, Hantian Ding, Varun Kumar, Nathan Fulton, Arash Farahani, Siddhartha Jain, Robert Giaquinto, Haifeng Qian, Murali~Krishna Ramanathan, and Ramesh Nallapati.
\newblock Multi-lingual evaluation of code generation models.
\newblock In {\em The Eleventh International Conference on Learning Representations, {ICLR} 2023, Kigali, Rwanda, May 1-5, 2023}. OpenReview.net, 2023.

\bibitem{mbpp}
Jacob Austin, Augustus Odena, Maxwell Nye, Maarten Bosma, Henryk Michalewski, David Dohan, Ellen Jiang, Carrie Cai, Michael Terry, Quoc Le, et~al.
\newblock Program synthesis with large language models.
\newblock {\em arXiv preprint arXiv:2108.07732}, 2021.

\bibitem{Qwen}
Jinze Bai, Shuai Bai, Yunfei Chu, Zeyu Cui, Kai Dang, Xiaodong Deng, Yang Fan, Wenbin Ge, Yu~Han, Fei Huang, Binyuan Hui, Luo Ji, Mei Li, Junyang Lin, Runji Lin, Dayiheng Liu, Gao Liu, Chengqiang Lu, Keming Lu, Jianxin Ma, Rui Men, Xingzhang Ren, Xuancheng Ren, Chuanqi Tan, Sinan Tan, Jianhong Tu, Peng Wang, Shijie Wang, Wei Wang, Shengguang Wu, Benfeng Xu, Jin Xu, An~Yang, Hao Yang, Jian Yang, Shusheng Yang, Yang Yao, Bowen Yu, Hongyi Yuan, Zheng Yuan, Jianwei Zhang, Xingxuan Zhang, Yichang Zhang, Zhenru Zhang, Chang Zhou, Jingren Zhou, Xiaohuan Zhou, and Tianhang Zhu.
\newblock Qwen technical report.
\newblock {\em arXiv preprint arXiv:2309.16609}, abs/2309.16609, 2023.

\bibitem{Qwen_VL}
Jinze Bai, Shuai Bai, Shusheng Yang, Shijie Wang, Sinan Tan, Peng Wang, Junyang Lin, Chang Zhou, and Jingren Zhou.
\newblock Qwen-vl: A versatile vision-language model for understanding, localization, text reading, and beyond.
\newblock {\em arXiv preprint arXiv:2308.12966}, 2023.

\bibitem{Qwen2.5-VL}
Shuai Bai, Keqin Chen, Xuejing Liu, Jialin Wang, Wenbin Ge, Sibo Song, Kai Dang, Peng Wang, Shijie Wang, Jun Tang, Humen Zhong, Yuanzhi Zhu, Mingkun Yang, Zhaohai Li, Jianqiang Wan, Pengfei Wang, Wei Ding, Zheren Fu, Yiheng Xu, Jiabo Ye, Xi~Zhang, Tianbao Xie, Zesen Cheng, Hang Zhang, Zhibo Yang, Haiyang Xu, and Junyang Lin.
\newblock Qwen2.5-vl technical report.
\newblock {\em arXiv preprint arXiv:2502.13923}, 2025.

\bibitem{codeplan}
Ramakrishna Bairi, Atharv Sonwane, Aditya Kanade, Arun Iyer, Suresh Parthasarathy, Sriram Rajamani, B~Ashok, and Shashank Shet.
\newblock Codeplan: Repository-level coding using llms and planning.
\newblock {\em Proceedings of the ACM on Software Engineering}, 1(FSE):675--698, 2024.

\bibitem{multipl_e}
Federico Cassano, John Gouwar, Daniel Nguyen, Sydney Nguyen, Luna Phipps-Costin, Donald Pinckney, Ming-Ho Yee, Yangtian Zi, Carolyn~Jane Anderson, Molly~Q Feldman, et~al.
\newblock Multipl-e: A scalable and polyglot approach to benchmarking neural code generation.
\newblock {\em IEEE Transactions on Software Engineering}, 2023.

\bibitem{mceval}
Linzheng Chai, Shukai Liu, Jian Yang, Yuwei Yin, Ke~Jin, Jiaheng Liu, Tao Sun, Ge~Zhang, Changyu Ren, Hongcheng Guo, et~al.
\newblock Mceval: Massively multilingual code evaluation.
\newblock {\em arXiv preprint arXiv:2406.07436}, 2024.

\bibitem{chen2021pass_k}
Mark Chen, Jerry Tworek, Heewoo Jun, Qiming Yuan, Henrique Ponde De~Oliveira Pinto, Jared Kaplan, Harri Edwards, Yuri Burda, Nicholas Joseph, Greg Brockman, et~al.
\newblock Evaluating large language models trained on code.
\newblock {\em arXiv preprint arXiv:2107.03374}, 2021.

\bibitem{internvl25}
Zhe Chen, Weiyun Wang, Yue Cao, Yangzhou Liu, Zhangwei Gao, Erfei Cui, Jinguo Zhu, Shenglong Ye, Hao Tian, Zhaoyang Liu, et~al.
\newblock Expanding performance boundaries of open-source multimodal models with model, data, and test-time scaling.
\newblock {\em arXiv preprint arXiv:2412.05271}, 2024.

\bibitem{gemini25}
Google Deepmind.
\newblock Gemini 2.5 pro preview: even better coding performance, 2025.

\bibitem{r2c2coder}
Ken Deng, Jiaheng Liu, He~Zhu, Congnan Liu, Jingxin Li, Jiakai Wang, Peng Zhao, Chenchen Zhang, Yanan Wu, Xueqiao Yin, et~al.
\newblock R2c2-coder: Enhancing and benchmarking real-world repository-level code completion abilities of code large language models.
\newblock {\em arXiv preprint arXiv:2406.01359}, 2024.

\bibitem{bert}
Jacob Devlin, Ming{-}Wei Chang, Kenton Lee, and Kristina Toutanova.
\newblock {BERT:} pre-training of deep bidirectional transformers for language understanding.
\newblock In {\em Proceedings of the 2019 Conference of the North American Chapter of the Association for Computational Linguistics: Human Language Technologies, {NAACL-HLT} 2019, Minneapolis, MN, USA, June 2-7, 2019, Volume 1 (Long and Short Papers)}, pages 4171--4186. Association for Computational Linguistics, 2019.

\bibitem{code_bert}
Zhangyin Feng, Daya Guo, Duyu Tang, Nan Duan, Xiaocheng Feng, Ming Gong, Linjun Shou, Bing Qin, Ting Liu, Daxin Jiang, and Ming Zhou.
\newblock Codebert: A pre-trained model for programming and natural languages.
\newblock In Trevor Cohn, Yulan He, and Yang Liu, editors, {\em Findings of the Association for Computational Linguistics: EMNLP 2020}, pages 1536--1547, Online, November 2020. Association for Computational Linguistics.

\bibitem{codegemma}
Google Gemma~Team.
\newblock Gemma: Open models based on gemini research and technology.
\newblock {\em arXiv preprint arXiv:2403.08295}, 2024.

\bibitem{vinvqa}
Yash Goyal, Tejas Khot, Douglas Summers-Stay, Dhruv Batra, and Devi Parikh.
\newblock Making the v in vqa matter: Elevating the role of image understanding in visual question answering.
\newblock In {\em Proceedings of the IEEE conference on computer vision and pattern recognition}, pages 6904--6913, 2017.

\bibitem{deepseek_coder}
Daya Guo, Qihao Zhu, Dejian Yang, Zhenda Xie, Kai Dong, Wentao Zhang, Guanting Chen, Xiao Bi, Y~Wu, YK~Li, et~al.
\newblock Deepseek-coder: When the large language model meets programming -- the rise of code intelligence.
\newblock {\em arXiv preprint arXiv:2401.14196}, 2024.

\bibitem{chartllama}
Yucheng Han, Chi Zhang, Xin Chen, Xu~Yang, Zhibin Wang, Gang Yu, Bin Fu, and Hanwang Zhang.
\newblock Chartllama: A multimodal llm for chart understanding and generation.
\newblock {\em arXiv preprint arXiv:2311.16483}, 2023.

\bibitem{he2022distribution}
Jingxuan He, Luca Beurer-Kellner, and Martin Vechev.
\newblock On distribution shift in learning-based bug detectors.
\newblock In Kamalika Chaudhuri, Stefanie Jegelka, Le~Song, Csaba Szepesvari, Gang Niu, and Sivan Sabato, editors, {\em Proceedings of the 39th International Conference on Machine Learning}, volume 162 of {\em Proceedings of Machine Learning Research}, pages 8559--8580. PMLR, 17--23 Jul 2022.

\bibitem{huang2024olympicarena}
Zhen Huang, Zengzhi Wang, Shijie Xia, Xuefeng Li, Haoyang Zou, Ruijie Xu, Run-Ze Fan, Lyumanshan Ye, Ethan Chern, Yixin Ye, et~al.
\newblock Olympicarena: Benchmarking multi-discipline cognitive reasoning for superintelligent ai.
\newblock {\em arXiv preprint arXiv:2406.12753}, 2024.

\bibitem{qwen25coder}
Binyuan Hui, Jian Yang, Zeyu Cui, Jiaxi Yang, Dayiheng Liu, Lei Zhang, Tianyu Liu, Jiajun Zhang, Bowen Yu, Kai Dang, et~al.
\newblock Qwen2. 5-coder technical report.
\newblock {\em arXiv preprint arXiv:2409.12186}, 2024.

\bibitem{gpt4o}
Aaron Hurst, Adam Lerer, Adam~P Goucher, Adam Perelman, Aditya Ramesh, Aidan Clark, AJ~Ostrow, Akila Welihinda, Alan Hayes, Alec Radford, et~al.
\newblock Gpt-4o system card.
\newblock {\em arXiv preprint arXiv:2410.21276}, 2024.

\bibitem{livecodebench}
Naman Jain, King Han, Alex Gu, Wen-Ding Li, Fanjia Yan, Tianjun Zhang, Sida Wang, Armando Solar-Lezama, Koushik Sen, and Ion Stoica.
\newblock Livecodebench: Holistic and contamination free evaluation of large language models for code.
\newblock {\em arXiv preprint arXiv:2403.07974}, 2024.

\bibitem{mistral}
Albert~Q Jiang, Alexandre Sablayrolles, Arthur Mensch, Chris Bamford, Devendra~Singh Chaplot, Diego de~las Casas, Florian Bressand, Gianna Lengyel, Guillaume Lample, Lucile Saulnier, et~al.
\newblock Mistral 7b.
\newblock {\em arXiv preprint arXiv:2310.06825}, 2023.

\bibitem{swe_bench}
Carlos~E Jimenez, John Yang, Alexander Wettig, Shunyu Yao, Kexin Pei, Ofir Press, and Karthik Narasimhan.
\newblock Swe-bench: Can language models resolve real-world github issues?
\newblock {\em arXiv preprint arXiv:2310.06770}, 2023.

\bibitem{vllm}
Woosuk Kwon, Zhuohan Li, Siyuan Zhuang, Ying Sheng, Lianmin Zheng, Cody~Hao Yu, Joseph~E. Gonzalez, Hao Zhang, and Ion Stoica.
\newblock Efficient memory management for large language model serving with pagedattention.
\newblock In {\em Proceedings of the ACM SIGOPS 29th Symposium on Operating Systems Principles}, 2023.

\bibitem{llavaonevision}
Bo~Li, Yuanhan Zhang, Dong Guo, Renrui Zhang, Feng Li, Hao Zhang, Kaichen Zhang, Yanwei Li, Ziwei Liu, and Chunyuan Li.
\newblock Llava-onevision: Easy visual task transfer.
\newblock {\em arXiv preprint arXiv:2408.03326}, 2024.

\bibitem{li2024mmcode}
Kaixin Li, Yuchen Tian, Qisheng Hu, Ziyang Luo, and Jing Ma.
\newblock Mmcode: Evaluating multi-modal code large language models with visually rich programming problems.
\newblock {\em arXiv preprint arXiv:2404.09486}, 2024.

\bibitem{starcoder}
Raymond Li, Loubna~Ben Allal, Yangtian Zi, Niklas Muennighoff, Denis Kocetkov, Chenghao Mou, Marc Marone, Christopher Akiki, Jia Li, Jenny Chim, Qian Liu, Evgenii Zheltonozhskii, Terry~Yue Zhuo, Thomas Wang, Olivier Dehaene, Mishig Davaadorj, Joel Lamy{-}Poirier, Jo{\~{a}}o Monteiro, Oleh Shliazhko, Nicolas Gontier, Nicholas Meade, Armel Zebaze, Ming{-}Ho Yee, Logesh~Kumar Umapathi, Jian Zhu, Benjamin Lipkin, Muhtasham Oblokulov, Zhiruo Wang, Rudra~Murthy V, Jason Stillerman, Siva~Sankalp Patel, Dmitry Abulkhanov, Marco Zocca, Manan Dey, Zhihan Zhang, Nour Moustafa{-}Fahmy, Urvashi Bhattacharyya, Wenhao Yu, Swayam Singh, Sasha Luccioni, Paulo Villegas, Maxim Kunakov, Fedor Zhdanov, Manuel Romero, Tony Lee, Nadav Timor, Jennifer Ding, Claire Schlesinger, Hailey Schoelkopf, Jan Ebert, Tri Dao, Mayank Mishra, Alex Gu, Jennifer Robinson, Carolyn~Jane Anderson, Brendan Dolan{-}Gavitt, Danish Contractor, Siva Reddy, Daniel Fried, Dzmitry Bahdanau, Yacine Jernite, Carlos~Mu{\~{n}}oz Ferrandis, Sean Hughes, Thomas
  Wolf, Arjun Guha, Leandro von Werra, and Harm de~Vries.
\newblock Starcoder: may the source be with you!
\newblock {\em arXiv preprint arXiv:2305.06161}, abs/2305.06161, 2023.

\bibitem{li-towards-real}
Yinghui Li, Zishan Xu, Shaoshen Chen, Haojing Huang, Yangning Li, Shirong Ma, Yong Jiang, Zhongli Li, Qingyu Zhou, Hai-Tao Zheng, and Ying Shen.
\newblock Towards real-world writing assistance: A {C}hinese character checking benchmark with faked and misspelled characters.
\newblock In Lun-Wei Ku, Andre Martins, and Vivek Srikumar, editors, {\em Proceedings of the 62nd Annual Meeting of the Association for Computational Linguistics (Volume 1: Long Papers)}, pages 8656--8668, Bangkok, Thailand, August 2024. Association for Computational Linguistics.

\bibitem{AlphaCode}
Yujia Li, David~H. Choi, Junyoung Chung, Nate Kushman, Julian Schrittwieser, R{\'{e}}mi Leblond, Tom Eccles, James Keeling, Felix Gimeno, Agustin~Dal Lago, Thomas Hubert, Peter Choy, Cyprien de~Masson~d'Autume, Igor Babuschkin, Xinyun Chen, Po{-}Sen Huang, Johannes Welbl, Sven Gowal, Alexey Cherepanov, James Molloy, Daniel~J. Mankowitz, Esme~Sutherland Robson, Pushmeet Kohli, Nando de~Freitas, Koray Kavukcuoglu, and Oriol Vinyals.
\newblock Competition-level code generation with alphacode.
\newblock {\em arXiv preprint arXiv:2203.07814}, abs/2203.07814, 2022.

\bibitem{quixbugs}
Derrick Lin, James Koppel, Angela Chen, and Armando Solar-Lezama.
\newblock Quixbugs: a multi-lingual program repair benchmark set based on the quixey challenge.
\newblock In {\em Proceedings Companion of the 2017 ACM SIGPLAN international conference on systems, programming, languages, and applications: software for humanity}, pages 55--56, 2017.

\bibitem{deepseekv3}
Aixin Liu, Bei Feng, Bing Xue, Bingxuan Wang, Bochao Wu, Chengda Lu, Chenggang Zhao, Chengqi Deng, Chenyu Zhang, Chong Ruan, et~al.
\newblock Deepseek-v3 technical report.
\newblock {\em arXiv preprint arXiv:2412.19437}, 2024.

\bibitem{llava_next}
Haotian Liu, Chunyuan Li, Yuheng Li, Bo~Li, Yuanhan Zhang, Sheng Shen, and Yong~Jae Lee.
\newblock Llava-next: Improved reasoning, ocr, and world knowledge, 2024.

\bibitem{llava}
Haotian Liu, Chunyuan Li, Qingyang Wu, and Yong~Jae Lee.
\newblock Visual instruction tuning.
\newblock {\em arXiv preprint arXiv:2304.08485}, 2023.

\bibitem{evalplus}
Jiawei Liu, Chunqiu~Steven Xia, Yuyao Wang, and Lingming Zhang.
\newblock Is your code generated by chatgpt really correct? rigorous evaluation of large language models for code generation.
\newblock {\em arXiv preprint arXiv:2305.01210}, abs/2305.01210, 2023.

\bibitem{repobench}
Tianyang Liu, Canwen Xu, and Julian McAuley.
\newblock Repobench: Benchmarking repository-level code auto-completion systems.
\newblock {\em arXiv preprint arXiv:2306.03091}, 2023.

\bibitem{logomotion}
Vivian Liu, Rubaiat~Habib Kazi, Li-Yi Wei, Matthew Fisher, Timothy Langlois, Seth Walker, and Lydia Chilton.
\newblock Logomotion: Visually grounded code generation for content-aware animation.
\newblock {\em arXiv preprint arXiv:2405.07065}, 2024.

\bibitem{visualagentbench}
Xiao Liu, Tianjie Zhang, Yu~Gu, Iat~Long Iong, Yifan Xu, Xixuan Song, Shudan Zhang, Hanyu Lai, Xinyi Liu, Hanlin Zhao, et~al.
\newblock Visualagentbench: Towards large multimodal models as visual foundation agents.
\newblock {\em arXiv preprint arXiv:2408.06327}, 2024.

\bibitem{liu2022flow2code}
Zejie Liu, Xiaoyu Hu, Deyu Zhou, Lin Li, Xu~Zhang, and Yanzheng Xiang.
\newblock Code generation from flowcharts with texts: A benchmark dataset and an approach.
\newblock In {\em Findings of the Association for Computational Linguistics: EMNLP 2022}, pages 6069--6077, 2022.

\bibitem{adamw}
Ilya Loshchilov and Frank Hutter.
\newblock Decoupled weight decay regularization.
\newblock {\em arXiv preprint arXiv:1711.05101}, 2017.

\bibitem{starcoder2}
Anton Lozhkov, Raymond Li, Loubna~Ben Allal, Federico Cassano, Joel Lamy-Poirier, Nouamane Tazi, Ao~Tang, Dmytro Pykhtar, Jiawei Liu, Yuxiang Wei, et~al.
\newblock Starcoder 2 and the stack v2: The next generation.
\newblock {\em arXiv preprint arXiv:2402.19173}, 2024.

\bibitem{wizardcoder}
Ziyang Luo, Can Xu, Pu~Zhao, Qingfeng Sun, Xiubo Geng, Wenxiang Hu, Chongyang Tao, Jing Ma, Qingwei Lin, and Daxin Jiang.
\newblock {WizardCoder}: Empowering code large language models with evol-instruct.
\newblock {\em arXiv preprint arXiv:2306.08568}, 2023.

\bibitem{llama4}
Meta.
\newblock Llama 4, leading intelligence., 2025.

\bibitem{robocodex}
Yao Mu, Junting Chen, Qinglong Zhang, Shoufa Chen, Qiaojun Yu, Chongjian Ge, Runjian Chen, Zhixuan Liang, Mengkang Hu, Chaofan Tao, et~al.
\newblock Robocodex: Multimodal code generation for robotic behavior synthesis.
\newblock {\em arXiv preprint arXiv:2402.16117}, 2024.

\bibitem{octopack}
Niklas Muennighoff, Qian Liu, Armel Zebaze, Qinkai Zheng, Binyuan Hui, Terry~Yue Zhuo, Swayam Singh, Xiangru Tang, Leandro von Werra, and Shayne Longpre.
\newblock {OctoPack}: Instruction tuning code large language models.
\newblock {\em arXiv preprint arXiv:2308.07124}, abs/2308.07124, 2023.

\bibitem{gpt4}
OpenAI.
\newblock Gpt-4 technical report.
\newblock {\em arXiv preprint arXiv:2303.08774}, 2023.

\bibitem{gpt45}
OpenAI.
\newblock Introducing gpt-4.5, 2025.

\bibitem{runbugrun}
Julian~Aron Prenner and Romain Robbes.
\newblock Runbugrun -- an executable dataset for automated program repair.
\newblock {\em arXiv preprint arXiv:2304.01102}, 2023.

\bibitem{clip}
Alec Radford, Jong~Wook Kim, Chris Hallacy, Aditya Ramesh, Gabriel Goh, Sandhini Agarwal, Girish Sastry, Amanda Askell, Pamela Mishkin, Jack Clark, et~al.
\newblock Learning transferable visual models from natural language supervision.
\newblock In {\em International conference on machine learning}, pages 8748--8763. PMLR, 2021.

\bibitem{gpt}
Alec Radford, Karthik Narasimhan, Tim Salimans, Ilya Sutskever, et~al.
\newblock Improving language understanding by generative pre-training.
\newblock {\em OpenAI blog}, 2018.

\bibitem{code_llama}
Baptiste Rozi{\`e}re, Jonas Gehring, Fabian Gloeckle, Sten Sootla, Itai Gat, Xiaoqing~Ellen Tan, Yossi Adi, Jingyu Liu, Tal Remez, J{\'e}r{\'e}my Rapin, et~al.
\newblock Code llama: Open foundation models for code.
\newblock {\em arXiv preprint arXiv:2308.12950}, 2023.

\bibitem{bloom}
Teven~Le Scao, Angela Fan, Christopher Akiki, Ellie Pavlick, Suzana Ili{\'c}, Daniel Hesslow, Roman Castagn{\'e}, Alexandra~Sasha Luccioni, Fran{\c{c}}ois Yvon, Matthias Gall{\'e}, et~al.
\newblock Bloom: A 176b-parameter open-access multilingual language model.
\newblock {\em arXiv preprint arXiv:2211.05100}, 2022.

\bibitem{doubao_thinking}
ByteDance Seed, Yufeng Yuan, Yu~Yue, Mingxuan Wang, Xiaochen Zuo, Jiaze Chen, Lin Yan, Wenyuan Xu, Chi Zhang, Xin Liu, et~al.
\newblock Seed-thinking-v1.5: Advancing superb reasoning models with reinforcement learning.
\newblock {\em arXiv preprint arXiv:2504.13914}, 2025.

\bibitem{shao2024visual}
Hao Shao, Shengju Qian, Han Xiao, Guanglu Song, Zhuofan Zong, Letian Wang, Yu~Liu, and Hongsheng Li.
\newblock Visual cot: Unleashing chain-of-thought reasoning in multi-modal language models.
\newblock {\em arXiv preprint arXiv:2403.16999}, 2024.

\bibitem{shi2024chartmimic}
Chufan Shi, Cheng Yang, Yaxin Liu, Bo~Shui, Junjie Wang, Mohan Jing, Linran Xu, Xinyu Zhu, Siheng Li, Yuxiang Zhang, et~al.
\newblock Chartmimic: Evaluating lmm's cross-modal reasoning capability via chart-to-code generation.
\newblock {\em arXiv preprint arXiv:2406.09961}, 2024.

\bibitem{design2code}
Chenglei Si, Yanzhe Zhang, Zhengyuan Yang, Ruibo Liu, and Diyi Yang.
\newblock Design2code: How far are we from automating front-end engineering?
\newblock {\em arXiv preprint arXiv:2403.03163}, 2024.

\bibitem{flowvqa}
Shubhankar Singh, Purvi Chaurasia, Yerram Varun, Pranshu Pandya, Vatsal Gupta, Vivek Gupta, and Dan Roth.
\newblock Flowvqa: Mapping multimodal logic in visual question answering with flowcharts.
\newblock {\em arXiv preprint arXiv:2406.19237}, 2024.

\bibitem{unicoder}
Tao Sun, Linzheng Chai, Jian Yang, Yuwei Yin, Hongcheng Guo, Jiaheng Liu, Bing Wang, Liqun Yang, and Zhoujun Li.
\newblock {U}ni{C}oder: Scaling code large language model via universal code.
\newblock In Lun-Wei Ku, Andre Martins, and Vivek Srikumar, editors, {\em Proceedings of the 62nd Annual Meeting of the Association for Computational Linguistics (Volume 1: Long Papers)}, pages 1812--1824, Bangkok, Thailand, August 2024. Association for Computational Linguistics.

\bibitem{mtvqa}
Jingqun Tang, Qi~Liu, Yongjie Ye, Jinghui Lu, Shu Wei, Chunhui Lin, Wanqing Li, Mohamad Fitri Faiz~Bin Mahmood, Hao Feng, Zhen Zhao, et~al.
\newblock Mtvqa: Benchmarking multilingual text-centric visual question answering.
\newblock {\em arXiv preprint arXiv:2405.11985}, 2024.

\bibitem{flowchartqa}
Simon Tannert, Marcelo~G Feighelstein, Jasmina Bogojeska, Joseph Shtok, Assaf Arbelle, Peter~WJ Staar, Anika Schumann, Jonas Kuhn, and Leonid Karlinsky.
\newblock Flowchartqa: the first large-scale benchmark for reasoning over flowcharts.
\newblock In {\em Proceedings of the 1st Workshop on Linguistic Insights from and for Multimodal Language Processing}, pages 34--46, 2023.

\bibitem{gemma3}
Gemma Team, Aishwarya Kamath, Johan Ferret, Shreya Pathak, Nino Vieillard, Ramona Merhej, Sarah Perrin, Tatiana Matejovicova, Alexandre Ram{\'e}, Morgane Rivi{\`e}re, et~al.
\newblock Gemma 3 technical report.
\newblock {\em arXiv preprint arXiv:2503.19786}, 2025.

\bibitem{kimivl}
Kimi Team, Angang Du, Bohong Yin, Bowei Xing, Bowen Qu, Bowen Wang, Cheng Chen, Chenlin Zhang, Chenzhuang Du, Chu Wei, Congcong Wang, Dehao Zhang, Dikang Du, Dongliang Wang, Enming Yuan, Enzhe Lu, Fang Li, Flood Sung, Guangda Wei, Guokun Lai, Han Zhu, Hao Ding, Hao Hu, Hao Yang, Hao Zhang, Haoning Wu, Haotian Yao, Haoyu Lu, Heng Wang, Hongcheng Gao, Huabin Zheng, Jiaming Li, Jianlin Su, Jianzhou Wang, Jiaqi Deng, Jiezhong Qiu, Jin Xie, Jinhong Wang, Jingyuan Liu, Junjie Yan, Kun Ouyang, Liang Chen, Lin Sui, Longhui Yu, Mengfan Dong, Mengnan Dong, Nuo Xu, Pengyu Cheng, Qizheng Gu, Runjie Zhou, Shaowei Liu, Sihan Cao, Tao Yu, Tianhui Song, Tongtong Bai, Wei Song, Weiran He, Weixiao Huang, Weixin Xu, Xiaokun Yuan, Xingcheng Yao, Xingzhe Wu, Xinxing Zu, Xinyu Zhou, Xinyuan Wang, Y.~Charles, Yan Zhong, Yang Li, Yangyang Hu, Yanru Chen, Yejie Wang, Yibo Liu, Yibo Miao, Yidao Qin, Yimin Chen, Yiping Bao, Yiqin Wang, Yongsheng Kang, Yuanxin Liu, Yulun Du, Yuxin Wu, Yuzhi Wang, Yuzi Yan, Zaida Zhou, Zhaowei Li, Zhejun
  Jiang, Zheng Zhang, Zhilin Yang, Zhiqi Huang, Zihao Huang, Zijia Zhao, and Ziwei Chen.
\newblock {Kimi-VL} technical report, 2025.

\bibitem{qvq-72b-preview}
Qwen Team.
\newblock Qvq: To see the world with wisdom, December 2024.

\bibitem{debugbench}
Runchu Tian, Yining Ye, Yujia Qin, Xin Cong, Yankai Lin, Zhiyuan Liu, and Maosong Sun.
\newblock Debugbench: Evaluating debugging capability of large language models.
\newblock {\em arXiv preprint arXiv:2401.04621}, 2024.

\bibitem{llama2}
Hugo Touvron, Louis Martin, Kevin Stone, Peter Albert, Amjad Almahairi, Yasmine Babaei, Nikolay Bashlykov, Soumya Batra, Prajjwal Bhargava, Shruti Bhosale, et~al.
\newblock Llama 2: Open foundation and fine-tuned chat models.
\newblock {\em arXiv preprint arXiv:2307.09288}, 2023.

\bibitem{trinh2024solvingolympiad}
Trieu~H Trinh, Yuhuai Wu, Quoc~V Le, He~He, and Thang Luong.
\newblock Solving olympiad geometry without human demonstrations.
\newblock {\em Nature}, 625(7995):476--482, 2024.

\bibitem{codevision}
Hanbin Wang, Xiaoxuan Zhou, Zhipeng Xu, Keyuan Cheng, Yuxin Zuo, Kai Tian, Jingwei Song, Junting Lu, Wenhui Hu, and Xueyang Liu.
\newblock Code-vision: Evaluating multimodal llms logic understanding and code generation capabilities.
\newblock {\em arXiv preprint arXiv:2502.11829}, 2025.

\bibitem{wang2023delving}
Liran Wang, Xunzhu Tang, Yichen He, Changyu Ren, Shuhua Shi, Chaoran Yan, and Zhoujun Li.
\newblock Delving into commit-issue correlation to enhance commit message generation models.
\newblock In {\em 2023 38th IEEE/ACM International Conference on Automated Software Engineering (ASE)}, pages 710--722. IEEE, 2023.

\bibitem{qwen_vl2}
Peng Wang, Shuai Bai, Sinan Tan, Shijie Wang, Zhihao Fan, Jinze Bai, Keqin Chen, Xuejing Liu, Jialin Wang, Wenbin Ge, et~al.
\newblock Qwen2-vl: Enhancing vision-language model's perception of the world at any resolution.
\newblock {\em arXiv preprint arXiv:2409.12191}, 2024.

\bibitem{codet5}
Yue Wang, Weishi Wang, Shafiq Joty, and Steven~CH Hoi.
\newblock Codet5: Identifier-aware unified pre-trained encoder-decoder models for code understanding and generation.
\newblock {\em arXiv preprint arXiv:2109.00859}, 2021.

\bibitem{odex}
Zhiruo Wang, Shuyan Zhou, Daniel Fried, and Graham Neubig.
\newblock Execution-based evaluation for open-domain code generation.
\newblock In {\em Findings of the Association for Computational Linguistics: EMNLP 2023}, pages 1271--1290, 2023.

\bibitem{wang2024charxiv}
Zirui Wang, Mengzhou Xia, Luxi He, Howard Chen, Yitao Liu, Richard Zhu, Kaiqu Liang, Xindi Wu, Haotian Liu, Sadhika Malladi, et~al.
\newblock Charxiv: Charting gaps in realistic chart understanding in multimodal llms.
\newblock {\em arXiv preprint arXiv:2406.18521}, 2024.

\bibitem{magicoder}
Yuxiang Wei, Zhe Wang, Jiawei Liu, Yifeng Ding, and Lingming Zhang.
\newblock Magicoder: Source code is all you need.
\newblock {\em arXiv preprint arXiv:2312.02120}, abs/2312.02120, 2023.

\bibitem{wu2024plot2code}
Chengyue Wu, Yixiao Ge, Qiushan Guo, Jiahao Wang, Zhixuan Liang, Zeyu Lu, Ying Shan, and Ping Luo.
\newblock Plot2code: A comprehensive benchmark for evaluating multi-modal large language models in code generation from scientific plots.
\newblock {\em arXiv preprint arXiv:2405.07990}, 2024.

\bibitem{deepseekvl2}
Zhiyu Wu, Xiaokang Chen, Zizheng Pan, Xingchao Liu, Wen Liu, Damai Dai, Huazuo Gao, Yiyang Ma, Chengyue Wu, Bingxuan Wang, Zhenda Xie, Yu~Wu, Kai Hu, Jiawei Wang, Yaofeng Sun, Yukun Li, Yishi Piao, Kang Guan, Aixin Liu, Xin Xie, Yuxiang You, Kai Dong, Xingkai Yu, Haowei Zhang, Liang Zhao, Yisong Wang, and Chong Ruan.
\newblock Deepseek-vl2: Mixture-of-experts vision-language models for advanced multimodal understanding, 2024.

\bibitem{osworld}
Tianbao Xie, Danyang Zhang, Jixuan Chen, Xiaochuan Li, Siheng Zhao, Ruisheng Cao, Toh~Jing Hua, Zhoujun Cheng, Dongchan Shin, Fangyu Lei, et~al.
\newblock Osworld: Benchmarking multimodal agents for open-ended tasks in real computer environments.
\newblock {\em arXiv preprint arXiv:2404.07972}, 2024.

\bibitem{qwen2}
An~Yang, Baosong Yang, Binyuan Hui, Bo~Zheng, Bowen Yu, Chang Zhou, Chengpeng Li, Chengyuan Li, Dayiheng Liu, Fei Huang, et~al.
\newblock Qwen2 technical report.
\newblock {\em arXiv preprint arXiv:2407.10671}, 2024.

\bibitem{swe_multimodal}
John Yang, Carlos~E Jimenez, Alex~L Zhang, Kilian Lieret, Joyce Yang, Xindi Wu, Ori Press, Niklas Muennighoff, Gabriel Synnaeve, Karthik~R Narasimhan, et~al.
\newblock Swe-bench multimodal: Do ai systems generalize to visual software domains?
\newblock {\em arXiv preprint arXiv:2410.03859}, 2024.

\bibitem{yang2024matplotagent}
Zhiyu Yang, Zihan Zhou, Shuo Wang, Xin Cong, Xu~Han, Yukun Yan, Zhenghao Liu, Zhixing Tan, Pengyuan Liu, Dong Yu, et~al.
\newblock Matplotagent: Method and evaluation for llm-based agentic scientific data visualization.
\newblock {\em arXiv preprint arXiv:2402.11453}, 2024.

\bibitem{minicpm_v}
Yuan Yao, Tianyu Yu, Ao~Zhang, Chongyi Wang, Junbo Cui, Hongji Zhu, Tianchi Cai, Haoyu Li, Weilin Zhao, Zhihui He, et~al.
\newblock Minicpm-v: A gpt-4v level mllm on your phone.
\newblock {\em arXiv preprint arXiv:2408.01800}, 2024.

\bibitem{mplug}
Qinghao Ye, Haiyang Xu, Guohai Xu, Jiabo Ye, Ming Yan, Yiyang Zhou, Junyang Wang, Anwen Hu, Pengcheng Shi, Yaya Shi, et~al.
\newblock mplug-owl: Modularization empowers large language models with multimodality.
\newblock {\em arXiv preprint arXiv:2304.14178}, 2023.

\bibitem{web2code}
Sukmin Yun, Haokun Lin, Rusiru Thushara, Mohammad~Qazim Bhat, Yongxin Wang, Zutao Jiang, Mingkai Deng, Jinhong Wang, Tianhua Tao, Junbo Li, et~al.
\newblock Web2code: A large-scale webpage-to-code dataset and evaluation framework for multimodal llms.
\newblock {\em arXiv preprint arXiv:2406.20098}, 2024.

\bibitem{humanevalv}
Fengji Zhang, Linquan Wu, Huiyu Bai, Guancheng Lin, Xiao Li, Xiao Yu, Yue Wang, Bei Chen, and Jacky Keung.
\newblock Humaneval-v: Evaluating visual understanding and reasoning abilities of large multimodal models through coding tasks.
\newblock {\em arXiv preprint arXiv:2410.12381}, 2024.

\bibitem{zhang2024codev}
Linhao Zhang, Daoguang Zan, Quanshun Yang, Zhirong Huang, Dong Chen, Bo~Shen, Tianyu Liu, Yongshun Gong, Pengjie Huang, Xudong Lu, et~al.
\newblock Codev: Issue resolving with visual data.
\newblock {\em arXiv preprint arXiv:2412.17315}, 2024.

\bibitem{internlm_xcomposer}
Pan Zhang, Xiaoyi Dong, Yuhang Zang, Yuhang Cao, Rui Qian, Lin Chen, Qipeng Guo, Haodong Duan, Bin Wang, Linke Ouyang, et~al.
\newblock Internlm-xcomposer-2.5: A versatile large vision language model supporting long-contextual input and output.
\newblock {\em arXiv preprint arXiv:2407.03320}, 2024.

\bibitem{EvalGPTFix}
Quanjun Zhang, Tongke Zhang, Juan Zhai, Chunrong Fang, Bowen Yu, Weisong Sun, and Zhenyu Chen.
\newblock A critical review of large language model on software engineering: An example from chatgpt and automated program repair.
\newblock {\em arXiv preprint arXiv:2310.08879}, 2023.

\bibitem{mavis}
Renrui Zhang, Xinyu Wei, Dongzhi Jiang, Yichi Zhang, Ziyu Guo, Chengzhuo Tong, Jiaming Liu, Aojun Zhou, Bin Wei, Shanghang Zhang, et~al.
\newblock Mavis: Mathematical visual instruction tuning.
\newblock {\em arXiv preprint arXiv:2407.08739}, 2024.

\bibitem{multimodalselfinstruct}
Wenqi Zhang, Zhenglin Cheng, Yuanyu He, Mengna Wang, Yongliang Shen, Zeqi Tan, Guiyang Hou, Mingqian He, Yanna Ma, Weiming Lu, et~al.
\newblock Multimodal self-instruct: Synthetic abstract image and visual reasoning instruction using language model.
\newblock {\em arXiv preprint arXiv:2407.07053}, 2024.

\bibitem{pybench}
Yaolun Zhang, Yinxu Pan, Yudong Wang, Jie Cai, Zhi Zheng, Guoyang Zeng, and Zhiyuan Liu.
\newblock Pybench: Evaluating llm agent on various real-world coding tasks.
\newblock {\em arXiv preprint arXiv:2407.16732}, 2024.

\bibitem{zhao2025chartcoder}
Xuanle Zhao, Xianzhen Luo, Qi~Shi, Chi Chen, Shuo Wang, Wanxiang Che, Zhiyuan Liu, and Maosong Sun.
\newblock Chartcoder: Advancing multimodal large language model for chart-to-code generation.
\newblock {\em arXiv preprint arXiv:2501.06598}, 2025.

\bibitem{codegeex}
Qinkai Zheng, Xiao Xia, Xu~Zou, Yuxiao Dong, Shan Wang, Yufei Xue, Zihan Wang, Lei Shen, Andi Wang, Yang Li, Teng Su, Zhilin Yang, and Jie Tang.
\newblock Codegeex: {A} pre-trained model for code generation with multilingual evaluations on humaneval-x.
\newblock {\em arXiv preprint arXiv:2303.17568}, abs/2303.17568, 2023.

\bibitem{opencodeinterpreter}
Tianyu Zheng, Ge~Zhang, Tianhao Shen, Xueling Liu, Bill~Yuchen Lin, Jie Fu, Wenhu Chen, and Xiang Yue.
\newblock Opencodeinterpreter: Integrating code generation with execution and refinement.
\newblock {\em arXiv preprint arXiv:2402.14658}, 2024.

\bibitem{llamafactory}
Yaowei Zheng, Richong Zhang, Junhao Zhang, Yanhan Ye, Zheyan Luo, Zhangchi Feng, and Yongqiang Ma.
\newblock Llamafactory: Unified efficient fine-tuning of 100+ language models.
\newblock In {\em Proceedings of the 62nd Annual Meeting of the Association for Computational Linguistics (Volume 3: System Demonstrations)}, Bangkok, Thailand, 2024. Association for Computational Linguistics.

\bibitem{webarena}
Shuyan Zhou, Frank~F. Xu, Hao Zhu, Xuhui Zhou, Robert Lo, Abishek Sridhar, Xianyi Cheng, Tianyue Ou, Yonatan Bisk, Daniel Fried, Uri Alon, and Graham Neubig.
\newblock Webarena: A realistic web environment for building autonomous agents.
\newblock In {\em The Twelfth International Conference on Learning Representations}, 2024.

\bibitem{minigpt4}
Deyao Zhu, Jun Chen, Xiaoqian Shen, Xiang Li, and Mohamed Elhoseiny.
\newblock Minigpt-4: Enhancing vision-language understanding with advanced large language models.
\newblock In {\em The Twelfth International Conference on Learning Representations}, 2023.

\bibitem{internvl3}
Jinguo Zhu, Weiyun Wang, Zhe Chen, Zhaoyang Liu, Shenglong Ye, Lixin Gu, Yuchen Duan, Hao Tian, Weijie Su, Jie Shao, et~al.
\newblock Internvl3: Exploring advanced training and test-time recipes for open-source multimodal models.
\newblock {\em arXiv preprint arXiv:2504.10479}, 2025.

\bibitem{bigcodebench}
Terry~Yue Zhuo, Minh~Chien Vu, Jenny Chim, Han Hu, Wenhao Yu, Ratnadira Widyasari, Imam Nur~Bani Yusuf, Haolan Zhan, Junda He, Indraneil Paul, et~al.
\newblock Bigcodebench: Benchmarking code generation with diverse function calls and complex instructions.
\newblock {\em arXiv preprint arXiv:2406.15877}, 2024.

\bibitem{zong2024mova}
Zhuofan Zong, Bingqi Ma, Dazhong Shen, Guanglu Song, Hao Shao, Dongzhi Jiang, Hongsheng Li, and Yu~Liu.
\newblock Mova: Adapting mixture of vision experts to multimodal context.
\newblock {\em arXiv preprint arXiv:2404.13046}, 2024.

\end{thebibliography}
